%% file: 0_main.tex
\pdfoutput=1
\documentclass{article}

\input{math_commands.tex}

\usepackage{microtype}
\usepackage{graphicx}
\usepackage{subfigure}
\usepackage{booktabs} 

\usepackage{multirow}
\usepackage{amsmath,amsfonts,amssymb, amsthm}
\usepackage{mathtools}
\usepackage{fixmath}
\usepackage{caption}
\usepackage{multirow, makecell}
\usepackage{tabularx}
\usepackage{float}
\usepackage{enumitem}
\usepackage{boxedminipage}
\usepackage{dashbox}
\usepackage{arydshln}
\usepackage{algorithm, algorithmic}
\usepackage{tikz}
\usepackage{pgfplots}
\usepgfplotslibrary{colorbrewer}

\newcommand{\rtable}[1]{\renewcommand{\arraystretch}{#1}}
\newcommand\Tstrut{\rule{0pt}{2.6ex}}         
\newcommand\Bstrut{\rule[-1.2ex]{0pt}{0pt}}   
\newcommand\TBstrut{\Tstrut\Bstrut}           

\usepackage{import}
\usepackage{pdfpages}

\newenvironment{customlegend}[1][]{%
	\begingroup
	\csname pgfplots@init@cleared@structures\endcsname
	\pgfplotsset{#1}%
}{%
	\csname pgfplots@createlegend\endcsname
	\endgroup
}%
\def\addlegendimage{\csname pgfplots@addlegendimage\endcsname}

\definecolor{azure}{HTML}{4385bc}
\definecolor{green(munsell)}{HTML}{53b251}
\definecolor{tractorred}{HTML}{e41e1e}
\definecolor{uclagold}{HTML}{9467bd}
\definecolor{tomato}{HTML}{ff7f0e}

\usepackage{hyperref}

\usepackage[accepted]{icml2020}

\icmltitlerunning{Adversarial Neural Pruning with Latent Vulnerability Suppression}

\begin{document}

\twocolumn[
\icmltitle{Adversarial Neural Pruning with Latent Vulnerability Suppression}



\icmlsetsymbol{equal}{*}

\begin{icmlauthorlist}
\icmlauthor{Divyam Madaan}{cs}
\icmlauthor{Jinwoo Shin}{ee,ai}
\icmlauthor{Sung Ju Hwang}{cs,ai,aitrics}
\end{icmlauthorlist}

\icmlaffiliation{cs}{School of Computing, KAIST, South Korea}
\icmlaffiliation{ee}{School of Electrical Engineering, KAIST, South Korea}
\icmlaffiliation{ai}{Graduate School of AI, KAIST, South Korea}
\icmlaffiliation{aitrics}{AITRICS, South Korea}

\icmlcorrespondingauthor{Divyam~Madaan}{dmadaan@kaist.ac.kr}

\icmlkeywords{Machine Learning, ICML}

\vskip 0.3in
]



\printAffiliationsAndNotice{} 

\begin{abstract}
Despite the remarkable performance of deep neural networks on various computer vision tasks, they are known to be susceptible to adversarial perturbations, which makes it challenging to deploy them in real-world safety-critical applications. In this paper, we conjecture that the leading cause of adversarial vulnerability is the distortion in the latent feature space, and provide methods to suppress them effectively. Explicitly, we define \emph{vulnerability} for each latent feature and then propose a new loss for adversarial learning, \emph{Vulnerability Suppression (VS)} loss, that aims to minimize the feature-level vulnerability during training. We further propose a Bayesian framework to prune features with high vulnerability to reduce both vulnerability and loss on adversarial samples. We validate our \emph{Adversarial Neural Pruning with Vulnerability Suppression (ANP-VS)} method on multiple benchmark datasets, on which it not only obtains state-of-the-art adversarial robustness but also improves the performance on clean examples, using only a fraction of the parameters used by the full network. Further qualitative analysis suggests that the improvements come from the suppression of feature-level vulnerability.
\end{abstract}

\input{1_introduction}
\input{2_related_work}
\input{3_motivation}
\input{4_approach}
\input{5_experiments}
\input{6_conclusion}

\bibliographystyle{icml2020}
\bibliography{refs}
\input{7_appendix}

\end{document}

%% file: math_commands.tex
\usepackage{xcolor}
\usepackage{amsmath,amsfonts,amssymb, amsthm}

\newenvironment{proof-sketch}{\noindent{\bf Sketch of Proof}\hspace*{1em}}{\qed\bigskip}
\newenvironment{proof-idea}{\noindent{\bf Proof Idea}\hspace*{1em}}{\qed\bigskip}
\newenvironment{proof-of-lemma}[1]{\noindent{\bf Proof of Lemma #1}\hspace*{1em}}{\qed\bigskip}
\newenvironment{proof-attempt}{\noindent{\bf Proof Attempt}\hspace*{1em}}{\qed\bigskip}

{\makeatletter
 \gdef\xxxmark{%
   \expandafter\ifx\csname @mpargs\endcsname\relax 
     \expandafter\ifx\csname @captype\endcsname\relax 
       \marginpar{\textcolor{red}{xxx~}}
     \else
       \textcolor{red}{xxx~}
     \fi
   \else
     \textcolor{red}{xxx~}
   \fi}
 \gdef\xxx{\@ifnextchar[\xxx@lab\xxx@nolab}
 \long\gdef\xxx@lab[#1]#2{{\bf [\xxxmark \textcolor{red}{#2} ---{\sc #1}]}}
 \long\gdef\xxx@nolab#1{{\bf [\xxxmark \textcolor{red}{#1}]}}
}

\DeclareMathOperator*{\sgn}{sgn}

\newcommand{\calB}{\ensuremath{\mathcal{B}}}

\newcommand{\calD}{\ensuremath{\mathcal{D}}}

\newcommand{\calL}{\ensuremath{\mathcal{L}}}

\theoremstyle{plain}

\theoremstyle{definition}

\DeclareMathOperator{\unif}{unif}
\DeclareMathOperator{\KL}{D_{KL}}
\DeclareMathOperator{\sgm}{sgm}

\DeclareMathOperator{\diag}{{diag}}

\renewcommand{\vec}[1]{\mathbold{#1}}

%% file: 1_introduction.tex
\section{Introduction}
In the last many years, deep neural networks (DNNs) have achieved impressive results on various computer vision tasks, e.g., image classification~\cite{alex12imagenet}, face/object recognition~\cite{DBLP:journals/corr/HeZR015,DBLP:journals/corr/abs-1801-07698}, and semantic segmentation~\cite{He2017MaskR}. The groundbreaking success of DNNs has motivated their use in safety-critical environments such as medical imaging~\cite{esteva2017dermatologist} and autonomous driving~\cite{DBLP:journals/corr/BojarskiTDFFGJM16}. However, DNNs are extremely brittle to carefully crafted imperceptible adversarial perturbations intentionally optimized to cause miss-prediction~\cite{Szegedy2013IntriguingPO,2014arXiv1412.6572G}.

While the field has primarily focused on the development of new attacks and defenses, a `cat-and-mouse' game between attacker and defender has arisen. There has been a long list of proposed defenses to mitigate the effect of adversarial examples, e.g., defenses~\cite{DBLP:journals/corr/PapernotMWJS15,Xu17feature,buckman2018thermometer,s.2018stochastic}, followed by successful attacks~\cite{DBLP:journals/corr/CarliniW16a,anish18obfuscated,DBLP:journals/corr/abs-1802-05666}. This shows that any defense mechanism that once looks successful could be circumvented with the invention of new attacks. In this paper, we tackle the problem by identifying a more fundamental cause of the adversarial vulnerability of DNNs.

What makes DNNs vulnerable to adversarial attacks? Our intuition is that the adversarial vulnerability of a DNN comes from the distortion in the \emph{latent feature} space incurred by the adversarial perturbation. If a perturbation at the input level is suppressed successfully at any layers of the DNN, then it would not cause miss-prediction. However, not all latent features will contribute equally to the distortion; some features may have more substantial distortion, by amplifying the perturbations at the input level, while others will remain relatively static. Under this motivation, we first formally define the vulnerability of the latent features and show that sparse networks can have a much smaller degree of network-level vulnerability in Figure~\ref{fig:intro_2_a}. Then, we propose an effective way to suppress vulnerability by pruning the latent features with high vulnerability. We refer to this defense mechanism as \emph{Adversarial Neural Pruning (ANP)}. Further, we propose a novel loss, which we refer to as \emph{Vulnerability Suppression (VS)} loss, that directly suppresses the vulnerability in the feature space by minimizing the feature-level vulnerability. To this end, we propose a novel defense mechanism coined, \emph{Adversarial Neural Pruning with Vulnerability Suppression (ANP-VS)}, that learns pruning masks for the features in a Bayesian framework to minimize both the adversarial loss and the feature-level vulnerability, as illustrated in Figure~\ref{fig:conceptfigure}. It effectively suppresses the distortion in the latent feature space with almost no extra cost relative to adversarial training and yields light-weighted networks that are more robust to adversarial perturbations. 

In summary, the contributions of this paper are as follows:
\input{1_concept_figure}
\begin{itemize}
	\item We hypothesize that the distortion in the latent features is the leading cause of DNN's susceptibility to adversarial attacks and formally describe the concepts of the vulnerability of latent-features based on the expectation of the distortion of latent-features with respect to input perturbations.
	
	\item Based on this finding, we introduce a novel defense mechanism, \emph{Adversarial Neural Pruning with Vulnerability Suppression (ANP-VS)}, to mitigate the feature-level vulnerability. The resulting framework learns a Bayesian pruning (dropout) mask to prune out the vulnerable features while preserving the robust ones by minimizing the adversarial and \emph{Vulnerability Suppression (VS)} loss. 
	
	\item {We experimentally validate our proposed method on MNIST, CIFAR-10, and CIFAR-100 datasets, on which it achieves state-of-the-art robustness with a substantial reduction in memory and computation, with qualitative analysis which suggests that the improvement on robustness comes from its suppression of feature-level vulnerability}.
\end{itemize}
While our primary focus is on improving the robustness of DNNs, we also found that ANP-VS achieves higher accuracy for clean/non-adversarial
inputs, compared to baseline adversarial training methods (see the results of CIFAR datasets in Table \ref{table:mainTable}). This is another essential benefit of ANP-VS as it has been well known that
adversarial training tends to hurt the clean accuracy~\cite{schmidt2018data,tsipras2018robustness, zhang2019theoretically}.
\input{1_intro_vis} 

%% file: 1_concept_figure.tex
\begin{figure}[t!]
    \centering
    \def\svgwidth{\linewidth}
    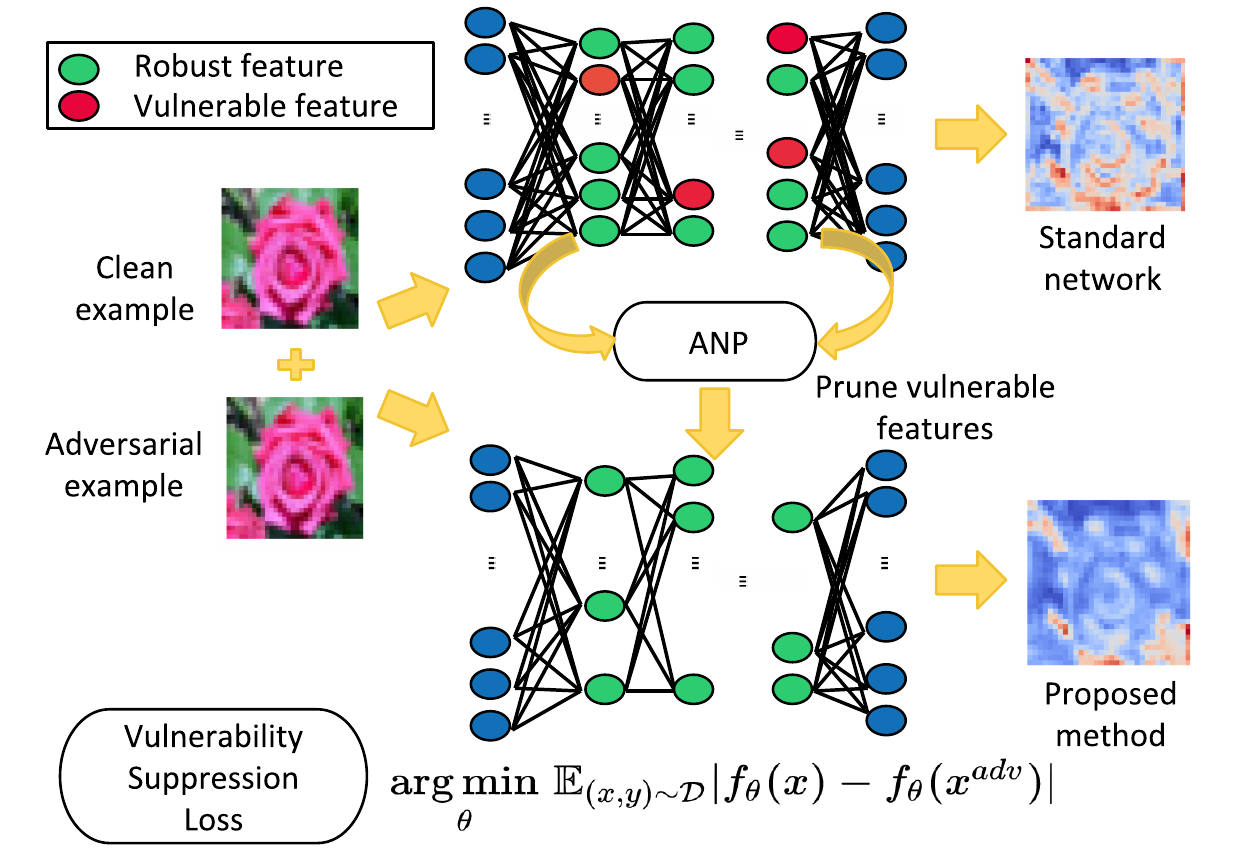

    \caption{\textbf{Concept:} We hypothesize that the distortion in the latent feature space is the leading cause of the adversarial vulnerability of deep neural networks and introduce a novel \emph{Vulnerability Suppression (VS)} loss that explicitly aims to minimize feature-level distortions. Further, we prune the features with significant distortions by learning pruning masks with \emph{Adversarial Neural Pruning (ANP)} that minimizes the adversarial loss.\label{fig:conceptfigure}}
\end{figure}

%% file: ANP_concept.pdf_tex
\begingroup%
  \makeatletter%
  \providecommand\color[2][]{%
    \errmessage{(Inkscape) Color is used for the text in Inkscape, but the package 'color.sty' is not loaded}%
    \renewcommand\color[2][]{}%
  }%
  \providecommand\transparent[1]{%
    \errmessage{(Inkscape) Transparency is used (non-zero) for the text in Inkscape, but the package 'transparent.sty' is not loaded}%
    \renewcommand\transparent[1]{}%
  }%
  \providecommand\rotatebox[2]{#2}%
  \newcommand*\fsize{\dimexpr\f@size pt\relax}%
  \newcommand*\lineheight[1]{\fontsize{\fsize}{#1\fsize}\selectfont}%
  \ifx\svgwidth\undefined%
    \setlength{\unitlength}{356.44081535bp}%
    \ifx\svgscale\undefined%
      \relax%
    \else%
      \setlength{\unitlength}{\unitlength * \real{\svgscale}}%
    \fi%
  \else%
    \setlength{\unitlength}{\svgwidth}%
  \fi%
  \global\let\svgwidth\undefined%
  \global\let\svgscale\undefined%
  \makeatother%
  \begin{picture}(1,0.69103477)%
    \lineheight{1}%
    \setlength\tabcolsep{0pt}%
    \put(0,0){\includegraphics[width=\unitlength,page=1]{ANP_concept.pdf}}%
  \end{picture}%
\endgroup%

%% file: 1_intro_vis.tex
\definecolor{s1}{RGB}{43, 140, 90}
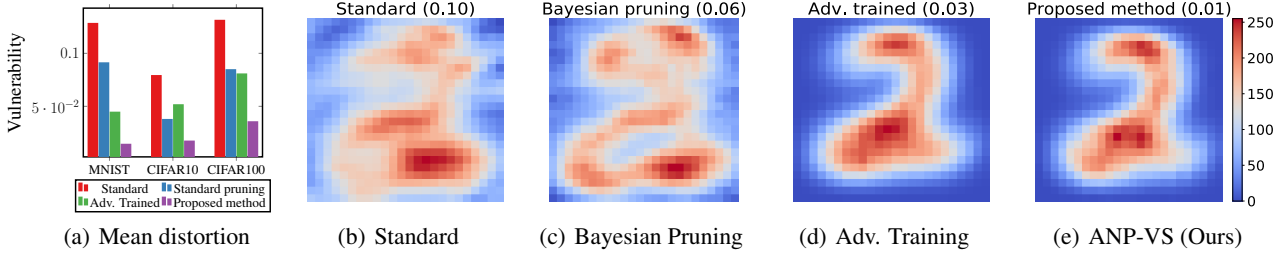
\begin{figure*}
	\begin{center}
	\label{fig:intro_results}
	\subfigure[Mean distortion]
	{\hspace{-0.2in}
		\resizebox{0.21\textwidth}{!}{
\begin{tikzpicture}
\begin{axis}[
ybar,
legend style={at={(0.5,-0.15)},	anchor=north,legend columns=2},
ylabel={Vulnerability},
xlabel={Dataset},
ylabel near ticks,
xlabel near ticks,
symbolic x coords={MNIST, CIFAR10, CIFAR100},
xtick={MNIST, CIFAR10, CIFAR100},
font=\Large,
ylabel style={font=\huge},
yticklabel style={font=\LARGE},
xlabel style={font=\huge},
enlarge x limits={abs=1cm},
style={ultra thick},
every axis plot/.append style={fill},
cycle list/Set1
]
\addplot coordinates {(MNIST,0.12802465) (CIFAR10,0.07881432) (CIFAR100,0.13095564)};
\addplot coordinates {(MNIST,0.09083603) (CIFAR10,0.03740915) (CIFAR100,0.08435563)};
\addplot coordinates {(MNIST,0.0442781) (CIFAR10,0.05128915) (CIFAR100,0.08035727)};
\addplot coordinates {(MNIST,0.014) (CIFAR10,0.017) (CIFAR100,0.035238896)};
\legend{Standard, Standard pruning, Adv. Trained, Proposed method}
\end{axis}
\end{tikzpicture}
		}
		\label{fig:intro_2_a}
	}
	\hspace{0in}
	\subfigure[Standard]
	{\includegraphics[width=2.8cm, height=2.8cm]{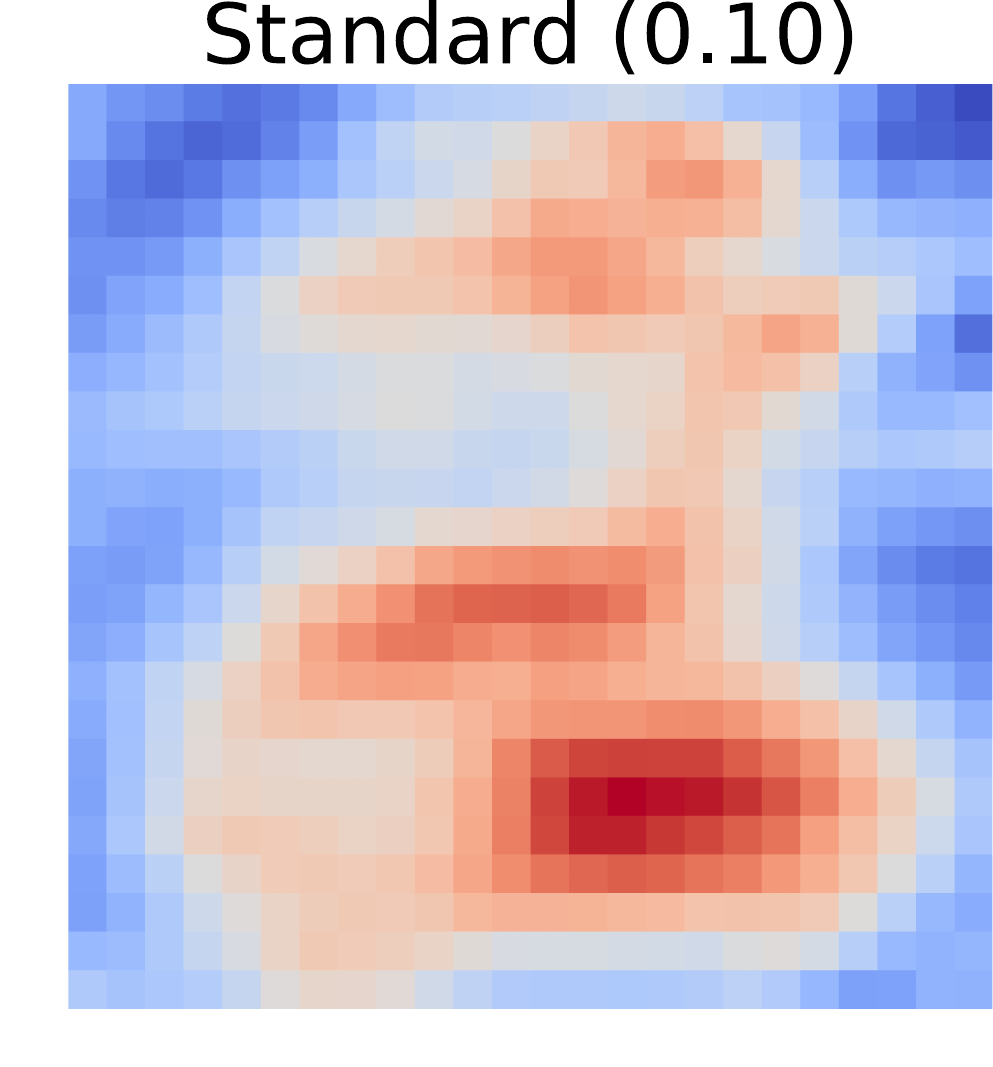}}
	\hspace{0.1in}
	\subfigure[Bayesian Pruning]
	{\includegraphics[width=2.8cm, height=2.8cm]{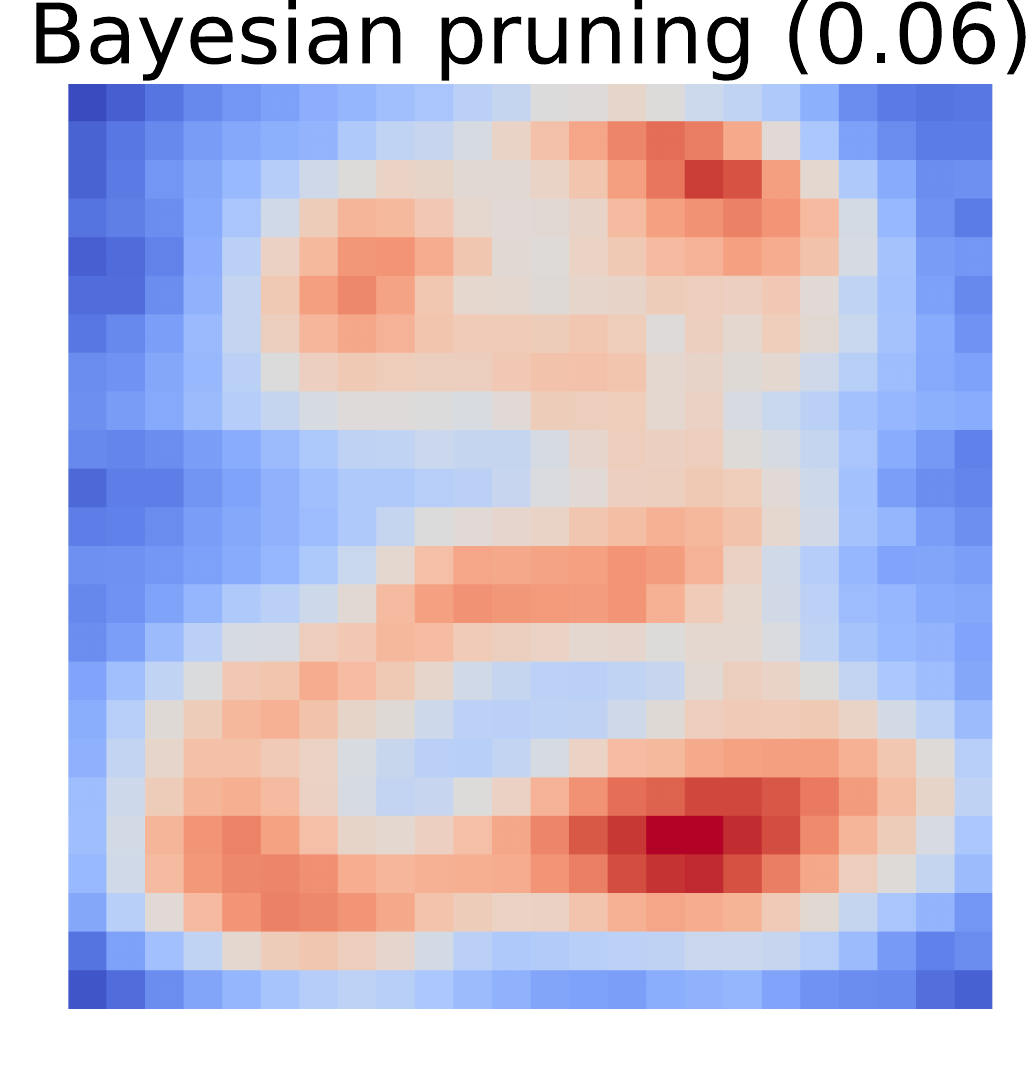}}
	\hspace{0.1in}
	\subfigure[Adv. Training]
	{\includegraphics[width=2.8cm, height=2.8cm]{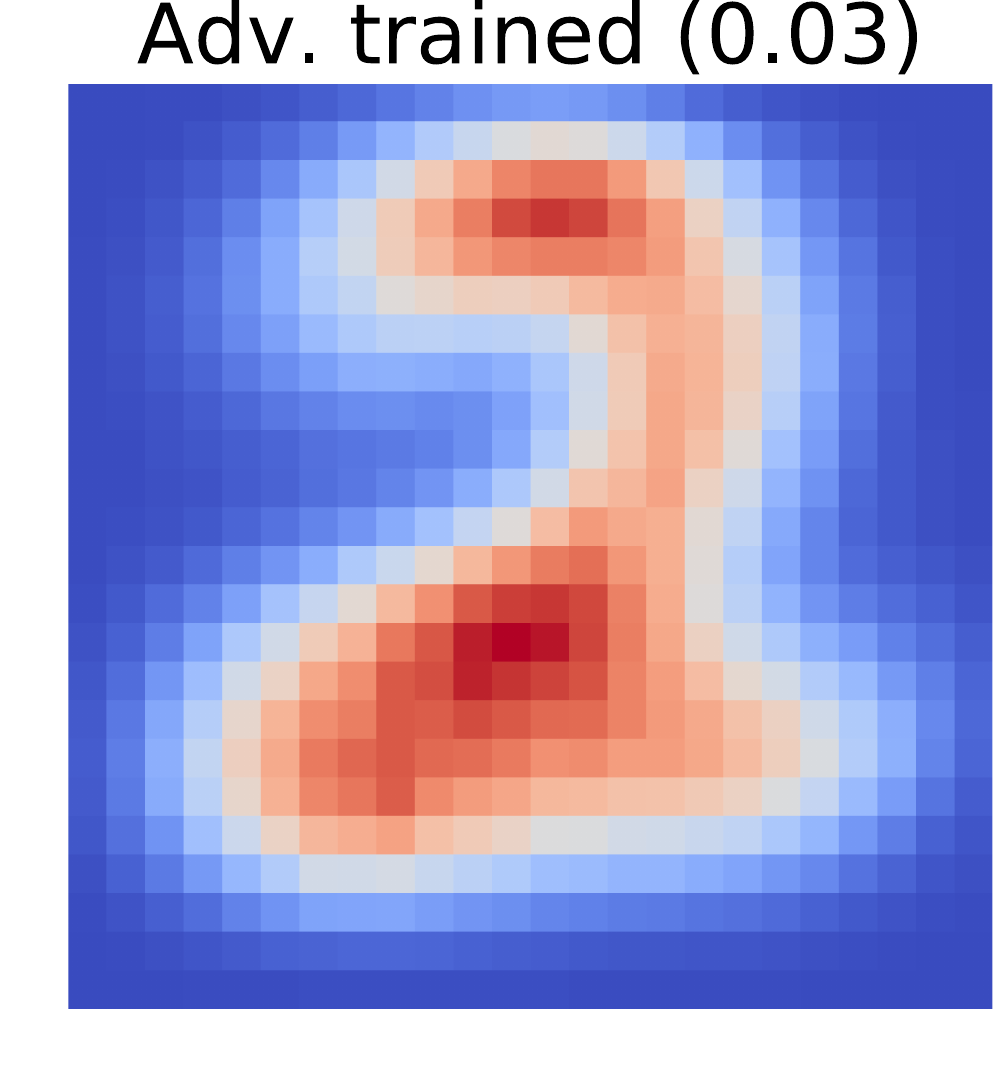}}
	\hspace{0.1in}
	\subfigure[ANP-VS (Ours)]
	{\includegraphics[width=3.35cm, height=2.8cm]{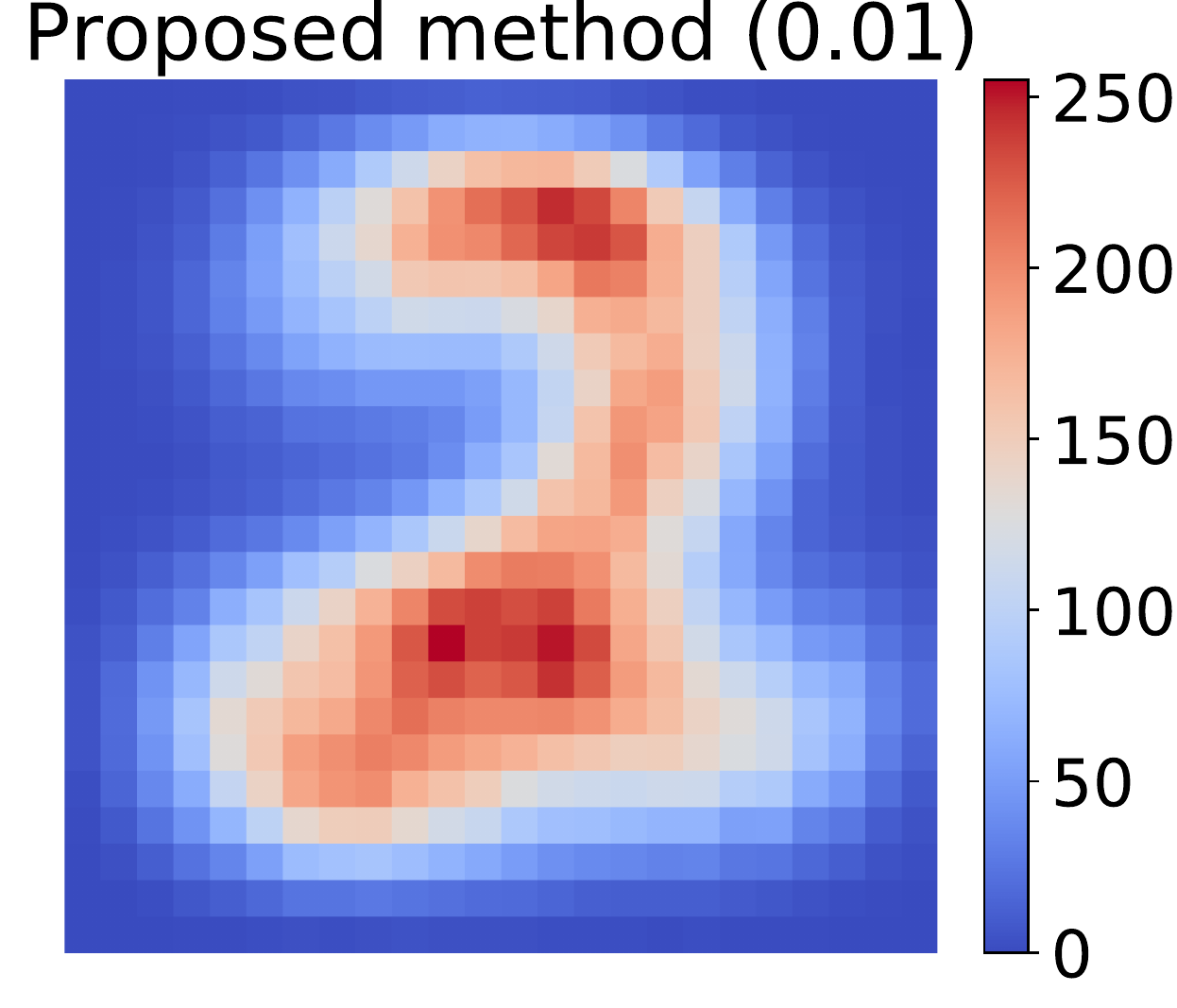}}
	\caption{(a) Mean distortion (average perturbation in latent features across all layers) for MNIST on Lenet-5-Caffe, CIFAR-10 and CIFAR-100 on VGG-16. Our method yields a network with minimum vulnerability (distortion) compared to all the other networks; we provide the formal definition of vulnerability in Section \ref{motivation}.   Visualization of the vulnerability for the input layer for Lenet-5-Caffe on MNIST for various methods where the vulnerability of the input layer is reported in the bracket (the smaller is, the better): (b) Standard trained model, (c) Standard Bayesian pruning, (d) Adversarial trained network and (e) ANP-VS. The standard network (a) has the maximum distortion, which is comparatively reduced in adversarial training (d) and further suppressed by ANP-VS (e). \label{fig:concept}}
\end{center}
\end{figure*}

%% file: 2_related_work.tex
\section{Related work} \label{related_work}
\paragraph{Adversarial robustness.} Since the literature on the adversarial robustness of neural networks is vast, we only discuss some of the most relevant studies. A large number of defenses~\cite{DBLP:journals/corr/PapernotMWJS15,DBLP:journals/corr/XuEQ17,buckman2018thermometer,s.2018stochastic,song2018pixeldefend} have been proposed and consequently broken by more sophisticated attack methods~\cite{DBLP:journals/corr/CarliniW16a,anish18obfuscated,DBLP:journals/corr/abs-1802-05666}. Adversarial training based defenses~\cite{madry2018towards,liu2018advbnn, hendrycks2019using, zhang2019theoretically} are widely considered to be the most effective since they largely avoid the obfuscated gradients problem~\cite{anish18obfuscated}. There has also been previous work that studied robust and vulnerable features at the input level.~\citet{garg17spectral} established a relation between adversarially robust features and the spectral property of the geometry of the dataset and~\citet{DBLP:journals/corr/GaoWQ17} proposed to remove unnecessary features to get robustness. Recently,~\citet{ilyas2019features} disentangle robust and non-robust features in the input-space. Our work is different from these existing works in that we consider and define the vulnerability at the latent feature level, which is more directly related to the model prediction.

\paragraph{Sparsification methods.} Sparsification of neural networks is becoming increasingly important with the increased deployments of deep network models to resource-limited devices. The most straightforward way to sparsify neural networks is by removing weights with small magnitude~\citep{Strom97sparseconnection,DBLP:journals/corr/CollinsK14}; however, such heuristics-based pruning often degenerates accuracy, and~\citet{Han2015LearningBW} proposed an iterative retraining and pruning approach to recover from the damage from pruning. 
Using sparsity-inducing regularization (e.g. $\ell_1$) is another popular approach for network sparsification. However elementwise sparsity does not yield practical speed-ups and~\citet{wei16_ssl} proposed to use group sparsity to drop a neuron or a filter as a whole, that will reduce the actual network size.~\citet{Molchanov:2017:VDS:3305890.3305939} proposed to learn the individual dropout rates per weight with sparsity-inducing priors to completely drop out unnecessary weights, and~\citet{Neklyudov2017StructuredBP} proposed to exploit structured sparsity by learning masks for each neuron or filter. \citet{lee2019adaptive} proposed a variational dropout whose dropout probabilities are drawn from sparsity-inducing beta-Bernoulli prior. Information-theoretic approaches have been also shown to be effective, such as~\citet{dai18vib} which minimizes the information theoretic bound to reduce the redundancy between layers. 

\paragraph{Robustness and sparsity.} The sparsity and robustness have been explored together in various recent works.~\citet{guo18sparsednns} analyzes sparsity and robustness from a theoretical and experimental perspective and demonstrate that appropriately higher sparsity leads to a more robust model and~\citet{ye2018defending} experimentally discuss how pruning shall effect robustness with a similar conclusion. In contrary,~\citet{Luyu18prunedrobustness} derived opposite conclusions showing that robustness decreases with increase in sparsity.~\citet{ye2019adversarial} proposed concurrent adversarial training and weight pruning to enable model compression while preserving robustness. However, ANP-VS is fundamentally different from all these methods; we sparsify networks while explicitly learning the pruning (dropout) mask to minimize the loss on adversarial examples.

%% file: 3_motivation.tex
\section{Robustness of deep representations}\label{motivation}
We first briefly introduce some notations and the concept of vulnerability in the deep latent representation space. We represent a L-layer neural network by a function $f:~\vec{X}~\rightarrow~\vec{Y}$ with dataset denoted by $\calD=\{\vec{X}, \vec{Y}\}$. Specifically, we denote any sample instance by $\vec{x}\in\vec{X}$, and it's corresponding label by $\vec{y}$. The output vector $f_{\vec{\theta}}(\vec{x})$ can then be represented by $f_{\vec{\theta}}(\vec{x}) = f_{L-1}(f_{L-2}(\cdots(f_1(\vec{x}))))$ where, $\vec{\theta} = \{\vec{W}_1,\ldots,\vec{W}_{L-1},\vec{b}_1,\ldots,\vec{b}_{L-1}\}$. 

Let $\vec{z}_l$ denote the latent-feature vector for the $l$-th layer with rectified linear unit (ReLU) as the activation function, then $f_l(\cdot)$ can be defined as:
\begin{align*}
\resizebox{\linewidth}{!}{$%
    \vec{z}_{l+1} = f_{l}(\vec{z}_l) = {\max}\{\vec{W}_l\vec{z}_l + \vec{b}_l, 0 \}, \quad \forall l\in\{1,2,\ldots ,L-2\}
    $}%
\end{align*}
where $\vec{W}_l$ and $\vec{b}_l$ denote the weight and bias vector respectively. Let $\vec{x}$ and $\tilde{\vec{x}}$ denote clean and adversarial data points, respectively ($\tilde{\vec{x}} = \vec{x} + \delta$) for any $\vec{x} \in \vec{X}$ with $\ell_p$-ball $\calB(\vec{x}, \varepsilon)$ around $\vec{x}:\{\tilde{\vec{x}} \in \vec{X}: \| \tilde{\vec{x}} - \vec{x}\| \leq \varepsilon\}$, $\vec{z}_l$ and $\tilde{\vec{z}}_l$ as their corresponding latent-feature map vectors for the $l$-th layer of the network.

\paragraph{Vulnerability of a latent-feature.}  The vulnerability of a $k$-th latent-feature for $l$-th layer can be measured by the distortion in that feature in the presence of an adversary. The vulnerability of a latent feature could then be defined as the expectation of the Manhattan distance between the feature value for a clean example ($\vec{z}_{lk}$) and its adversarial example ($\tilde{\vec{z}}_{lk}$). This could be formally defined as follows:
\begin{equation}\label{eq:feature_vulnerability}
{\rm{v}}(\vec{z}_{lk},\tilde{\vec{z}}_{lk}) = \mathbb{E}_{(\vec{x},\vec{y})\sim\mathcal{D}}|| \vec{z}_{lk} - \tilde{\vec{z}}_{lk} ||
\end{equation}

\paragraph{Vulnerability of a network.} We can measure the vulnerability of an entire network $f_{\vec{\theta}}(\vec{X})$ by computing the sum of the vulnerability of all the latent features vectors of the network before the logit layer, then ${\rm{V}}(f_{\vec{\theta}}(\vec{X}), f_{\vec{\theta}}(\tilde{\vec{X}}))$ can be defined as:
\begin{equation}
\begin{aligned}\label{eq:network_vulnerability}
\overline{{\rm{v}}_l} &= \frac{1}{N_l} \sum_{k=1}^{k=N_l} {\rm{v}}(\vec{z}_{lk},\tilde{\vec{z}}_{lk}), \\
{\rm{V}}(f_{\vec{\theta}}(\vec{X}), f_{\vec{\theta}}(\tilde{\vec{X}})) &= \frac{1}{L-2}\sum_{l=1}^{l=L-2} \overline{{\rm{v}}_l}
\end{aligned}
\end{equation}
where $\overline{{\rm{v}}_l}$ represents the vulnerability of the layer $l$ composed of $N_l$ latent-features. Figure \ref{fig:intro_2_a} shows the vulnerability of different networks across various datasets. {It can be observed that although adversarial training suppresses the vulnerability at the input level, the latent feature space is still vulnerable to adversarial perturbation and that our proposed method achieves the minimum distortion.}
\paragraph{Adversarial training.}
Adversarial Training~\cite{2014arXiv1412.6572G,DBLP:journals/corr/KurakinGB16a} was proposed as a data augmentation method to train the network on the mixture of clean and adversarial examples until the loss converges. ~\citet{madry2018towards} incorporated the adversarial search inside the training process by solving the following non-convex outer minimization problem and a non-concave inner maximization problem:
\begin{equation}\label{eq:adv_madry}
\min_{\vec{\theta}}~ \mathbb{E}_{(\vec{x},\vec{y}) \sim \calD} \biggl[\max_{\delta \in \calB(\vec{x}, \varepsilon)} \calL(\vec{\theta}, \tilde{\vec{x}}, \vec{y})  \biggr]
\end{equation}
Figure~\ref{fig:concept} shows that adversarial training can distinguish between robust and vulnerable features. While the standard training results in obtaining vulnerable features, adversarial training reduces the vulnerability of the latent-features and selects the robust features which are necessarily required to attain adversarial robustness.

\paragraph{Relationship between robustness and sparsity.}~\citet{guo18sparsednns, ye2018defending} shows that sparsifying networks leads to more robust networks, which is evident by our definitions: sparsity suppresses vulnerability to \(0\) and thus reduces the network vulnerability.~\citet{ye2019adversarial} uses a pre-trained adversarial defense model and investigated different pruning schemes for robustness. However, the network still does not take into account the robustness of a latent-feature. To address these limitations, we introduce \emph{ANP-VS} in the next section, which is a novel method to prune and suppress these vulnerable features and only requires a standard network. In the experiments section, we show that our method significantly outperforms their method in adversarial robustness and computational efficiency.

%% file: 4_approach.tex
\section{Adversarial neural pruning with vulnerability suppression}\label{approach}
In this section, we propose our method to reduce the vulnerability in the latent space. Let \(\calL(\vec{\theta}\odot \vec{M}, \vec{x}, \vec{y})\) be the loss function at data point \(\vec{x}\) with class \(\vec{y}\) for any $\vec{x} \in \vec{X}$ for the model with parameters $\vec{\theta}$ and mask parameters \(\vec{M}\), we use Projected Gradient Descent (PGD)~\cite{madry2018towards} to generate the adversarial examples:
\begin{equation}\label{eq:pgd_attack}
\tilde{\vec{x}}^{k+1} = \underset{\calB(\vec{x}, \varepsilon)}{\prod}(\tilde{\vec{x}}^k + \alpha\cdot \sgn(\triangledown_{\vec{x}} \calL(\vec{\theta} \odot \vec{M}, \tilde{\vec{x}}^k,\vec{y})))
\end{equation}
where $\alpha$ is the step size, $\sgn(\cdot)$ returns the sign of the vector, $\odot$ is the hadamard product, $\prod(\cdot)$ is the projection operator on \(\ell_{\infty}\) norm-ball $\calB(\vec{x}, \varepsilon)$ around $\vec{x}$ with radius $\varepsilon$ for each example, and $\vec{x}^{k+1}$ denotes the adversarial example at the $k$-th PGD step. Specifically, PGD perturbs the clean example $\vec{x}$ for $K$ number of steps, and projects the adversarial example $\tilde{x}^{k+1}$ onto the norm-ball of $\vec{x}$, if it goes beyond $\calB(\vec{x}, \varepsilon)$ after each step of perturbation.

In order to minimize the vulnerability of the network, we first propose \emph{Adversarial Neural Pruning (ANP)}.  The basic idea of ANP is to achieve robustness while suppressing the distortion, by explicitly pruning out the latent features with high distortion. Specifically, ANP learns pruning masks for the features in a Bayesian framework to minimize the following adversarial loss:
\begin{equation}\label{eq:mask_loss}
\min_{\vec{M}}~~ \mathbb{E}_{(\vec{x},\vec{y}) \sim \calD}\biggl\{\max_{\delta \in \calB(\vec{x}, \varepsilon)} \mathcal{L}(\vec{\theta} \odot \vec{M}, \tilde{\vec{x}}, \vec{y})  \biggr\}
\end{equation}
We further combine our proposed pruning scheme with our novel \emph{vulnerability suppression loss (VS)} that minimizes the network vulnerability and further improves the robustness of the model. Let $J(\vec{\theta}\odot\vec{M}, \vec{x}, \vec{y})$ be the adversarial training loss on a batch of samples (comprised of the cross-entropy loss on the clean and adversarial samples, plus any weight decay, etc.). In particular, Adversarial Neural Pruning with Vulnerability Suppression (ANP-VS) optimizes the network parameters $\vec{\theta}$ using the following loss:
\begin{equation}\label{eq:weight_loss}
\begin{aligned}
\min_{\vec{\theta}}~ \mathbb{E}_{(\vec{x},\vec{y}) \sim \calD} &~ \biggl\{\underbrace{J(\vec{\theta}\odot\vec{M}, \vec{x}, \vec{y})}_{\text{classification loss}} + \underbrace{\lambda\cdot {\rm{V}}(f_{\vec{\theta}}(\vec{x}), f_{\vec{\theta}}(\tilde{\vec{x}}))}_{\text{vulnerability suppression loss}}\biggr\}
\end{aligned}
\end{equation}
where $\lambda$ is the hyper-parameter determining the strength of the vulnerability suppression loss. The overall training algorithm is displayed in Algorithm~\ref{alg:algorithm}.

\paragraph{Intution behind ANP-VS.}
ANP in Equation~\eqref{eq:mask_loss} encourages the pruning mask to be optimized by minimizing the adversarial loss and preserving the robustness of the model. The VS loss encourages the suppression of the distortion of the latent features in the presence of adversarial perturbations via minimizing the difference between latent-features of clean examples and those of adversarial examples. We empirically found that ANP-VS also has an effect of increasing the smoothness of the model's output and its loss surface, which is conceptually consistent with the indispensable properties of robust models~\cite{cisse2017parseval, tsipras2018robustness} (see Figure \ref{fig:loss_surface}).

\input{4_algorithm}
\paragraph{Adversarial beta-Bernoulli dropout.}
Beta-Bernoulli Dropout~\cite{lee2019adaptive} learns to set the dropout rate by generating the dropout mask from sparsity-inducing beta-Bernoulli prior to each neuron. 
Let $\vec{\theta} \in \mathbb{R}^{K\times L \times M}$ be a parameter tensor of neural network layer with $K$ channels and \(\vec{M} = \{\vec{m}_1,\ldots,\vec{m}_n\}\) for any $\vec{m} \in \{0,1\}^K$ be the binary mask sampled from the finite-dimensional beta-Bernoulli prior to be applied for the n-th observation \(\vec{x}_{n}\).

The goal of the variational inference is to compute the posterior distribution $p(\vec{\theta}, \vec{M}, \vec{\pi}|\calD)$ and we approximate this posterior using an approximate variational distribution $q(\vec{\theta},\vec{M}, \vec{\pi}|\tilde{\vec{X}})$ of known parametric form. For \(\vec{\pi}\), we use the Kumaraswamy distribution~\cite{KUMARASWAMY198079} with parameters \(a\) and \(b\) following following~\citet{lee2019adaptive, nalisnick2016stick}, and \(\vec{m}_{k}\) is sampled by reparametrization with continuous relaxation:
\begin{equation} 
\begin{aligned}
q(\pi_{k};a_{k}, b_{k}) = a_{k}b_{k}\pi_{k}^{a_{k} - 1}(1 - \pi_{k}^{a_{k}})^{b_{k}-1} \\
	{\vec{m}_{k}} = \sgm \left( \frac{1}{\tau} \left( \log \frac{\pi_{k}}{1 - {\pi}_{k}} + \log \frac{u}{1 - u} \right)  \right) 
\end{aligned}
\end{equation}

where \(\sgm(x) = \frac{1}{1 + e^{-x}}\), \(u \sim \unif[0,1]\), and \(\tau\) is a temperature continuous relaxation. The KL-divergence between the prior and variational distribution can then be computed in a closed form~\cite{nalisnick2016stick}:
\begin{align}\label{eq:bbdrop_kl}
& D_{KL}[q(\vec{M}, \vec{\pi})\Vert p(\vec{M}, \vec{\pi})] \nonumber \\
&= \sum_{k=1}^K \Bigg\{ \frac{a_k-\alpha/K}{a_k}\bigg( - \gamma - \Psi(b_k) 
 -\frac{1}{b_k}\bigg) \nonumber \\
& \qquad \qquad + 
\log \frac{a_k b_k}{\alpha/K} - \frac{b_k-1}{b_k}\Bigg\}
\end{align}  
where \(\gamma\) is Euler-Mascheroni constant, and \(\Psi(.)\) is digamma function. Using the Stochastic Gradient Variational Bayes (SGVB) framework~\cite{kingma2015sgvb}, we can optimize the variational parameters and get the final loss as follows:
\begin{equation}\label{eq:bb_final_loss}
\begin{aligned}
 \min_{\vec{M}}~~\biggl\{ &\sum_{n=1}^{N} \mathbb{E}_{q} [\log p (\vec{y}_{n}|f(\tilde{\vec{x}}_{n}; \vec{\theta}\odot \vec{M}))] \\&- \beta \cdot D_{KL}[q(\vec{M};\vec{\pi})||p(\vec{M}|\vec{\pi})] \biggr\}\\
\end{aligned}
\end{equation}
where $\beta$ is the trade-off parameter between network robustness and pruned network size to individually tailor the degree of compression across each layer of the network. The first term in the loss measures the log-likelihood of the adversarial samples w.r.t. \(q(\vec{M}; \vec{\pi})\) and the second term regularizes \(q(\vec{M}; \vec{\pi})\) so it doesn't deviate from the prior distribution. We refer ANP-VS to Adversarial Beta Bernoulli dropout with VS for the rest of our paper. We further extend ANP-VS to the Variational information bottleneck~\cite{dai18vib} in the supplementary material. 

%% file: 4_algorithm.tex
\begin{algorithm}[t]
\begin{algorithmic}[1]
\INPUT Dataset $\calD$, training iterations $T$, trained model $f_{\vec{\theta}}$ with parameters $\vec{\theta}$, mask parameters $\vec{M}$, batch size~$n$,  PGD iterations K, and PGD step size $\alpha$ for some norm-ball $\mathcal{B}$.
\OUTPUT  Pruned state of network $f_{\vec{\theta}}$
\FOR{\text{$t = \{1, \dots, T\}$}}
\STATE{\text{Sample mini-batch $B = \{\vec{x}_1, \dots, \vec{x}_m\} \subset \calD$}}
\FOR{\text{$i = \{1, \dots, n\}$}}
\STATE \textit{// Run PGD adversary}
\FOR{\text{$k = \{1, \dots, K\}$}}
    \STATE \small{$\tilde{\vec{x}}^{k+1}_i = \underset{\calB(\vec{x_i}, \varepsilon)}{\prod}(\tilde{\vec{x}}^{k}_i + \alpha \sgn(\triangledown_{\vec{x}} \calL(\vec{\theta} \odot \vec{M}, \tilde{\vec{x}}^{k}_i,\vec{y}_i)))$}
\ENDFOR
    \STATE Optimize $\vec{\theta}$ by Equation \ref{eq:weight_loss} and $\vec{M}$ by Equation \ref{eq:mask_loss} using gradient descent.
    \ENDFOR
    \ENDFOR
 \caption{Adversarial training by ANP-VS \label{alg:algorithm}}
\end{algorithmic}
\end{algorithm}

%% file: 5_experiments.tex
\section{Experiments}\label{experiments}
\input{table_1}
\input{table_2}
\subsection{Experimental setup}
\paragraph{Baselines and our model.} We first introduce various baselines and our model. We compare against following adversarial robustness and compression baselines:
\begin{enumerate}[itemsep=0em, topsep=-1ex, itemindent=0em, leftmargin=1.2em, partopsep=0em]
\item{\bf Standard.} Base convolution neural network.
\item{\bf Bayesian Pruning (BP).} Base network with Beta-Bernoulli dropout~\cite{lee2019adaptive}.
\item{\bf Adversarial Training (AT).} Adversarial trained network~\cite{madry2018towards}.
\item{\bf Adversarial Bayesian Neural Network (AT BNN).} Adversarial Bayesian trained network~\cite{liu2018advbnn}.
\item{\bf Pre-trained Adversarial Training (Pretrained AT).} Adversarial training on a pre-trained base model~\cite{hendrycks2019using}.
\item{\bf Alternating Direction Method of Multipliers (ADMM).} Concurrent weight pruning and adversarial training~\cite{ye2019adversarial}.
\item{\bf Theoretically Principled Trade-off between Robustness and Accuracy (TRADES).} Explicit trade off between natural and robust generalization~\cite{zhang2019theoretically}.
\item{\bf Adversarial Neural Pruning with vulnerability suppression (ANP-VS).} Adversarial neural pruning regularized with vulnerability suppression loss.
\end{enumerate}
\paragraph{Datasets.} We validate our method on following benchmark datasets for adversarial robustness:
\begin{enumerate}[itemsep=0em, topsep=-1ex, itemindent=0em, leftmargin=1.2em, partopsep=0em]
\item{\bf MNIST.} This dataset~\cite{lecun1998mnist} contains 60,000 grey scale images of handwritten digits sized \(28\times28\), where there are 5,000 training instances and 1,000 test instances per class. As for the base network, we use LeNet 5-Caffe \footnote{https://github.com/BVLC/caffe/tree/master/examples/mnist} for this dataset.

\item{\bf CIFAR-10.} This dataset~\cite{alex12cifar} consists of 60,000 images sized \(32\times32\), from ten animal and  vehicle classes. For each class, there are 5,000 images for training and 1,000 images for test. We use VGG-16~\cite{Simonyan2015VeryDC} for this dataset with 13 convolutional and two fully connected layers with pre-activation batch normalization and Binary Dropout.

\item{\bf CIFAR-100.} This dataset~\cite{alex12cifar} also consists of 60,000 images of \(32\times32\) pixels as in CIFAR-10 but has 100 generic object classes instead of 10. Each class has 500 images for training and 100 images for test. We  use VGG-16~\cite{Simonyan2015VeryDC} similar to CIFAR-10 dataset as the base network for this dataset.
\end{enumerate}

\paragraph{Evaluation setup.} We validate our model with three metrics for computational efficiency: i) \emph{Memory footprint} (Memory) - The ratio of space for storing hidden feature maps in pruned model versus original model. ii) \emph{Floating point operations} (xFLOPs) - The ratio of the number of floating-point operations for the original model versus pruned model. iii) \emph{Model size} (Sparsity) - The ratio of the number of zero units in the original model versus the pruned model. We report the clean, adversarial accuracy and vulnerability (Equation~\eqref{eq:network_vulnerability}) for $\ell_{\infty}$ white box and black box attack~\cite{PapernotMGJCS16}. For generating black-box adversarial examples, we used AT and standard network for adversarial training and standard methods respectively. We utilize the clean and adversarial examples for adversarial training methods, all our results are measured by computing mean and standard deviation across 5 trials with random seeds. We list the hyper-parameters in the supplementary material, and the code is available online~\footnote{\url{https://github.com/divyam3897/ANP_VS}}.

\input{white_box_pruning_percentage}
\subsection{Comparison of robustness and generalization}
\paragraph{Evaluation on MNIST.}
For MNIST, we consider a Lenet-5-Caffe model with a perturbation radius of $\varepsilon = 0.3$, perturbation per step of $0.01$, $20$ PGD steps for training, and $40$ PGD steps with random restarts for evaluating the trained model. Our pretrained standard Lenet 5-Caffe baseline model reaches over $99.29\%$ accuracy after 200 epochs averaged across five runs. The results in Table~\ref{table:mainTable} show that ANP-VS significantly improves the adversarial accuracy and vulnerability of the network over all the competing baselines. Namely, with the robust-features, ANP-VS achieves $\sim3\%$ improvement in adversarial accuracy and $\sim65\%$ reduction in the vulnerability against white-box and black-box attacks over the state-of-the-art adversarial training frameworks. In addition, it achieves $\sim 93\%$ reduction in memory footprint with significant speedup over standard adversarial training baselines. The results indicate that ANP-VS better suppresses the vulnerability of the network, which we can directly attribute to increased adversarial robustness.
\input{vulnerability_vis}
\paragraph{Evaluation on CIFAR-10 and CIFAR-100.}
In order to show that ANP-VS is capable of scaling to more complex datasets, we evaluate our method on CIFAR-10 and CIFAR-100 dataset with VGG-16 architecture. We use $\varepsilon = 0.03$, $10$ PGD steps for training and $40$ steps with random restart for evaluation. The results are summarized in Table~\ref{table:mainTable}. We make the following observations from the results: (1) ANP-VS achieves $\sim7\%$ improvement in adversarial accuracy, $\sim58\%$ reduction in vulnerability of the network against white-box and black-box attack over all the baselines for both the datasets. (2) Our proposed method is not only effective on adversarial robustness but also outperforms all the compared adversarial baselines on the standard generalization performance, and strongly support our hypothesis of obtaining robust features. (3) Compared to the standard adversarial training baselines, our method shows a reduction $\sim 85\%$ in the memory-footprint with $2 \times$ fewer FLOPS, demonstrating the effectiveness of our method.

One might also be concerned regarding the number of PGD steps and different $\varepsilon$ values as for certain defenses; the robustness decreased as the number of PGD steps were increased~\cite{engstrom2018evaluating}. Table~\ref{fig:table2} shows the results for different $\ell_{\infty}$ epsilon values and PGD steps up to 1000. Remarkably, in both scenarios, our proposed defense outperforms all the compared baselines, illustrating the practicality of our method. These results indicate that even if the attacker uses greater resources to attack, the effect on our proposed method is negligible. This allows us to conclude that ANP-VS can induce reliable and effective adversarial robustness with much less computation cost.
\subsection{Ablation studies}
\paragraph{Analysis of individual components.}
To further gain insights into the performance of ANP-VS, we perform ablation studies to dissect the effectiveness of various components (ANP and VS) for robustness in Table~\ref{table:ablationResults}. First, we study the impact of VS loss on AT, whether VS produces better robustness, and curtails the vulnerability of the network. One can observe that AT-VS improves the adversarial accuracy~($\sim 3\%$) and leads to~$\sim50\%$ reduction in vulnerability over AT for MNIST and CIFAR-10 dataset. In addition, note that the clean accuracy for AT-VS is approximately the same as AT across all the datasets, which indicates that VS may have a limited effect on standard accuracy, but has a substantial impact on robustness.

Second, we investigate the impact of ANP on AT, whether pruning the vulnerable latent-features is more effective than merely suppressing the vulnerability of those features. Table~\ref{table:ablationResults} shows that ANP not only improves the robustness~($\sim10\%)$ but also the standard generalization~($\sim 3\%$) over AT. Furthermore, we can observe that ANP on AT significantly outperforms VS on AT with CIFAR-10 dataset by $\sim 4\%$ on the adversarial accuracy while significantly reducing the computational cost. This observation supports our conjecture that ANP incorporates the vulnerability of latent-features, and pruning the vulnerable latent-features improves robustness. We know from these results that using VS or ANP can already outperform AT, while incorporating them together in ANP-VS additionally further boosts the performance, as testified. 

\paragraph{Robustness and sparsity.}
Our proposed method could be vital when we want to obtain a lightweight yet robust network. To show that we can achieve both goals at once, we experiment with different scaling coefficient for the KL term in Equation~\eqref{eq:bb_final_loss} to obtain architectures with varying degrees of sparsity, whose details can be found in the supplementary material.

Figure \ref{fig:final_results} clearly shows that ANP-VS outperforms AT up to a sparsity level of $\sim 80\%$ for CIFAR-10 and CIFAR-100 after which there is a decrease in the robust and standard generalization. The results are not surprising, as it is an overall outcome of the model capacity reduction and the removal of the robust features. We further analyze a baseline PAT where we first perform Bayesian pruning, freeze the dropout mask, and then perform AT (see Figure \ref{fig:final_results}). We can observe that PAT marginally improves the robustness over AT but loses on clean accuracy. Note that ANP-VS significantly outperforms PAT, supporting our hypothesis that just naive approach of AT over pruning can hurt performance.

\subsection{Further analysis on defense performance}
\paragraph{Vulnerability analysis.}
We conduct additional experiments to visualize the vulnerability of the latent-feature space to investigate the robust and vulnerable features. Figure~\ref{fig:vulnerability_fig} (Top) compares the vulnerability of various models. One can observe that the latent-features of the standard model are the most vulnerable, and the vulnerability decreases with the AT and is further suppressed by half with ANP-VS. Further, note that the latent features of our proposed method capture more local information of the objects and align much better with human perception. 

In addition, Figure~\ref{fig:vulnerability_fig} (Bottom) shows the histogram of the vulnerability of latent-features for input-layer defined in Equation~\eqref{eq:feature_vulnerability} for various methods. We consistently see that standard Bayesian pruning zeros out some of the distortions in the latent-features, and AT reduces the distortion level of all the latent-features. On the other hand, ANP-VS does both, with the largest number of latent-features with zero distortion and low distortion level in general. In this vein, we can say that ANP-VS demonstrates the effectiveness of our method as a defense mechanism by obtaining robust features utilizing pruning and latent vulnerability suppression. 

\paragraph{Loss landscape visualization.} It has been observed that adversarial training flatters the adversarial loss landscape and obfuscated gradients can often lead to bumpier loss landscapes~\cite{engstrom2018evaluating}. To investigate this problem of obfuscated gradients, we visualize the adversarial loss landscape of the baseline models and ANP-VS in Figure~\ref{fig:loss_surface}.  We vary the input along a linear space defined by the sign of gradient where x and y-axes represent the perturbation added in each direction, and the z-axis represents the loss.

We can observe that the loss is highly curved in the vicinity of the data point x for the standard networks, which reflects that the gradient poorly models the global landscape. On the other hand, we observe that both sparsity and adversarial training make the loss surface smooth, with our model obtaining the smoothest surface. It demonstrates that our proposed method is not susceptible to the problem of gradients obfuscation. In addition, it indicates that suppressing the network vulnerability leads to robust optimization, uncovering some unexpected benefits of our proposed method over adversarial training. 

%% file: table_1.tex
\begin{table*}[t!]
\centering
	\rtable{1.2}
			\caption{Robustness and compression performance for MNIST on Lenet-5-Caffe, CIFAR-10 and CIFAR-100 on VGG-16 architecture under $\ell_{\infty}$-PGD attack. All the values are measured by computing mean and standard deviation across 5 trials upon randomly chosen seeds. The best results over adversarial baselines are highlighted in bold. $\uparrow$ ($\downarrow$) indicates that the higher (lower) number is the better. \label{table:mainTable}}
			\resizebox{\linewidth}{!}{
			\begin{tabular}{lllllllllllllll}
				\toprule
				{} & {} & {} & \multicolumn{2}{c}{Adversarial accuracy $(\uparrow)$} & \multicolumn{2}{c}{Vulnerability ($\downarrow$)} & \multicolumn{3}{c}{Computational efficiency}\\
				\cmidrule(lr){4-5}\cmidrule(lr){6-7}\cmidrule(lr){8-10}
				& Model & \vtop{\hbox{\strut Clean}\hbox{\strut accuracy ($\uparrow$)}} & \vtop{\hbox{\strut White box}\hbox{\strut attack}} & \vtop{\hbox{\strut Black box}\hbox{\strut attack}} & \vtop{\hbox{\strut White box}\hbox{\strut attack}} & \vtop{\hbox{\strut Black box }\hbox{\strut attack}} & Memory ($\downarrow$) & xFLOPS ($\uparrow$) & Sparsity ($\uparrow$)\\
				\midrule
				\parbox[t]{2mm}{\multirow{9}{*}{\rotatebox[origin=c]{90}{MNIST}}}
				& Standard & 99.29{\scriptsize $\pm$0.02} & 0.00{\scriptsize $\pm$0.0} & 8.02{\scriptsize $\pm$0.9} & 0.129{\scriptsize $\pm$0.001} & 0.113{\scriptsize $\pm$0.000} & 100.0{\scriptsize $\pm$0.00} & 1.00{\scriptsize $\pm$0.00} &  0.00{\scriptsize $\pm$0.00}\\
				& BP & 99.34{\scriptsize $\pm$0.05} & 0.00{\scriptsize $\pm$0.0} & 12.99{\scriptsize $\pm$0.5} & 0.091{\scriptsize $\pm$0.001} & 0.078{\scriptsize $\pm$0.001} & 4.14{\scriptsize $\pm$0.29} & 9.68{\scriptsize $\pm$0.36} & 83.48{\scriptsize $\pm$0.54}\\
				\cline{2-10}
				& AT& 99.14{\scriptsize $\pm$0.02} & 88.03{\scriptsize $\pm$0.7}  & 94.18{\scriptsize $\pm$0.8} & 0.045{\scriptsize $\pm$0.001} & 0.040{\scriptsize $\pm$0.000} & 100.0{\scriptsize $\pm$0.00} & 1.00{\scriptsize $\pm$0.00} &  0.00{\scriptsize $\pm$0.00}\\
				& AT BNN & 99.16{\scriptsize $\pm$0.05} & 88.44{\scriptsize $\pm$0.4} & 94.87{\scriptsize $\pm$0.2} & 0.364{\scriptsize $\pm$0.023} & 0.199{\scriptsize $\pm$0.031}  & 200.0{\scriptsize $\pm$0.00} & 0.50{\scriptsize $\pm$0.00} & 0.00{\scriptsize $\pm$0.00} \\
				& Pretrained AT  & \textbf{99.18{\scriptsize $\pm$0.06}} & 88.26{\scriptsize $\pm$0.6} & 94.49{\scriptsize $\pm$0.7} & 0.412{\scriptsize $\pm$0.035} & 0.381{\scriptsize $\pm$0.029} & 100.0{\scriptsize $\pm$0.00} & 1.00{\scriptsize $\pm$0.00} &  0.00{\scriptsize $\pm$0.00}\\
				& ADMM  & 99.01{\scriptsize $\pm$0.02} & 88.47{\scriptsize $\pm$0.4} & 94.61{\scriptsize $\pm$0.7} & 0.041{\scriptsize $\pm$0.002} & 0.038{\scriptsize $\pm$0.001} & 100.0{\scriptsize $\pm$0.00} & 1.00{\scriptsize $\pm$0.00} &  80.00{\scriptsize $\pm$0.00}\\
				& TRADES  & 99.07{\scriptsize $\pm$0.04} & 89.67{\scriptsize $\pm$0.4} & 95.04{\scriptsize $\pm$0.6} & 0.037{\scriptsize $\pm$0.001} & 0.033{\scriptsize $\pm$0.001} & 100.0{\scriptsize $\pm$0.00} & 1.00{\scriptsize $\pm$0.00} &  0.00{\scriptsize $\pm$0.00}\\
				\cline{2-10}
				& ANP-VS (ours) & 99.05{\scriptsize $\pm$0.08} & \textbf{91.31{\scriptsize $\pm$0.9}} & \textbf{95.43{\scriptsize $\pm$0.8}} & \textbf{0.017{\scriptsize $\pm$0.001}} & \textbf{0.015{\scriptsize $\pm$0.001}} & \textbf{6.81{\scriptsize $\pm$0.35}} & \textbf{10.57{\scriptsize $\pm$1.15}} & \textbf{84.16{\scriptsize $\pm$0.36}}\\
				\hline  \TBstrut
				
				\parbox[t]{2mm}{\multirow{9}{*}{\rotatebox[origin=c]{90}{CIFAR-10}}}
				& Standard & 92.76{\scriptsize $\pm$0.1} & 13.79{\scriptsize $\pm$0.8} & 41.65{\scriptsize $\pm$0.9} & 0.077{\scriptsize $\pm$0.001} & 0.065{\scriptsize $\pm$0.001} & 100.0{\scriptsize $\pm$0.00} &  1.00{\scriptsize $\pm$0.00}  & 0.00{\scriptsize $\pm$0.00} \\
				& BP & 92.91{\scriptsize $\pm$0.1} & 14.30{\scriptsize $\pm$0.5} & 42.88{\scriptsize $\pm$1.3} & 0.037{\scriptsize $\pm$0.001} & 0.033{\scriptsize $\pm$0.001} & 12.41{\scriptsize $\pm$0.14} & 2.34{\scriptsize $\pm$0.0.03} & 75.92{\scriptsize $\pm$0.13}\\
				\cline{2-10}
				& AT & 87.50{\scriptsize $\pm$0.5} & 49.85{\scriptsize $\pm$0.9} & 63.70{\scriptsize $\pm$0.6} & 0.050{\scriptsize $\pm$0.002} & 0.047{\scriptsize $\pm$0.001} & 100.0{\scriptsize $\pm$0.00} &  1.00{\scriptsize $\pm$0.00} & 0.00{\scriptsize $\pm$0.00} \\
				& AT BNN & 86.69{\scriptsize $\pm$0.5} & 51.87{\scriptsize $\pm$0.9} & 64.92{\scriptsize $\pm$0.9} & 0.267{\scriptsize $\pm$0.013} & 0.238{\scriptsize $\pm$0.011} & 200.0{\scriptsize $\pm$0.00} & 0.50{\scriptsize $\pm$0.00} & 0.00{\scriptsize $\pm$0.00} \\
				& Pretrained AT & 87.50{\scriptsize $\pm$0.4} &  52.25{\scriptsize $\pm$0.7} & 66.10{\scriptsize $\pm$0.8} & 0.041{\scriptsize $\pm$0.002} & 0.036{\scriptsize $\pm$0.001} & 100.0{\scriptsize $\pm$0.00} &  1.00{\scriptsize $\pm$0.00}  & 0.00{\scriptsize $\pm$0.00} \\
				& ADMM  & 78.15{\scriptsize $\pm$0.7} & 47.37{\scriptsize $\pm$0.6} & 62.15{\scriptsize $\pm$0.8} & 0.034{\scriptsize $\pm$0.002} & 0.030{\scriptsize $\pm$0.002} & 100.0{\scriptsize $\pm$0.00} & 1.00{\scriptsize $\pm$0.00} & 75.00{\scriptsize $\pm$0.00}\\
				& TRADES & 80.33{\scriptsize $\pm$0.5} & 52.08{\scriptsize $\pm$0.7} & 64.80{\scriptsize $\pm$0.5} & 0.045{\scriptsize $\pm$0.001} & 0.042{\scriptsize $\pm$0.005} & 100.0{\scriptsize $\pm$0.00} &  1.00{\scriptsize $\pm$0.00}  & 0.00{\scriptsize $\pm$0.00} \\
				\cline{2-10}
				& ANP-VS (ours) & \textbf{88.18{\scriptsize $\pm$0.5}} &  \textbf{56.21{\scriptsize $\pm$0.1}} & \textbf{71.44{\scriptsize $\pm$0.6}} & \textbf{0.019{\scriptsize $\pm$0.000}} & \textbf{0.016{\scriptsize $\pm$0.000}} & \textbf{12.27{\scriptsize $\pm$0.18}} & \textbf{2.41{\scriptsize $\pm$0.04}} &  \textbf{76.53{\scriptsize $\pm$0.16}}\\
				\hline  \TBstrut
				
				\parbox[t]{2mm}{\multirow{9}{*}{\rotatebox[origin=c]{90}{CIFAR-100}}}
				& Standard & 67.44{\scriptsize $\pm$0.7} & 2.81{\scriptsize $\pm$0.2} & 14.94{\scriptsize $\pm$0.8} & 0.143{\scriptsize $\pm$0.007} & 0.119{\scriptsize $\pm$0.005} & 100.0{\scriptsize $\pm$0.00} &  1.00{\scriptsize $\pm$0.00}  & 0.00{\scriptsize $\pm$0.00} \\
				& BP & 69.40{\scriptsize $\pm$0.7} & 3.12{\scriptsize $\pm$0.1} & 16.39{\scriptsize $\pm$0.2} & 0.067{\scriptsize $\pm$0.001} & 0.059{\scriptsize $\pm$0.001} & 18.59{\scriptsize $\pm$0.56} & 1.95{\scriptsize $\pm$0.04} & 63.48{\scriptsize $\pm$0.88}\\
				\cline{2-10}
				& AT & 57.79{\scriptsize $\pm$0.8} & 19.07{\scriptsize $\pm$0.8} & 32.47{\scriptsize $\pm$1.4} & 0.079{\scriptsize $\pm$0.003} & 0.071{\scriptsize $\pm$0.003} & 100.0{\scriptsize $\pm$0.00} &  1.00{\scriptsize $\pm$0.00}  & 0.00{\scriptsize $\pm$0.00} \\
				& AT BNN & 53.75{\scriptsize $\pm$0.7} & 19.40{\scriptsize $\pm$0.6} & 30.38{\scriptsize $\pm$0.2} & 0.446{\scriptsize $\pm$0.029} & 0.385{\scriptsize $\pm$0.051} & 200.0{\scriptsize $\pm$0.00} & 0.50{\scriptsize $\pm$0.00} & 0.00{\scriptsize $\pm$0.00} \\
				& Pretrained AT & 57.14{\scriptsize $\pm$0.9} & 19.86{\scriptsize $\pm$0.6} & 35.42{\scriptsize $\pm$0.4} & 0.071{\scriptsize $\pm$0.001} & 0.065{\scriptsize $\pm$0.002} & 100.0{\scriptsize $\pm$0.00} &  1.00{\scriptsize $\pm$0.00}  & 0.00{\scriptsize $\pm$0.00} \\
				& ADMM  & 52.52{\scriptsize $\pm$0.5} & 19.65{\scriptsize $\pm$0.5} & 31.30{\scriptsize $\pm$0.3} & 0.060{\scriptsize $\pm$0.001} & 0.056{\scriptsize $\pm$0.001} &  100.0{\scriptsize $\pm$0.00} & 1.00{\scriptsize $\pm$0.00} &  65.00{\scriptsize $\pm$0.00}\\
				& TRADES & 56.70{\scriptsize $\pm$0.7} & 21.21{\scriptsize $\pm$0.3} & 32.81{\scriptsize $\pm$0.6} & 0.065{\scriptsize $\pm$0.003} & 0.060{\scriptsize $\pm$0.003} & 100.0{\scriptsize $\pm$0.00} &  1.00{\scriptsize $\pm$0.00} & 0.00{\scriptsize $\pm$0.00} \\
				\cline{2-10}
				& ANP-VS (ours) & \textbf{59.15{\scriptsize $\pm$1.2}} & \textbf{22.35{\scriptsize $\pm$0.6}} & \textbf{37.01{\scriptsize $\pm$1.1}} & \textbf{0.035{\scriptsize $\pm$0.001}} & \textbf{0.030{\scriptsize $\pm$0.003}} & \textbf{16.74{\scriptsize $\pm$0.52}} & \textbf{2.02{\scriptsize $\pm$0.05}} &  \textbf{66.80{\scriptsize $\pm$0.75}} \\
				\bottomrule
			\end{tabular}}
	\end{table*}

%% file: table_2.tex
	\begin{table*}[t!]
	\centering
			\caption{Adversarial accuracy of CIFAR-10 and CIFAR-100 for VGG-16 architecture under $\ell_{\infty}$-PGD white box attack for various $\varepsilon$ values and PGD iterations with the perturbation per step of 0.007. The best results are highlighted in bold. \label{fig:table2}}
	\resizebox{\linewidth}{!}{
\begin{tabular}{llllll||llll}
		\toprule
		\centering
		{} & {} & \multicolumn{4}{c||}{Epsilon ($\varepsilon$)} & \multicolumn{4}{c}{\# Iterations}\\
		\cmidrule(lr){3-6}\cmidrule(lr){7-10}
		&  & 0.01 & 0.03 & 0.05 & 0.07 & 100 & 200 & 500 & 1000\\
		\midrule
		
		\parbox[t]{2mm}{\multirow{8}{*}{\rotatebox[origin=c]{90}{CIFAR-10}}}
        & Standard & 40.54{\scriptsize $\pm$0.9} & 13.79{\scriptsize $\pm$0.8} & 4.16{\scriptsize $\pm$0.5} & 1.46{\scriptsize $\pm$0.1} & 11.76{\scriptsize $\pm$0.4}  & 11.73{\scriptsize $\pm$0.7} & 13.71{\scriptsize $\pm$0.6} & 13.68{\scriptsize $\pm$0.7}\\
        & BP & 41.67{\scriptsize $\pm$0.6} & 14.30{\scriptsize $\pm$0.5} & 4.21{\scriptsize $\pm$0.4} & 1.50{\scriptsize $\pm$0.1} & 14.24{\scriptsize $\pm$0.5} & 14.18{\scriptsize $\pm$0.3} & 14.14{\scriptsize $\pm$0.2} & 14.16{\scriptsize $\pm$0.3}\\
        \cline{2-10}
		& AT & 65.86{\scriptsize $\pm$0.8} & 49.85{\scriptsize $\pm$0.9} & 34.54{\scriptsize $\pm$0.8} & 22.69{\scriptsize $\pm$0.7} & 49.74{\scriptsize $\pm$0.9} & 49.75{\scriptsize $\pm$0.8} & 49.76{\scriptsize $\pm$0.9} & 49.75{\scriptsize $\pm$0.9} \\
		& AT BNN & 67.02{\scriptsize $\pm$0.6} & 51.87{\scriptsize $\pm$0.9} & 36.76{\scriptsize $\pm$0.9} & 25.06{\scriptsize $\pm$0.7} & 51.81{\scriptsize $\pm$0.9} & 51.80{\scriptsize $\pm$0.9} & 51.81{\scriptsize $\pm$0.9} & 51.80{\scriptsize $\pm$0.9} \\
		& Pretrained AT & 70.55{\scriptsize $\pm$0.7} & 52.25{\scriptsize $\pm$0.7} & 37.11{\scriptsize $\pm$0.8} & 24.03{\scriptsize $\pm$0.7} & 52.19{\scriptsize $\pm$0.7} & 52.18{\scriptsize $\pm$0.7} & 52.16{\scriptsize $\pm$0.7} & 52.13{\scriptsize $\pm$0.7} \\
	    & ADMM & 65.94{\scriptsize $\pm$0.4} & 47.37{\scriptsize $\pm$0.6} & 32.77{\scriptsize $\pm$0.6} & 22.45{\scriptsize $\pm$0.5} & 47.34{\scriptsize $\pm$0.5} & 47.31{\scriptsize $\pm$0.5} & 47.30{\scriptsize $\pm$0.4} & 47.31{\scriptsize $\pm$0.5} \\
		& TRADES & 69.54{\scriptsize $\pm$0.2} & 52.08{\scriptsize $\pm$0.7} & 34.54{\scriptsize $\pm$0.3} & 23.01{\scriptsize $\pm$0.4} & 52.03{\scriptsize $\pm$0.5} & 52.01{\scriptsize $\pm$0.6} & 52.03{\scriptsize $\pm$0.5} & 52.02{\scriptsize $\pm$0.5} \\
		\cline{2-10}
		& ANP-VS (ours) & \textbf{72.33{\scriptsize $\pm$0.4}} & \textbf{56.21{\scriptsize $\pm$0.1}} & \textbf{40.89{\scriptsize $\pm$0.3}} & \textbf{26.90{\scriptsize $\pm$0.2}}& \textbf{56.17{\scriptsize $\pm$0.1}} & \textbf{56.16{\scriptsize $\pm$0.1}} & \textbf{56.15{\scriptsize $\pm$0.1}} & \textbf{56.16{\scriptsize $\pm$0.1}} \\
		\hline  \TBstrut
		
		\parbox[t]{2mm}{\multirow{8}{*}{\rotatebox[origin=c]{90}{CIFAR-100}}}
		& Standard & 11.68{\scriptsize $\pm$0.7} & 2.81{\scriptsize $\pm$0.2} & 0.91{\scriptsize $\pm$0.1} & 0.41{\scriptsize $\pm$0.0} & 2.77{\scriptsize $\pm$0.1} & 2.76{\scriptsize $\pm$0.2} & 2.76{\scriptsize $\pm$0.1} & 2.75{\scriptsize $\pm$0.1}\\
		& BP & 13.01{\scriptsize $\pm$0.3} & 3.12{\scriptsize $\pm$0.1} & 1.04{\scriptsize $\pm$0.1} & 0.47{\scriptsize $\pm$0.0} & 3.02{\scriptsize $\pm$0.3} & 3.06{\scriptsize $\pm$0.2} & 3.03{\scriptsize $\pm$0.2} & 3.04{\scriptsize $\pm$0.1}\\
		\cline{2-10}
		& AT & 35.10{\scriptsize $\pm$0.9} & 19.07{\scriptsize $\pm$0.8} & 10.62{\scriptsize $\pm$0.4} & 5.89{\scriptsize $\pm$0.8} & 19.06{\scriptsize $\pm$0.8} & 19.03{\scriptsize $\pm$0.8} & 19.02{\scriptsize $\pm$0.8} & 19.04{\scriptsize $\pm$0.7}\\
		& AT BNN & 30.13{\scriptsize $\pm$0.9} & 19.40{\scriptsize $\pm$0.6} & 10.79{\scriptsize $\pm$0.5} & 5.99{\scriptsize $\pm$0.2} & 19.34{\scriptsize $\pm$0.9} & 19.37{\scriptsize $\pm$0.9} & 19.36{\scriptsize $\pm$0.9} & 19.35{\scriptsize $\pm$0.8} \\
		& Pretrained AT & 33.92{\scriptsize $\pm$0.2} & 19.86{\scriptsize $\pm$0.6} & 11.39{\scriptsize $\pm$0.3} & 6.27{\scriptsize $\pm$0.1} & 19.85{\scriptsize $\pm$0.5} & 19.84{\scriptsize $\pm$0.6} & 19.81{\scriptsize $\pm$0.6} & 19.83{\scriptsize $\pm$0.6} \\
		& ADMM & 34.59{\scriptsize $\pm$0.2} & 19.65{\scriptsize $\pm$0.5} & 10.50{\scriptsize $\pm$0.3} & 4.77{\scriptsize $\pm$0.4} & 19.60{\scriptsize $\pm$0.3} & 19.57{\scriptsize $\pm$0.3} & 19.59{\scriptsize $\pm$0.3} & 19.59{\scriptsize $\pm$0.2} \\
		& TRADES & 33.89{\scriptsize $\pm$0.1} & 21.22{\scriptsize $\pm$0.3} & 10.80{\scriptsize $\pm$0.1} & 4.51{\scriptsize $\pm$0.2} & 21.15{\scriptsize $\pm$0.4} & 21.16{\scriptsize $\pm$0.3} & 21.20{\scriptsize $\pm$0.4} & 21.15{\scriptsize $\pm$0.4} \\
		\cline{2-10}
		& ANP-VS (ours) & \textbf{35.70{\scriptsize $\pm$0.8}} & \textbf{22.35{\scriptsize $\pm$0.6}} & \textbf{12.95{\scriptsize $\pm$0.6}} & \textbf{7.28{\scriptsize $\pm$0.3}} & \textbf{22.32{\scriptsize $\pm$0.7}} & \textbf{22.26{\scriptsize $\pm$0.7}} & \textbf{22.25{\scriptsize $\pm$0.6}} & \textbf{22.26{\scriptsize $\pm$0.7}}\\
		\bottomrule
	\end{tabular}}
		\end{table*}

%% file: white_box_pruning_percentage.tex
\begin{table*}[t]
	\begin{minipage}[b]{0.50\linewidth}
\subfigure{
		\hspace{4.6em}\resizebox{0.50\linewidth}{!}{%
			\begin{tikzpicture}
			\begin{customlegend}[legend columns=5,legend style={column sep=.1ex},
			legend entries={\textsc{\footnotesize  Original},\textsc{\footnotesize  BP},\textsc{\footnotesize  AT},\textsc{\footnotesize  PAT},\textsc{\footnotesize ANP-VS}
			}]
			\addlegendimage{mark=square,solid,tomato,line width=2.4pt} 
				\addlegendimage{mark=square,solid,azure,line width=2.4pt} 
 	
			\addlegendimage{mark=square,solid,uclagold,line width=2.4pt}
			\addlegendimage{mark=square,solid,green(munsell),line width=2.4pt} 
			\addlegendimage{mark=square,solid,line legend,tractorred,line width=2.4pt}
			\end{customlegend}
		\end{tikzpicture}}}

		\subfigure{
			\resizebox{0.31\linewidth}{!}{
				\begin{tikzpicture}
				\begin{axis}[
				title={Clean Accuracy},
				xlabel={Sparsity (\%)},
				ylabel={Clean Accuracy (\%)},
				ylabel near ticks,
				xlabel near ticks,
				xmin=0, xmax=35,
				ymin=0, ymax=25,
				enlarge x limits=true,
				xtick={0,5,10,15,20,25,30,35},
				xticklabels={0,65,70,75,80,85,90,95},
				ytick={0, 5, 10,15,20,25},
				yticklabels={0,75,80,85,90,95},
				legend pos=south east,
				ymajorgrids=true,
				grid style=dashed,
				tickwidth=0.1cm,
				max space between ticks=250,
				grid=both,
				font=\LARGE,
				style={ultra thick}
				]
				\addplot[
				color=tractorred,
				mark=square,
				line width=2.6pt
				]
				coordinates {
					(9.4,18.63)(14.39,18.225)(16.345,18.74)(21.47,17.415)(25.595,15.112)(29.13,11.13)(30.75,7.4)
				};
				\addplot[
				color=uclagold,
				mark=square,
				mark size=2pt,
				line width=2.6pt
				]
				coordinates {
					(0,17.37)
				};
				\addplot[
				color=green(munsell),
				mark=square,
				line width=2.6pt
				]
				coordinates {
					(15.74,12.24)(20.75,10.72)(24.32,10.22)(27,9.36)
				};
				\addplot[
				color=tomato,
				mark=square,
				mark size=2pt,
				line width=2.6pt
				]
				coordinates {
					(0,12.65)
				};
				\addplot[
				color=azure,
				mark=square,
				line width=2.6pt
				]
				coordinates {
					(15.825,23.0025)(20.735,22.365)(24.4325, 21.1325)(26.785,19.724)(28.74,18.19)
				};
				\end{axis}
				\end{tikzpicture}}}
		\subfigure{
			\resizebox{0.31\linewidth}{!}{%
				\begin{tikzpicture}
				\begin{axis}[
				title={White Box Attack},
				xlabel={Sparsity (\%)},
				ylabel={Adversarial Accuracy(\%)},
				ylabel near ticks,
				xlabel near ticks,
				xmin=0, xmax=35,
				ymin=0, ymax=60,
				enlarge x limits=true,
				xtick={0,5,10,15,20,25,30,35,40},
				xticklabels={0,65,70,75,80,85,90,95},
				ytick={0, 10, 20, 30,40,50,60,70,80,90,100},
				grid=both,
				ymajorgrids=true,
				grid style=dashed,
				font=\LARGE,
				style={ultra thick},
				]
				\addplot[
				color=tractorred,
				mark=square,
				line width=2.6pt
				]
				coordinates {
					(9.4,53.97)(14.39,53.90)	(16.345,54.9445)(21.47,54.629)(25.595,52.85333333)(29.13,48.445)(30.75,43.765)
				};
				\addplot[
				color=green(munsell),
				mark=square,
				line width=2.6pt
				]
				coordinates {
					(15.74,51.73)(20.75,49.42)(24.32,50.39)(27,48.27)
				};
				\addplot[
				color=uclagold,
				mark=square,
				mark size=2pt,
				line width=2.6pt
				]
				coordinates {
					(0,48.93)
				};
				\addplot[
				color=tomato,
				mark=square,
				mark size=2pt,
				line width=2.6pt
				]
				coordinates {
					(0,12.80975)
				};
				\addplot[
				color=azure,
				mark=square,
				line width=2.6pt
				]
				coordinates {
					(15.825,13.3685)(20.735,14.91)(24.4325, 15.3225)(26.785,15.65725)(28.74,16.007)
				};
				\end{axis}
				\end{tikzpicture}}}
		\subfigure
		{
			\resizebox{0.31\linewidth}{!}{%
				\begin{tikzpicture}
				\begin{axis}[
				title={Black Box Attack},
				xlabel={Sparsity (\%)},
				ylabel={Adversarial Accuracy(\%)},
				ylabel near ticks,
				xlabel near ticks,
				xmin=0, xmax=35,
				ymin=0, ymax=50,
				enlarge x limits=true,
				xtick={0,5,10,15,20,25,30,35},
				xticklabels={0,65,70,75,80,85,90,95},
				ytick={0, 10, 20, 30,40,50},
				yticklabels={0,40,50,60,70,80},
				ymajorgrids=true,
				grid style=dashed,
				font=\LARGE,
				grid=both,
				style={ultra thick}
				]
				\addplot[
				color=tractorred,
				mark=square,
				line width=2.6pt
				]
				coordinates {
					(9.4,39.13)(14.39,38.84)(16.345,40.2465)(21.47,39.8015)(25.595,37.72)(29.13,33.61)(30.75,31.792)
				};
				\addplot[
				color=green(munsell),
				mark=square,
				line width=2.6pt
				]
				coordinates {
					(15.74,36.85)(20.75,35.24)(24.32,35.95)(27,33.44)
				};
				\addplot[
				color=uclagold,
				mark=square,
				mark size=2pt,
				line width=2.6pt
				]
				coordinates {
					(0,33.61)
				};
				\addplot[
				color=tomato,
				mark=square,
				mark size=2pt,
				line width=2.6pt
				]
				coordinates {
					(0,10.8925)
				};
				\addplot[
				color=azure,
				mark=square,
				line width=2.6pt
				]
				coordinates {
					(15.825,12.11)(20.735,13.3075)(24.4325, 13.49)(26.785,12.88)(28.74,13.8475)
				};
				\end{axis}
				\end{tikzpicture}}}
		
		\subfigure{
			\resizebox{0.31\linewidth}{!}{%
				\begin{tikzpicture}
				\begin{axis}[
				title={Clean Accuracy},
				xlabel={Sparsity (\%)},
				ylabel={Clean Accuracy (\%)},
				ylabel near ticks,
				xlabel near ticks,
				enlarge x limits=true,
				xmin=0, xmax=40,
				ymin=0, ymax=50,
				xtick={0,5,10,15,20,25,30,35,40},
				xticklabels={0,55,60,65,70,75,80,85,90},
				yticklabels={0,40,50,60,70,80},
				ytick={0, 10, 20, 30, 40,50},
				legend pos=south west,
				ymajorgrids=true,
				grid style=dashed,
				grid=both,
				font=\LARGE,
				style={ultra thick}
				]
				\addplot[
				color=tractorred,
				mark=square,
				line width=2.6pt
				]
				coordinates {(9.265,25)(13.93,26.29)(16.77,29.265)(20.005,27.76)(25.505,23.2)(31.525,19.15)(36.5,10.095)
				};
				\addplot[
				color=green(munsell),
				mark=square,
				line width=2.6pt
				]
				coordinates {
					(13.64,21.76)(19.86,20.35)(24.73,21.57)(28.37,20)
				};
				\addplot[
				color=uclagold,
				mark=square,
				mark size=2pt,
				line width=2.6pt
				]
				coordinates {
					(0,26.43)
				};
				\addplot[
				color=tomato,
				mark=square,
				mark size=2pt,
				line width=2.6pt
				]
				coordinates {
					(0,37.1025)
				};
				\addplot[
				color=azure,
				mark=square,
				line width=2.6pt
				]
				coordinates {
					(13.095,38.945)(19.5275,37.36)(23.5725,35.275)(26.95666667,31.96333333)(31.28,26.98)
				};
				\end{axis}
				\end{tikzpicture}}}
		\subfigure{
			\resizebox{0.31\linewidth}{!}{%
				\begin{tikzpicture}
				\begin{axis}[
				title={White Box Attack},
				xlabel={Sparsity (\%)},
				ylabel={Adversarial Accuracy(\%)},
				ylabel near ticks,
				xlabel near ticks,
				enlarge x limits=true,
				xmin=0, xmax=40,
				ymin=0, ymax=25,
				xtick={0,5,10,15,20,25,30,35,40},
				xticklabels={0,55,60,65,70,75,80,85,90},
				ytick={0, 5, 10, 15, 20, 25},
				grid=both,
				ymajorgrids=true,
				font=\LARGE,
				grid style=dashed,
				style={ultra thick}
				]
				\addplot[
				color=tractorred,
				mark=square,
				line width=2.6pt
				]
				coordinates {
					(9.265,21.5)(13.93,21.16)	(16.77,23.328)(20.005,22.0725)(25.505,20.605)(31.525,17.8445)(36.5,13.03)
				};
				\addplot[
				color=uclagold,
				mark=square,
				mark size=2pt,
				line width=2.6pt
				]
				coordinates {
					(0,19.9155)
				};
				\addplot[
				color=green(munsell),
				mark=square,
				line width=2.6pt
				]
				coordinates {
					(13.64,19.7)(19.86,21)(24.73,19.6)(28.3,19.0)
				};
				\addplot[
				color=tomato,
				mark=square,
				mark size=2pt,
				line width=2.6pt
				]
				coordinates {
					(0,2.9575)
				};
				\addplot[
				color=azure,
				mark=square,
				line width=2.6pt
				]
				coordinates {
					(13.095,3.33225)(19.5275,3.54)(23.5725,3.376666667)(26.95666667,3.361666667)(31.28,3.015)
				};
				\end{axis}
				\end{tikzpicture}}}
		\subfigure
		{
			\resizebox{0.31\linewidth}{!}{%
				\begin{tikzpicture}
				\begin{axis}[
				title={Black Box Attack},
				xlabel={Sparsity (\%)},
				ylabel={Adversarial Accuracy(\%)},
				ylabel near ticks,
				xlabel near ticks,
				enlarge x limits=true,
				xmin=0, xmax=40,
				ymin=0, ymax=40,
				xtick={0,5,10,15,20,25,30,35,40},
				xticklabels={0,55,60,65,70,75,80,85,90},
				ymajorgrids=true,
				grid style=dashed,
				font=\LARGE,
				grid=both,
				style={ultra thick}
				]
				\addplot[
				color=tractorred,
				mark=square,
				line width=2.6pt
				]
				coordinates {
					(9.265,36.86)(13.93,36.11)	(16.77,38.075)(20.005,36.27)(25.505,33.735)(31.525,29.7485)(36.5,22.81)
				};
				\addplot[
				color=uclagold,
				mark=square,
				mark size=2pt,
				line width=2.6pt
				]
				coordinates {
					(0,33.39)
				};

				\addplot[
				color=tomato,
				mark=square,
				mark size=2pt,
				line width=2.6pt
				]
				coordinates {
					(0,15.9)
				};
				\addplot[
				color=azure,
				mark=square,
				line width=2.6pt
				]
				coordinates {
					(13.095,16.68375)(19.5275,17.07)(23.5725,16.59333333)(26.95666667,16.02)(31.28,15.13666667)
				};
				\addplot[
				color=green(munsell),
				mark=square,
				line width=2.6pt
				]
				coordinates {
					(13.64,32.99)(19.86,34.68)(24.73,32.23)(28.3,31.24)
				};
				\end{axis}
				\end{tikzpicture}}}
		\par\vspace{0pt}
		\captionof{figure}{Comparison of clean and adversarial accuracy for different sparsity levels. {Top:} VGG-16 on CIFAR10. {Bottom:} VGG-16 on CIFAR100 dataset.\label{fig:final_results}}
	\end{minipage}\hfill
	\noindent
	\begin{minipage}[b]{0.47\linewidth}
			\caption{Ablation studies for Vulnerability Suppression loss (VS) and Adversarial Neural Pruning (ANP) to investigate the impact of VS and ANP against $\ell_\infty$ white-box PGD attack. $\uparrow$ ($\downarrow$) indicates that the higher (lower) number is the better. The best results are highlighted in bold. \label{table:ablationResults}}

		\resizebox{\linewidth}{!}{
			\begin{tabular}{llllllll}
				\toprule
				& Model & Clean Acc. ($\uparrow$) & Adv. acc ($\uparrow$) & Vulnerability ($\downarrow$)\\
				\midrule
				\hline  \TBstrut
				\parbox[t]{2mm}{\multirow{4}{*}{\rotatebox[origin=c]{90}{MNIST}}}
				& AT &  99.14{\scriptsize $\pm$0.02} & 88.03{\scriptsize $\pm$0.7} &  0.045{\scriptsize $\pm$0.001} \\
				& AT-VS & \textbf{99.24{\scriptsize $\pm$0.05}} & 90.36{\scriptsize $\pm$0.8} &  0.022{\scriptsize $\pm$0.001}\\
				& ANP & 98.64{\scriptsize $\pm$0.07} & 90.12{\scriptsize $\pm$0.5} & 0.020{\scriptsize $\pm$0.002}\\
				& ANP-VS & 99.05{\scriptsize $\pm$0.08} & \textbf{91.31{\scriptsize $\pm$0.9}} & \textbf{0.011{\scriptsize $\pm$0.002}} \\
				\hline  \TBstrut
				\parbox[t]{2mm}{\multirow{4}{*}{\rotatebox[origin=c]{90}{CIFAR-10}}}
				& AT &  87.50{\scriptsize $\pm$0.5} & 49.85{\scriptsize $\pm$0.9}  &  0.050{\scriptsize $\pm$0.002}\\
				& AT-VS &  87.44{\scriptsize $\pm$0.5} & 51.52{\scriptsize $\pm$0.4} &  0.024{\scriptsize $\pm$0.002}\\
				& ANP & \textbf{88.36{\scriptsize $\pm$0.4}} & 55.63{\scriptsize $\pm$0.9} & 0.022{\scriptsize $\pm$0.001}\\
				& ANP-VS & 88.18{\scriptsize $\pm$0.5} &  \textbf{56.21{\scriptsize $\pm$0.1}} & \textbf{0.016{\scriptsize $\pm$0.000}}\\
				\hline  \TBstrut
				
				\parbox[t]{2mm}{\multirow{4}{*}{\rotatebox[origin=c]{90}{CIFAR-100}}}
				& AT &  57.79{\scriptsize $\pm$0.8} & 19.07{\scriptsize $\pm$0.8}  &  0.079{\scriptsize $\pm$0.003}\\
				& AT-VS & 57.74{\scriptsize $\pm$0.6} & 20.06{\scriptsize $\pm$0.9} &  0.061{\scriptsize $\pm$0.005} \\
				& ANP & 58.47{\scriptsize $\pm$1.02} & 22.20{\scriptsize $\pm$1.1} & 0.037{\scriptsize $\pm$0.003} \\
				& ANP-VS & \textbf{59.15{\scriptsize $\pm$1.2}} & \textbf{22.35{\scriptsize $\pm$0.6}} & \textbf{0.035{\scriptsize $\pm$0.001}}\\
				\bottomrule
				\bottomrule
			\end{tabular}
		}
		\par\vspace{20pt}
	\end{minipage}
\end{table*}

%% file: vulnerability_vis.tex
\begin{figure*}[t]
	\begin{center}
	\begin{subfigure}
	\centering
	\setcounter{subfigure}{0}%
	\renewcommand\thesubfigure{(\alph{subfigure})}%
	\subfigure
		{\includegraphics[width=4.55cm, height=5.85cm]{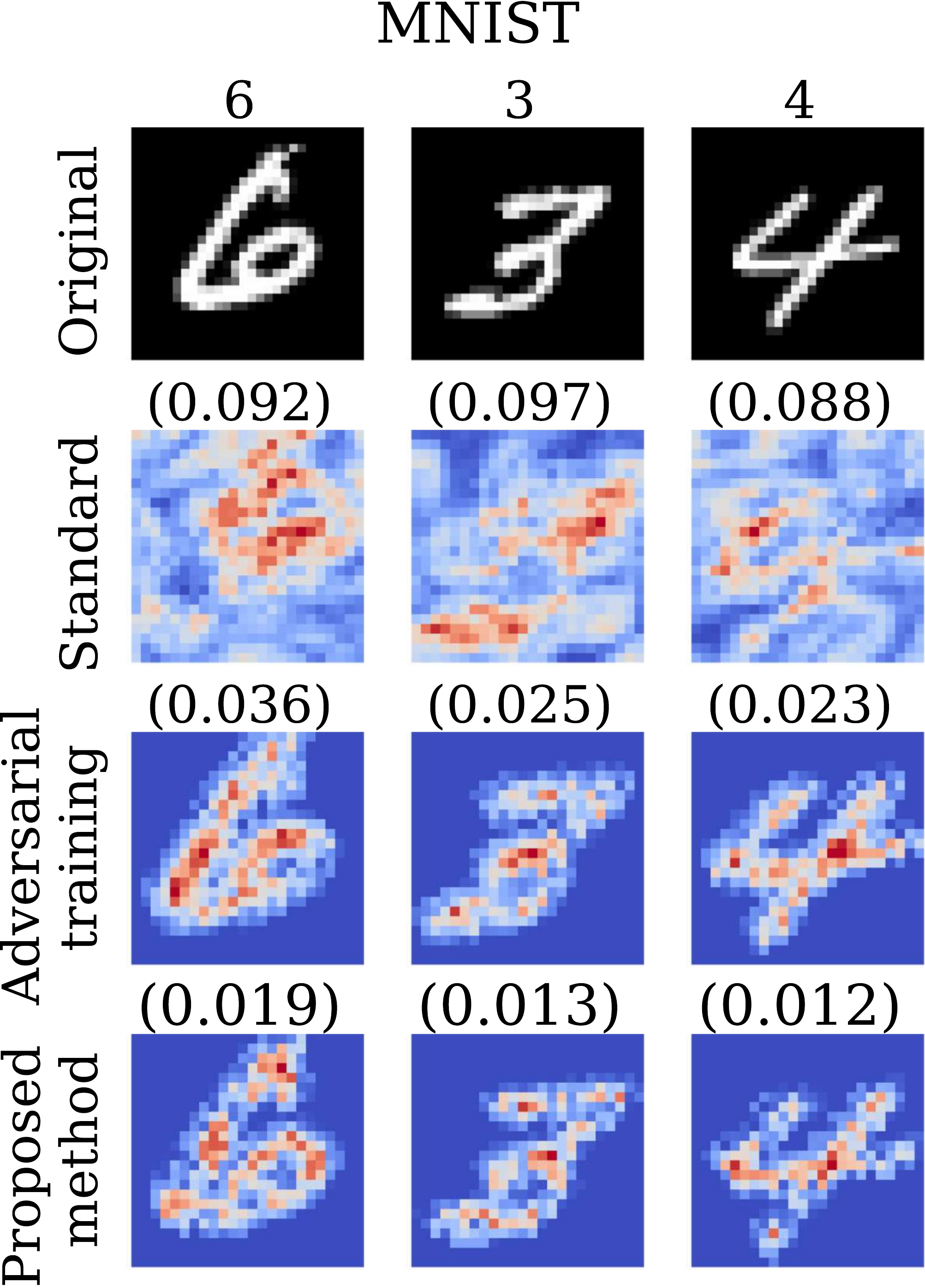}}
		\label{a) MNIST}
		\hspace{0.25in}
	\subfigure
		{\includegraphics[width=4.55cm, height=5.85cm]{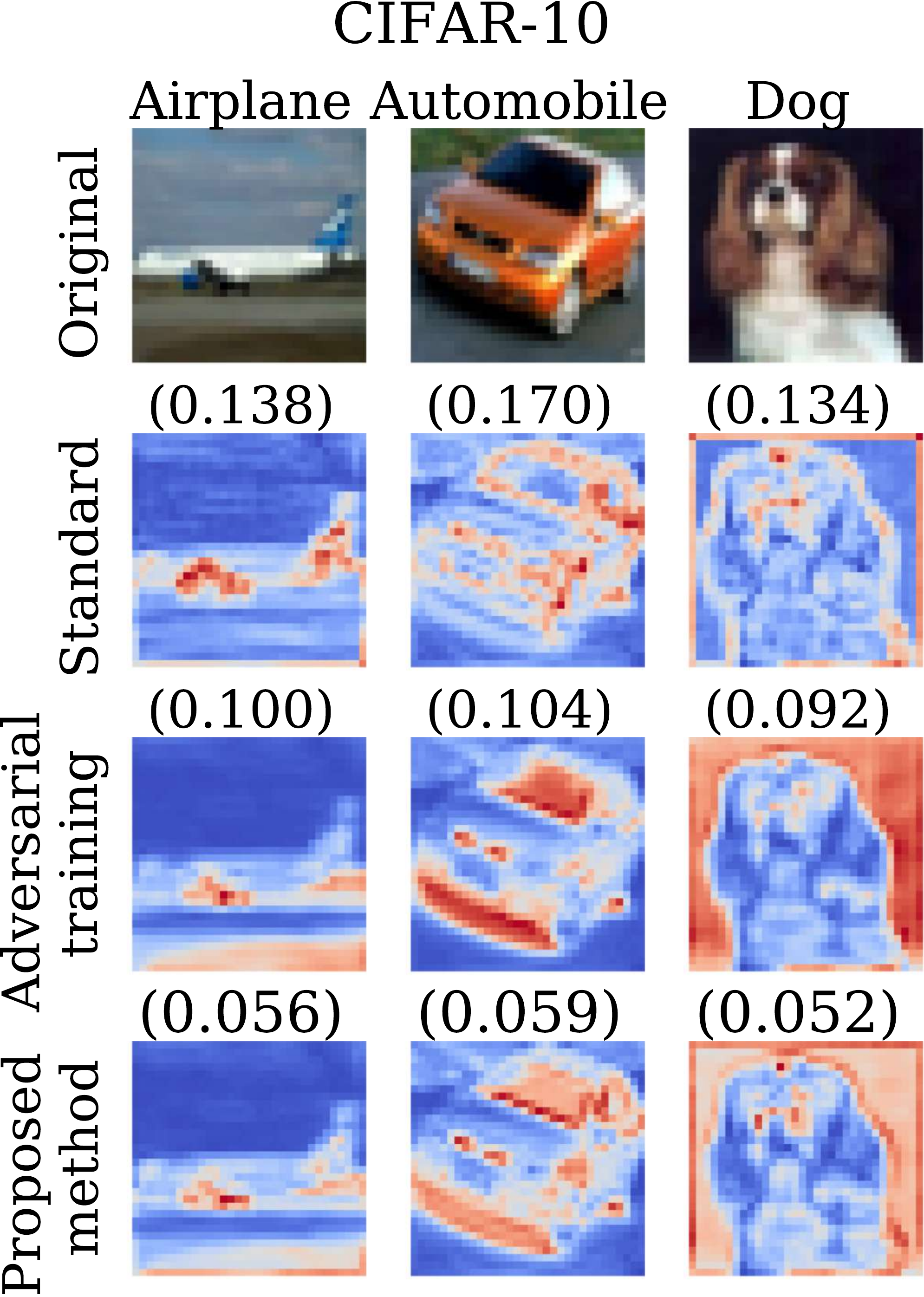}}
		\label{b) CIFAR10}
			\hspace{0.25in}
	\subfigure
		{\includegraphics[width=4.55cm, height=5.85cm]{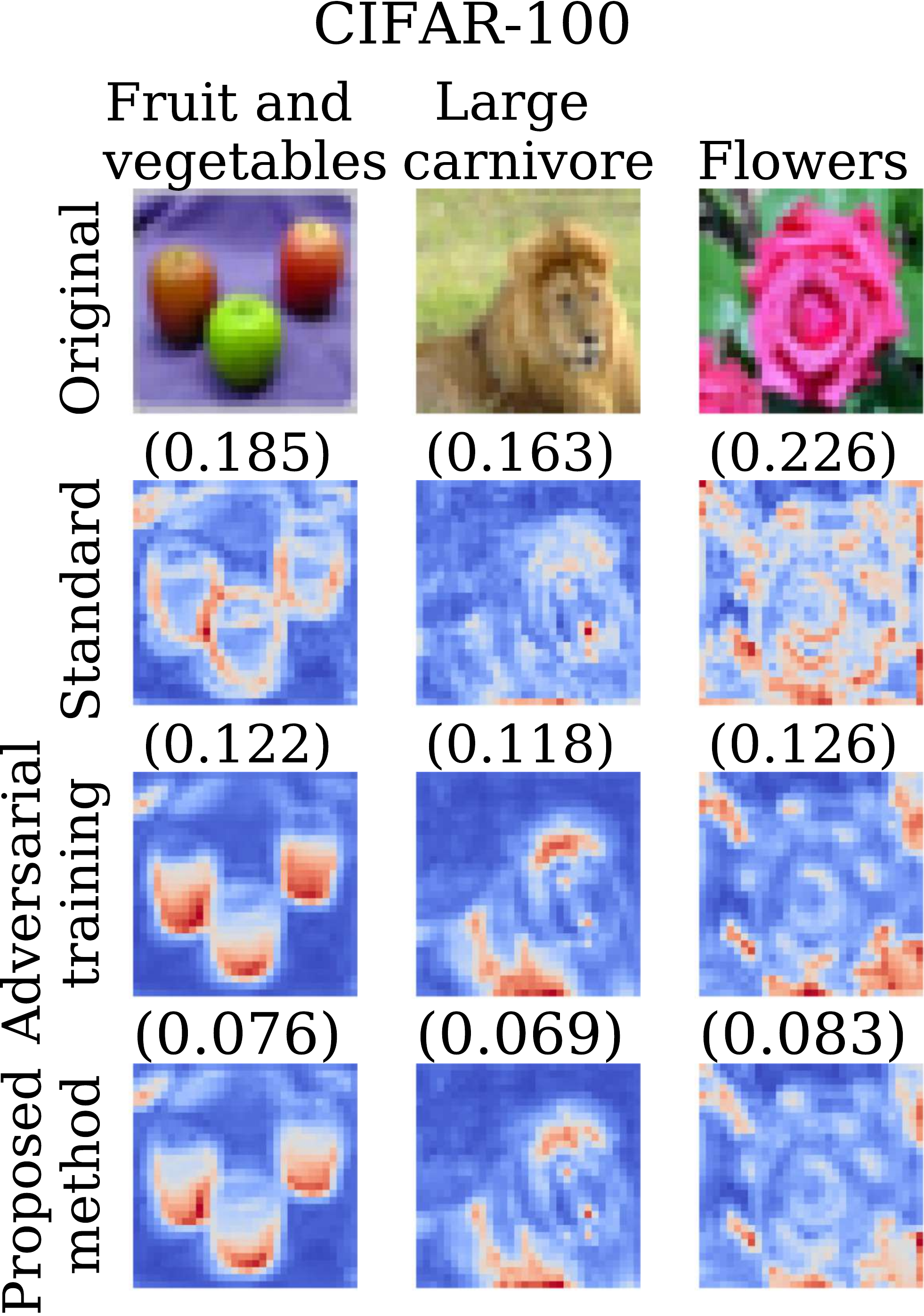}}
		\label{c) CIFAR100}
\end{subfigure}%
\begin{subfigure}
\centering
    \def\svgwidth{0.92\linewidth}
    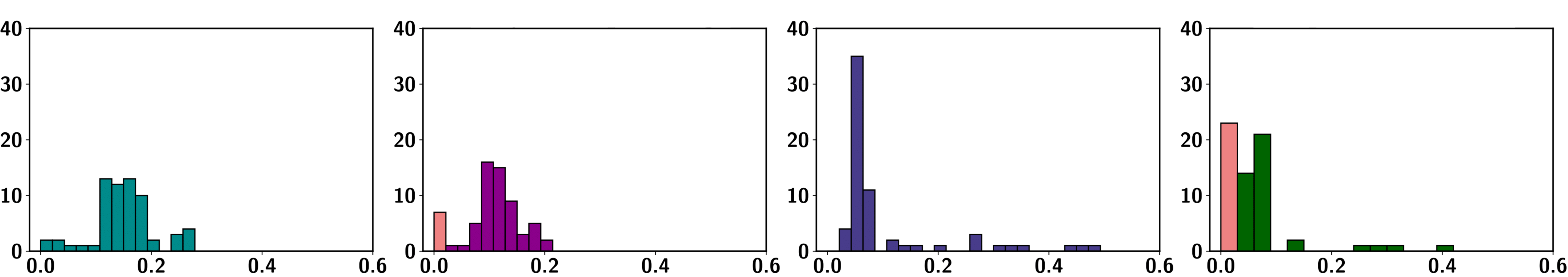

\end{subfigure}
		\caption{{Top:} Visualization of the vulnerability of the latent-features with respect to the input pixels for various set of datasets. {Bottom:} Histogram of vulnerability of the features for the input layer for CIFAR-10 with the number of zeros shown in orange color. \label{fig:vulnerability_fig}}
	\end{center}
\end{figure*}
\begin{figure*}
	\centering
    \def\svgwidth{0.80\linewidth}
    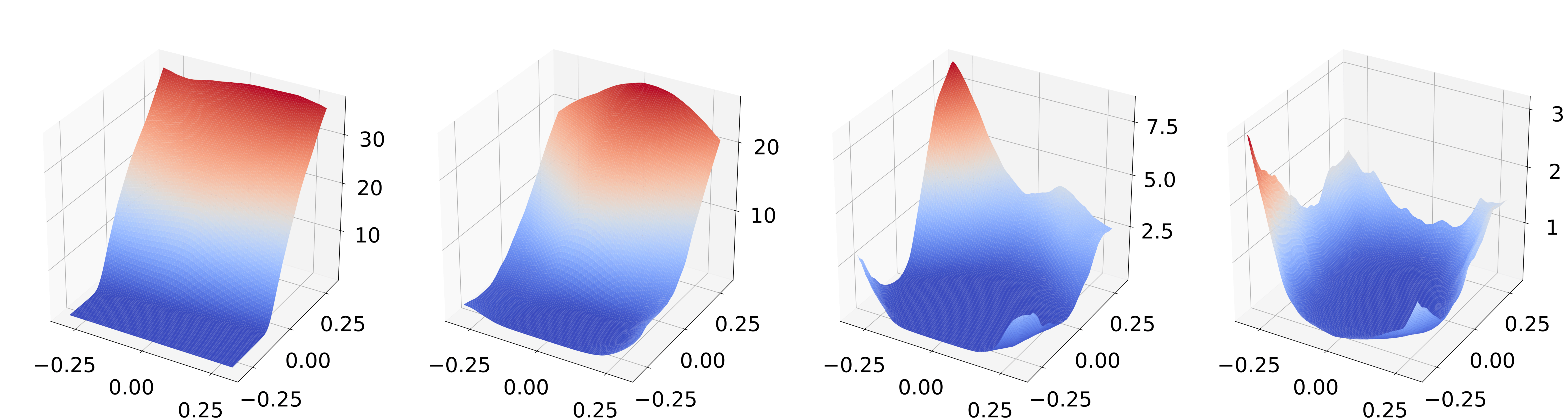

	\caption{Comparison of loss landscapes for various methods. We observe that our proposed method results in a much smoother and flattened loss surface compared to adversarial training. The z axis represents the loss projected along two random directions. \label{fig:loss_surface}}
\end{figure*}

%% file: vis_plots.pdf_tex
\begingroup%
  \makeatletter%
  \providecommand\color[2][]{%
    \errmessage{(Inkscape) Color is used for the text in Inkscape, but the package 'color.sty' is not loaded}%
    \renewcommand\color[2][]{}%
  }%
  \providecommand\transparent[1]{%
    \errmessage{(Inkscape) Transparency is used (non-zero) for the text in Inkscape, but the package 'transparent.sty' is not loaded}%
    \renewcommand\transparent[1]{}%
  }%
  \providecommand\rotatebox[2]{#2}%
  \newcommand*\fsize{\dimexpr\f@size pt\relax}%
  \newcommand*\lineheight[1]{\fontsize{\fsize}{#1\fsize}\selectfont}%
  \ifx\svgwidth\undefined%
    \setlength{\unitlength}{3186.04760742bp}%
    \ifx\svgscale\undefined%
      \relax%
    \else%
      \setlength{\unitlength}{\unitlength * \real{\svgscale}}%
    \fi%
  \else%
    \setlength{\unitlength}{\svgwidth}%
  \fi%
  \global\let\svgwidth\undefined%
  \global\let\svgscale\undefined%
  \makeatother%
  \begin{picture}(1,0.17679108)%
    \lineheight{1}%
    \setlength\tabcolsep{0pt}%
    \put(0,0){\includegraphics[width=\unitlength,page=1]{vis_plots.pdf}}%
    \put(0.08029584,0.17955723){\color[rgb]{0,0,0}\makebox(0,0)[lt]{\begin{minipage}{0.33007645\unitlength}\raggedright Standard\\ \end{minipage}}}%
    \put(0.30392715,0.17955723){\color[rgb]{0,0,0}\makebox(0,0)[lt]{\begin{minipage}{0.33007645\unitlength}\raggedright Bayesian pruning\end{minipage}}}%
    \put(0.54497816,0.17955723){\color[rgb]{0,0,0}\makebox(0,0)[lt]{\begin{minipage}{0.33007645\unitlength}\raggedright Adversarial training\end{minipage}}}%
    \put(0.80344887,0.17955723){\color[rgb]{0,0,0}\makebox(0,0)[lt]{\begin{minipage}{0.33007645\unitlength}\raggedright Proposed method\end{minipage}}}%
  \end{picture}%
\endgroup%

%% file: loss_plots.pdf_tex
\begingroup%
  \makeatletter%
  \providecommand\color[2][]{%
    \errmessage{(Inkscape) Color is used for the text in Inkscape, but the package 'color.sty' is not loaded}%
    \renewcommand\color[2][]{}%
  }%
  \providecommand\transparent[1]{%
    \errmessage{(Inkscape) Transparency is used (non-zero) for the text in Inkscape, but the package 'transparent.sty' is not loaded}%
    \renewcommand\transparent[1]{}%
  }%
  \providecommand\rotatebox[2]{#2}%
  \newcommand*\fsize{\dimexpr\f@size pt\relax}%
  \newcommand*\lineheight[1]{\fontsize{\fsize}{#1\fsize}\selectfont}%
  \ifx\svgwidth\undefined%
    \setlength{\unitlength}{4752.29882813bp}%
    \ifx\svgscale\undefined%
      \relax%
    \else%
      \setlength{\unitlength}{\unitlength * \real{\svgscale}}%
    \fi%
  \else%
    \setlength{\unitlength}{\svgwidth}%
  \fi%
  \global\let\svgwidth\undefined%
  \global\let\svgscale\undefined%
  \makeatother%
  \begin{picture}(1,0.26680742)%
    \lineheight{1}%
    \setlength\tabcolsep{0pt}%
    \put(0,0){\includegraphics[width=\unitlength,page=1]{loss_plots.pdf}}%
    \put(0.03451945,0.27325981){\color[rgb]{0,0,0}\makebox(0,0)[lt]{\begin{minipage}{0.18819839\unitlength}\raggedright \end{minipage}}}%
    \put(0.03877251,0.27963942){\color[rgb]{0,0,0}\makebox(0,0)[lt]{\begin{minipage}{0.27219654\unitlength}\raggedright \end{minipage}}}%
    \put(0.04727866,0.2902721){\color[rgb]{0,0,0}\makebox(0,0)[lt]{\begin{minipage}{0.02020208\unitlength}\raggedright \end{minipage}}}%
    \put(0.00900102,0.46464801){\color[rgb]{0,0,0}\makebox(0,0)[lt]{\begin{minipage}{0.11376964\unitlength}\raggedright \end{minipage}}}%
    \put(0.01538062,0.37958659){\color[rgb]{0,0,0}\makebox(0,0)[lt]{\begin{minipage}{0.44976225\unitlength}\raggedright \end{minipage}}}%
    \put(0.07009134,0.24754888){\color[rgb]{0,0,0}\makebox(0,0)[lt]{\lineheight{1000}\smash{\begin{tabular}[t]{l}\small{Standard}\end{tabular}}}}%
    \put(0.30297643,0.25073315){\color[rgb]{0,0,0}\makebox(0,0)[lt]{\lineheight{1000}\smash{\begin{tabular}[t]{l}\small{Bayesian pruning}\end{tabular}}}}%
    \put(0.53587377,0.25052603){\color[rgb]{0,0,0}\makebox(0,0)[lt]{\lineheight{1000}\smash{\begin{tabular}[t]{l}\small{Adversarial training}\end{tabular}}}}%
    \put(0.78147029,0.25027085){\color[rgb]{0,0,0}\makebox(0,0)[lt]{\lineheight{1000}\smash{\begin{tabular}[t]{l}\small{Proposed method}\end{tabular}}}}%
  \end{picture}%
\endgroup%

%% file: 6_conclusion.tex
\section{Conclusion}\label{conclusion}

We hypothesized that the adversarial vulnerability of deep neural networks comes from the distortion in the latent feature space since if they are suppressed at any layers of the deep network, they will not affect the prediction. Based on this hypothesis, we formally defined the vulnerability of a latent feature and proposed \emph{Adversarial Neural Pruning}(ANP) as a defense mechanism to achieve adversarial robustness as well as a means of achieving a memory- and computation-efficient deep neural networks. Specifically, we proposed a Bayesian formulation that trains a Bayesian pruning (dropout) mask for adversarial robustness of the network. Then we introduced \emph{Vulnerabiltiy Suppression (VS)} loss, that minimizes network vulnerability. To this end, we proposed \emph{Adversarial Neural Pruning with Vulnerability Suppression (ANP-VS)}, which prunes the vulnerable features by learning pruning masks for them, to minimize the adversarial loss and feature-level vulnerability. We experimentally validate ANP-VS on three datasets against recent baselines, and the results show that it significantly improves the robustness of the deep network, achieving state-of-the-art results on all the datasets. Further qualitative analysis shows that our method obtains more interpretable latent features compared to standard counterparts, effectively suppresses feature-level distortions, and obtains smoother loss surface. We hope that our work leads to more follow-up works on adversarial learning that investigates the distortion in the latent feature space, which we believe is a more direct cause of adversarial vulnerability.

\section*{Acknowledgements}
We thank the anonymous reviewers for their insightful comments and suggestions. We are also grateful to the authors of~\citet{lee2019adaptive} (Juho Lee, Saehoon Kim, and Hae Beom Lee)---for their insights and efforts replicating, extending and discussing our experimental results. This work was supported by Google AI Focused Research Award, Institute of Information \& communications Technology Planning \& Evaluation (IITP) grant funded by the Korea government (MSIT) (No.2020-0-00153), Penetration Security Testing of ML Model Vulnerabilities and Defense), Engineering Research Center Program through the National Research Foundation of Korea (NRF) funded by the Korean Government MSIT (NRF-2018R1A5A1059921), Institute of Information \& communications Technology Planning \& Evaluation (IITP) grant funded by the Korea government (MSIT)  
(No.2019-0-00075), and Artificial Intelligence Graduate School Program (KAIST). Any opinions, findings, and conclusions or recommendations expressed in this material are those of the authors and do not necessarily reflect the views of the funding agencies.

%% file: 7_appendix.tex
\newpage
\appendix

\input{appendix_results_table}
\section{Adversarial variational information bottleneck}\label{appendix:bbd}
\input{neurons_table}
In this section, we extend the idea of Adversarial Neural Pruning to Variational Information Bottleneck (VIB). Variational information bottleneck~\cite{dai18vib} uses information theoretic bound to reduce the redundancy between adjacent layers. Let \(p(\vec{h_{i}}|\vec{h_{i-1}})\) define the conditional probability and  \(I(\vec{h_{i}}; \vec{h_{i-1}})\) define the mutual information between hidden layer activations \(\vec{h_{i}}\) and \(\vec{h_{i-1}}\) for every hidden layer in the network. For every hidden layer \(\vec{h_{i}}\), we would like to minimize the information bottleneck~\cite{tishby2000information} \(I(\vec{h_{i}; h_{i-1}})\) to remove interlayer redundancy, while simultaneously maximizing the mutual information \(I(\vec{h_{i}};\vec{y})\) between \(\vec{h_{i}}\) and the output \(\vec{y}\) to encourage accurate predictions of adversarial examples. The layer-wise energy \(\mathcal{L}_{i}\) can be written as:
\begin{equation}
\mathcal{L}_{i} = \beta_{i}I(\vec{h_{i}}; \vec{h_{i-1}}) - I(\vec{h_{i}};\vec{y})
\end{equation}
	
 The output layer approximates the true distribution \(p(\vec{y}|\vec{h_{L}})\) via some tractable alternative \(q(\vec{y}|\vec{h_{L}})\). 
Using variational bounds, we can invoke the upper bound as:
 \begin{equation}\label{eq:vib_initial_loss}
 \begin{aligned}
	\calL_{i} = \beta_{i} \mathbb{E}_{\vec{h_{i-1}} \sim p(\vec{h_{i-1}})}[\KL[p(\vec{h_{i}}|\vec{h_{i-1}})||q(\vec{h_{i}})]] - \\ \mathbb{E}_{\{\vec{x},\vec{y}\} \sim D, h \sim p(\vec{h}|\tilde{\vec{x}})}[\log q(\vec{y}|\vec{h_{L}})] \geq \calL_{i}
\end{aligned}
\end{equation}
	
\(\calL_{i}\) in Equation~\ref{eq:vib_initial_loss} is composed of two terms, the first is the KL divergence between \(p(\vec{h_{i}}|\vec{h_{i-1}})\) and \(q(\vec{h_{i}})\), which approximates information extracted by \(\vec{h_{i}}\) from \(\vec{h_{i-1}}\) and the second term represents constancy with respect to the adversarial data distribution. In order to optimize Equation~\ref{eq:vib_initial_loss}, we can define the parametric form for the distributions \(p(\vec{h_{i}}|\vec{h_{i-1}})\) and \(q(\vec{h_{i}})\) as follow:
	\begin{align}\label{eq:vib_parametric_form}
	\begin{split}
	p(\vec{h_{i}}|\vec{h_{i-1}}) &= \mathcal{N}(\vec{h_{i}}; f_{i}(\vec{h_{i-1}}) \odot \mu_{i}, \diag[f_{i}(\vec{h_{i-1}})^2 \odot \sigma_{i}^2] \\
	q(\vec{h_{i}}) &= \mathcal{N}(\vec{h_{i}};0,\diag[\vec{\xi_{i}}])
	\end{split}
	\end{align}
	where \(\vec{\xi}_i\) is an unknown vector of variances that can be learned from data. The gaussian assumptions help us to get an interpretable, closed-form approximation for the KL term from Equation~\ref{eq:vib_initial_loss}, which allows us to directly optimize \(\vec{\xi}_i\) out of the model.
	\begin{equation}\label{eq:vib_kl}
	\begin{aligned}
	\mathbb{E}_{\vec{h_{i-1}} \sim p(\vec{h_{i-1}})}[\KL[p(\vec{h_{i}}|\vec{h_{i-1}})||q(\vec{h_{i}})]] = \\ \sum_{j} \left[\log\left(1+\frac{\mu_{i,j}^2}{\sigma_{i,j}^2}\right)\right]
	\end{aligned}
	\end{equation}
	
	The final variational information bottleneck can thus be obtained using Equation~\ref{eq:vib_kl}:
	\begin{equation}\label{appendix:vib_final_loss}
	\begin{aligned}
	\mathcal{L} = \sum_{i=1}^{L}\beta_{i} \sum_{j=1}^{r_{i}}\left[\log\left(1+\frac{\mu_{i,j}^2}{\sigma_{i,j}^2}\right)\right] - \\ \mathbb{E}_{\{\vec{x},\vec{y}\}\sim D, \vec{h} \sim p(\vec{h}|\tilde{\vec{x}})}[\log q(\vec{y}|\vec{h_{L}})]
	\end{aligned}
	\end{equation}
	
where \(\beta \geq 0\) is a coefficient that determines the strength of the bottleneck that can be defined as the degree to which we value compression over robustness.
\section{Experiment setup} In this section, we describe our experimental settings for all the experiments. We follow the two-step pruning procedure where we pretrain all the networks using the standard-training procedure followed by network sparsification using various sparsification methods. We train each model with 200 epochs with a fixed batch size of 64. All the results are measured by computing mean and standard deviation across 5 trials upon randomly chosen seeds. 
	
Our pretrained standard Lenet 5-Caffe baseline model reaches over 99.29\% accuracy on MNIST and VGG-16 architecture reaches 92.76\% and 67.44\% on CIFAR-10 and CIFAR-100 dataset respectively after 200 epochs. We use Adam~\cite{Kingma2015AdamAM} with the learning rate for the weights to be 0.1 times smaller than those for the variational parameters as in~\cite{Neklyudov2017StructuredBP,lee2019adaptive}. For Beta-Bernoulli Dropout, we set \(\alpha/K = 10^{-4}\) for all the layers and prune the neurons/filters whose expected drop probability are smaller than a fixed threshold  \(10^{-3}\) as originally proposed in the paper. For Beta-Bernoulli Dropout, we scaled the KL-term by different values of trade-off parameter \(\beta\) where \(\beta \in \{1, 4, 8, 10, 12\}\) for Lenet-5-Caffe and \(\beta \in \{1, 2, 4, 6, 8\}\) for VGG-16. For Variational Information Bottleneck (VIBNet), we tested with trade-off parameter \(\beta\) in Equation~\ref{appendix:vib_final_loss} where \(\beta \in \{10, 30, 50, 80, 100\}\) for Lenet-5-Caffe with MNIST and \(\beta \in \{10^{-4}, 1, 20, 40, 60\}\) for VGG-16 with CIFAR-10 and CIFAR-100 dataset. For generating black-box adversarial examples, we used an adversarial trained full network for adversarial neural pruning and the standard base network for the standard Bayesian compression method.

\input{appendix_hist}	
\input{motivation_vis}
\section{More experimental results}
Due to the length limit of our paper, some results are illustrated here.
\subsection{Robustness of adversarial variational information bottleneck}
The results for ANP-VS with Variational Information Bottleneck are summarized in Table~\ref{table:appendix_mainTable}. We can observe that ANP-VS with Variational Information Bottleneck significantly outperforms the base adversarial training for robustness of adversarial examples by achieving an improvement of $\sim 2\%$ in adversarial accuracy. Note that, ANP-VS leads to $\sim 50\%$ and $\sim 25\%$ reduction in vulnerability for CIFAR-10 and CIFAR-100 dataset with memory and computation efficiency. We emphasize that ANP can similarly be extended to any existing or future sparsification method to improve performance. Table~\ref{table:neuronsResults} further shows the number of units for the baselines and our proposed method.

\subsection{Features vulnerability}
Figure~\ref{fig:hist_appendix} shows the histogram of the feature vulnerability for various datasets. We consistently observe that standard Bayesian pruning zeros out some of the distortions, AT reduces the distortion level of all the features and ANP-VS does both, with the most significant number of features with zero distortion and low distortion level in general which confirms that our proposed method works successfully as a defense against adversarial attacks. All these results overall confirm the effectiveness of our defense.

\subsection{Features visualization}
One might also be curious about the representation of the robust and vulnerable features in the latent-feature space. We visualize the robust and vulnerable features based on the vulnerability of a feature in the latent-feature space from our paper. Figure \ref{fig:motivation} shows the visualization of robust and vulnerable features in the latent space for adversarial training. Note that, AT contains features with high vulnerability (vulnerable feature) and features with less vulnerability (robust feature), which aligns with our observation that the latent features have a varying degree of susceptibility to adversarial perturbations to the input. As future work, we plan to explore more effective ways to suppress perturbation at the intermediate latent features of deep networks.

%% file: appendix_results_table.tex
\begin{table*}[t!]
\centering
	\rtable{1.2}
			\caption{\small Robustness and compression performance for MNIST on Lenet-5-Caffe, CIFAR-10 and CIFAR-100 on VGG-16 architecture under $\ell_{\infty}$-PGD attack for ANP-VS with Variational Information Bottleneck. All the values are measured by computing mean and standard deviation across 5 trials upon randomly chosen seeds. The best results over adversarial baselines are highlighted in bold. \label{table:appendix_mainTable}}
			\resizebox{0.95\linewidth}{!}{
			\begin{tabular}{lllllllllllllll}
				\toprule
				{} & {} & {} & \multicolumn{2}{c}{Adversarial accuracy $(\uparrow)$} & \multicolumn{2}{c}{Vulnerability ($\downarrow$)} & \multicolumn{3}{c}{Computational efficiency}\\
				\cmidrule(lr){4-5}\cmidrule(lr){6-7}\cmidrule(lr){8-10}
				& Model & \vtop{\hbox{\strut Clean}\hbox{\strut accuracy ($\uparrow$)}} & \vtop{\hbox{\strut White box}\hbox{\strut attack}} & \vtop{\hbox{\strut Black box}\hbox{\strut attack}} & \vtop{\hbox{\strut White box}\hbox{\strut attack}} & \vtop{\hbox{\strut Black box }\hbox{\strut attack}} & Memory ($\downarrow$) & xFLOPS ($\uparrow$) & Sparsity ($\uparrow$)\\
				\midrule
				\parbox[t]{2mm}{\multirow{9}{*}{\rotatebox[origin=c]{90}{MNIST}}}
				& Standard & 99.29{\scriptsize $\pm$0.02} & 0.00{\scriptsize $\pm$0.0} & 8.02{\scriptsize $\pm$0.9} & 0.129{\scriptsize $\pm$0.001} & 0.113{\scriptsize $\pm$0.000} & 100.0{\scriptsize $\pm$0.00} & 1.00{\scriptsize $\pm$0.00} &  0.00{\scriptsize $\pm$0.00}\\
				& BP & 99.32{\scriptsize $\pm$0.04} & 5.66{\scriptsize $\pm$0.4} & 15.47{\scriptsize $\pm$0.3} & 0.091{\scriptsize $\pm$0.001} & 0.078{\scriptsize $\pm$0.001} & 4.34{\scriptsize $\pm$0.34} & 9.39{\scriptsize $\pm$0.25} & 82.46{\scriptsize $\pm$0.61}\\
				\cline{2-10}
				& AT& 99.14{\scriptsize $\pm$0.02} & 88.03{\scriptsize $\pm$0.7}  & 94.18{\scriptsize $\pm$0.8} & 0.045{\scriptsize $\pm$0.001} & 0.040{\scriptsize $\pm$0.000} & 100.0{\scriptsize $\pm$0.00} & 1.00{\scriptsize $\pm$0.00} &  0.00{\scriptsize $\pm$0.00}\\
				& AT BNN & 99.16{\scriptsize $\pm$0.05} & 88.44{\scriptsize $\pm$0.4} & 94.87{\scriptsize $\pm$0.2} & 0.364{\scriptsize $\pm$0.023} & 0.199{\scriptsize $\pm$0.031}  & 200.0{\scriptsize $\pm$0.00} & 0.50{\scriptsize $\pm$0.00} & 0.00{\scriptsize $\pm$0.00} \\
				& Pretrained AT  & \textbf{99.18{\scriptsize $\pm$0.06}} & 88.26{\scriptsize $\pm$0.6} & 94.49{\scriptsize $\pm$0.7} & 0.412{\scriptsize $\pm$0.035} & 0.381{\scriptsize $\pm$0.029} & 100.0{\scriptsize $\pm$0.00} & 1.00{\scriptsize $\pm$0.00} &  0.00{\scriptsize $\pm$0.00}\\
				& ADMM  & 99.01{\scriptsize $\pm$0.02} & 88.47{\scriptsize $\pm$0.4} & 94.61{\scriptsize $\pm$0.7} & 0.041{\scriptsize $\pm$0.002} & 0.038{\scriptsize $\pm$0.001} & 100.00{\scriptsize $\pm$0.00} & 1.00{\scriptsize $\pm$0.00} &  80.00{\scriptsize $\pm$0.00}\\
				& TRADES  & 99.07{\scriptsize $\pm$0.04} & 89.67{\scriptsize $\pm$0.4} & 95.04{\scriptsize $\pm$0.6} & 0.037{\scriptsize $\pm$0.001} & 0.033{\scriptsize $\pm$0.001} & 100.0{\scriptsize $\pm$0.00} & 1.00{\scriptsize $\pm$0.00} &  0.00{\scriptsize $\pm$0.00}\\
				\cline{2-10}
				& ANP-VS (ours) & 98.86{\scriptsize $\pm$0.02} & \textbf{90.11{\scriptsize $\pm$0.9}} & \textbf{95.14{\scriptsize $\pm$0.8}} & \textbf{0.017{\scriptsize $\pm$0.001}} & \textbf{0.015{\scriptsize $\pm$0.001}} & \textbf{4.87{\scriptsize $\pm$0.21}} & \textbf{10.06{\scriptsize $\pm$0.87}} & \textbf{78.48{\scriptsize $\pm$0.42}}\\
				\hline  \TBstrut
				
				\parbox[t]{2mm}{\multirow{9}{*}{\rotatebox[origin=c]{90}{CIFAR-10}}}
				& Standard & 92.76{\scriptsize $\pm$0.1} & 13.79{\scriptsize $\pm$0.8} & 41.65{\scriptsize $\pm$0.9} & 0.077{\scriptsize $\pm$0.001} & 0.065{\scriptsize $\pm$0.001} & 100.0{\scriptsize $\pm$0.00} &  1.00{\scriptsize $\pm$0.00}  & 0.00{\scriptsize $\pm$0.00} \\
				& BP & 92.73{\scriptsize $\pm$0.1} & 12.28{\scriptsize $\pm$0.3} & 76.35{\scriptsize $\pm$0.8} & 0.035{\scriptsize $\pm$0.002} & 0.032{\scriptsize $\pm$0.001} & 12.38{\scriptsize $\pm$0.12} & 2.38{\scriptsize $\pm$0.0.05} & 76.35{\scriptsize $\pm$0.23}\\
				\cline{2-10}
				& AT & 87.50{\scriptsize $\pm$0.5} & 49.85{\scriptsize $\pm$0.9} & 63.70{\scriptsize $\pm$0.6} & 0.050{\scriptsize $\pm$0.002} & 0.047{\scriptsize $\pm$0.001} & 100.0{\scriptsize $\pm$0.00} &  1.00{\scriptsize $\pm$0.00} & 0.00{\scriptsize $\pm$0.00} \\
				& AT BNN & 86.69{\scriptsize $\pm$0.5} & 51.87{\scriptsize $\pm$0.9} & 64.92{\scriptsize $\pm$0.9} & 0.267{\scriptsize $\pm$0.013} & 0.238{\scriptsize $\pm$0.011} & 200.0{\scriptsize $\pm$0.00} & 0.50{\scriptsize $\pm$0.00} & 0.00{\scriptsize $\pm$0.00} \\
				& Pretrained AT & 87.50{\scriptsize $\pm$0.4} &  52.25{\scriptsize $\pm$0.7} & 66.10{\scriptsize $\pm$0.8} & 0.041{\scriptsize $\pm$0.002} & 0.036{\scriptsize $\pm$0.001} & 100.0{\scriptsize $\pm$0.00} &  1.00{\scriptsize $\pm$0.00}  & 0.00{\scriptsize $\pm$0.00} \\
				& ADMM  & 78.15{\scriptsize $\pm$0.7} & 47.37{\scriptsize $\pm$0.6} & 62.15{\scriptsize $\pm$0.8} & 0.034{\scriptsize $\pm$0.002} & 0.030{\scriptsize $\pm$0.002} & 100.00{\scriptsize $\pm$0.00} & 1.00{\scriptsize $\pm$0.00} &  75.00{\scriptsize $\pm$0.00}\\
				& TRADES & 80.33{\scriptsize $\pm$0.5} & 52.08{\scriptsize $\pm$0.7} & 64.80{\scriptsize $\pm$0.5} & 0.045{\scriptsize $\pm$0.001} & 0.042{\scriptsize $\pm$0.005} & 100.0{\scriptsize $\pm$0.00} &  1.00{\scriptsize $\pm$0.00}  & 0.00{\scriptsize $\pm$0.00} \\
				\cline{2-10}
				& ANP-VS (ours) & \textbf{87.56{\scriptsize $\pm$0.2}} &  \textbf{53.41{\scriptsize $\pm$0.5}} & \textbf{68.12{\scriptsize $\pm$0.7}} & \textbf{0.025{\scriptsize $\pm$0.002}} & \textbf{0.021{\scriptsize $\pm$0.001}} & \textbf{12.09{\scriptsize $\pm$0.26}} & \textbf{2.43{\scriptsize $\pm$0.02}} &  \textbf{77.02{\scriptsize $\pm$0.32}}\\
				\hline  \TBstrut
				
				\parbox[t]{2mm}{\multirow{9}{*}{\rotatebox[origin=c]{90}{CIFAR-100}}}
				& Standard & 67.44{\scriptsize $\pm$0.7} & 2.81{\scriptsize $\pm$0.2} & 14.94{\scriptsize $\pm$0.8} & 0.143{\scriptsize $\pm$0.007} & 0.119{\scriptsize $\pm$0.005} & 100.0{\scriptsize $\pm$0.00} &  1.00{\scriptsize $\pm$0.00}  & 0.00{\scriptsize $\pm$0.00} \\
				& BP & 69.09{\scriptsize $\pm$0.5} & 2.73{\scriptsize $\pm$0.3} & 19.53{\scriptsize $\pm$0.4} & 0.084{\scriptsize $\pm$0.001} & 0.073{\scriptsize $\pm$0.001} & 18.46{\scriptsize $\pm$0.42} & 1.95{\scriptsize $\pm$0.03} & 63.84{\scriptsize $\pm$0.62}\\
				\cline{2-10}
				& AT & 57.79{\scriptsize $\pm$0.8} & 19.07{\scriptsize $\pm$0.8} & 32.47{\scriptsize $\pm$1.4} & 0.079{\scriptsize $\pm$0.003} & 0.071{\scriptsize $\pm$0.003} & 100.0{\scriptsize $\pm$0.00} &  1.00{\scriptsize $\pm$0.00}  & 0.00{\scriptsize $\pm$0.00} \\
				& AT BNN & 53.75{\scriptsize $\pm$0.7} & 19.40{\scriptsize $\pm$0.6} & 30.38{\scriptsize $\pm$0.2} & 0.446{\scriptsize $\pm$0.029} & 0.385{\scriptsize $\pm$0.051} & 200.0{\scriptsize $\pm$0.00} & 0.50{\scriptsize $\pm$0.00} & 0.00{\scriptsize $\pm$0.00} \\
				& Pretrained AT & 57.14{\scriptsize $\pm$0.9} & 19.86{\scriptsize $\pm$0.6} & 35.42{\scriptsize $\pm$0.4} & 0.071{\scriptsize $\pm$0.001} & 0.065{\scriptsize $\pm$0.002} & 100.0{\scriptsize $\pm$0.00} &  1.00{\scriptsize $\pm$0.00}  & 0.00{\scriptsize $\pm$0.00} \\
				& ADMM  & 52.52{\scriptsize $\pm$0.5} & 19.65{\scriptsize $\pm$0.5} & 31.30{\scriptsize $\pm$0.3} & 0.060{\scriptsize $\pm$0.001} & 0.056{\scriptsize $\pm$0.001} & 100.00{\scriptsize $\pm$0.00} & 1.00{\scriptsize $\pm$0.00} &  65.00{\scriptsize $\pm$0.00}\\
				& TRADES & 56.70{\scriptsize $\pm$0.7} & 21.21{\scriptsize $\pm$0.3} & 32.81{\scriptsize $\pm$0.6} & 0.065{\scriptsize $\pm$0.003} & 0.060{\scriptsize $\pm$0.003} & 100.0{\scriptsize $\pm$0.00} &  1.00{\scriptsize $\pm$0.00} & 0.00{\scriptsize $\pm$0.00} \\
				\cline{2-10}
				& ANP-VS (ours) & \textbf{59.77{\scriptsize $\pm$0.4}} & \textbf{21.53{\scriptsize $\pm$0.6}} & \textbf{36.82{\scriptsize $\pm$0.7}} & \textbf{0.048{\scriptsize $\pm$0.002}} & \textbf{0.042{\scriptsize $\pm$0.002}} & \textbf{16.46{\scriptsize $\pm$0.34}} & \textbf{2.06{\scriptsize $\pm$0.02}} &  \textbf{67.19{\scriptsize $\pm$0.57}} \\
				\bottomrule
			\end{tabular}}
	\end{table*}

%% file: neurons_table.tex
	\begin{table*}
		\centering
		\resizebox{0.88\textwidth}{!}{
		\begin{tabular}{llcc}
			\toprule
			{} & {} & {} \\
			& Model & No of neurons\\
			\midrule
			\parbox[t]{2mm}{\multirow{6}{*}{\rotatebox[origin=c]{90}{MNIST}}}	
			& Standard & 20 -- 50 -- 800 -- 500 \\
			& BP (BBD) & 14 -- 21 -- 150 -- 49\\
			& BP (VIB) & 12 -- 19 -- 160 -- 37\\
			& AT & 20 -- 50 -- 800 -- 500 \\
			& ANP-VS (BBD) & 7 --\hspace{0.4em} 21 -- 147 -- 46\\
			& ANP-VS (VIB) & 10 -- 23 -- 200 -- 53\\
			
			\hline
			 \TBstrut
			\parbox[t]{2mm}{\multirow{6}{*}{\rotatebox[origin=c]{90}{CIFAR-10}}}	
			& Standard & 64 -- 64 -- 128 -- 128 -- 256 -- 256 -- 256 -- 512 -- 512 -- 512 -- 512 -- 512 -- 512 -- 512 -- 512\\
			& BP (BBD) & 57 -- 59 -- 127 -- 101 -- 150 -- 71 -- \hspace{0.4em} 31 --\hspace{0.4em} 41 -- \hspace{0.6em}35 -- \hspace{0.6em}10 -- \hspace{0.6em}46 -- \hspace{0.6em}48 -- \hspace{0.4em}16 --\hspace{0.4em} 16 -- \hspace{0.4em}25\\
			& BP (VIB) & 49 -- 56 -- 106 -- 92 --\hspace{0.4em} 157 -- 74 -- \hspace{0.2em} 26 -- \hspace{0.4em}43 --\hspace{0.6em} 32 -- \hspace{0.6em}10 --\hspace{0.6em} 39 -- \hspace{0.6em}40 --\hspace{0.4em} 7 --\hspace{0.9em} 7 --\hspace{0.9em} 13 \\
			& AT & 64 -- 64 -- 128 -- 128 -- 256 -- 256 -- 256 -- 512 -- 512 -- 512 -- 512 -- 512 -- 512 -- 512 -- 512\\
			& ANP-VS (BBD) & 42 -- 57 -- 113 -- 96 -- \hspace{0.4em}147 --\hspace{0.4em} 68 --\hspace{0.4em} 25 --\hspace{0.4em} 37 --\hspace{0.4em} 27 --\hspace{0.6em} 9 --\hspace{0.6em} 39 --\hspace{0.6em} 40 --\hspace{0.6em} 13 --\hspace{0.6em} 13 --\hspace{0.6em} 12\\
			& ANP-VS (VIB) & 40 -- 57 -- 104 -- 93 -- \hspace{0.4em}174 --\hspace{0.4em} 96 --\hspace{0.4em} 30 --\hspace{0.4em} 48 -- \hspace{0.4em}39 --\hspace{0.6em} 9 -- \hspace{0.6em}49 --\hspace{0.6em} 57 --\hspace{0.6em} 10 --\hspace{0.6em} 10 --\hspace{0.6em} 12 \\
			
			\hline
			 \TBstrut
			\parbox[t]{2mm}{\multirow{6}{*}{\rotatebox[origin=c]{90}{CIFAR-100}}}	
			& Standard & 64 -- 64 -- 128 -- 128 -- 256 -- 256 -- 256 -- 512 -- 512 -- 512 -- 512 -- 512 -- 512 -- 512 -- 512\\
			& BP (BBD) & 62 -- 64 -- 128 -- 123 -- 244 -- 203 -- 84 --\hspace{0.2em} 130 -- \hspace{0.6em}95 -- \hspace{0.6em}18 -- \hspace{0.2em}152 -- \hspace{0.2em}157 -- \hspace{0.2em}32 --\hspace{0.2em} 32 -- \hspace{0.2em}101\\
			& BP (VIB) & 52 -- 64 -- 119 -- 116 --\hspace{0.2em} 229 -- 179 -- 83 -- \hspace{0.4em}99 --\hspace{0.5em} 71 -- \hspace{0.4em}17 --\hspace{0.6em} 107 -- \hspace{0.5em}110 --\hspace{0.2em} 12 --\hspace{0.2em} 11 --\hspace{0.2em} 49 \\
			& AT & 64 -- 64 -- 128 -- 128 -- 256 -- 256 -- 256 -- 512 -- 512 -- 512 -- 512 -- 512 -- 512 -- 512 -- 512\\
			& ANP-VS (BBD) & 60 -- 64 -- 126 -- 122 --\hspace{0.2em} 235 -- 185 -- 77 -- \hspace{0.4em}128 --\hspace{0.2em} 101 -- \hspace{0.4em}17 --\hspace{0.2em} 165 -- \hspace{0.5em}177 --\hspace{0.2em} 35 --\hspace{0.2em} 35 --\hspace{0.2em} 45 \\
			& ANP-VS (VIB) & 44 -- 58 -- 110 -- 109 --\hspace{0.2em} 207 -- 155 -- 81 -- \hspace{0.4em}86 --\hspace{0.6em} 66 -- \hspace{0.8em}19 --\hspace{0.6em} 88 -- \hspace{0.6em}86 --\hspace{0.6em} 15 --\hspace{0.2em} 15 --\hspace{0.2em} 36 \\
			\bottomrule
		\end{tabular}}
				\caption{\small Distribution of neurons for all the layers of Lenet-5 Caffe for MNIST and VGG-16 architecture for CIFAR-10 and CIFAR-100 datasets.\label{table:neuronsResults}}
	\end{table*}

%% file: appendix_hist.tex
\begin{figure*}[!ht]\label{fig:motivation_vis}
	\centering
	\subfigure[]
	{\includegraphics[width=3.2cm, height=3.2cm]{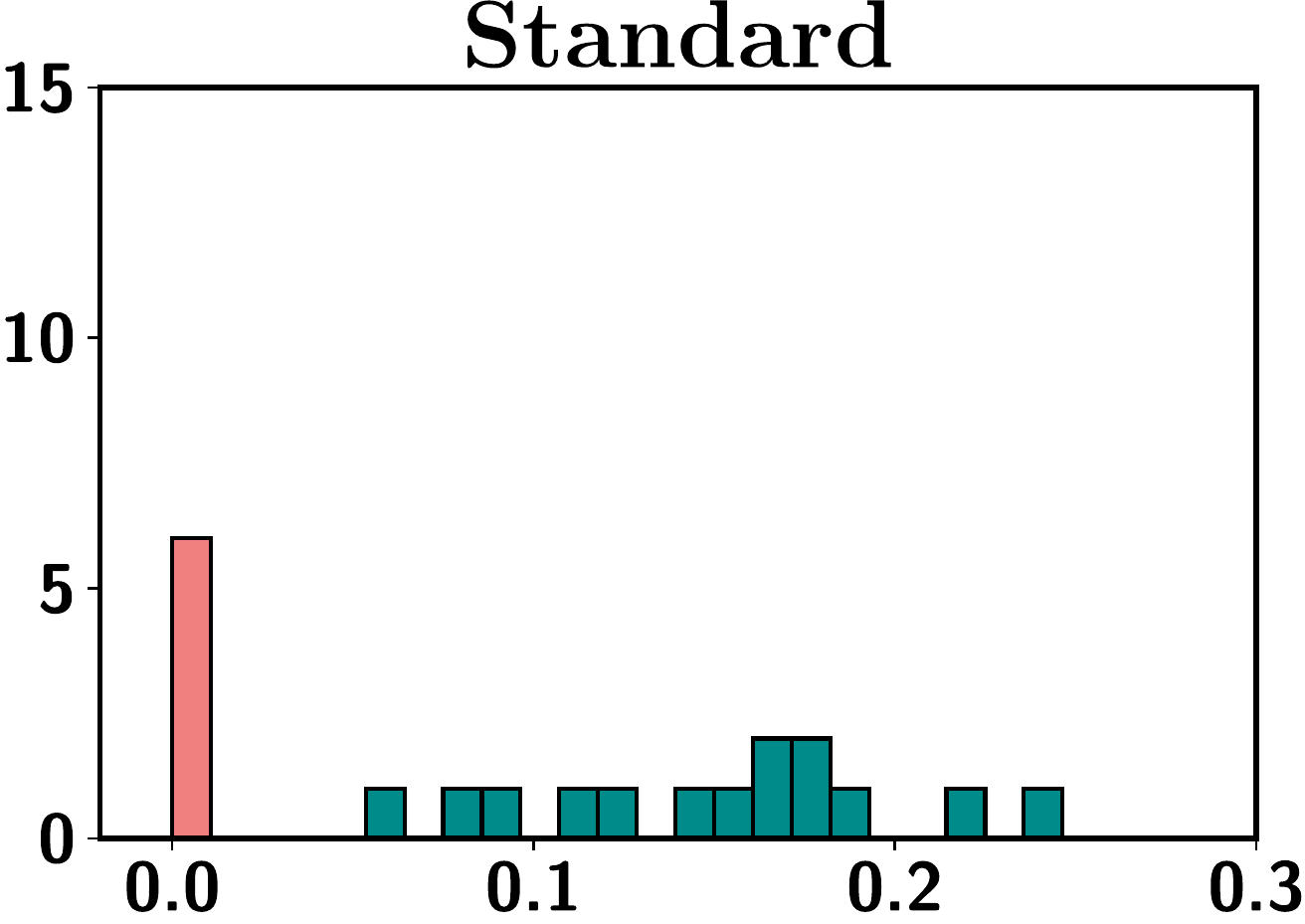}}
	\label{}
	\hspace{0.1in}
	\subfigure[]
	{\includegraphics[width=3.2cm, height=3.2cm]{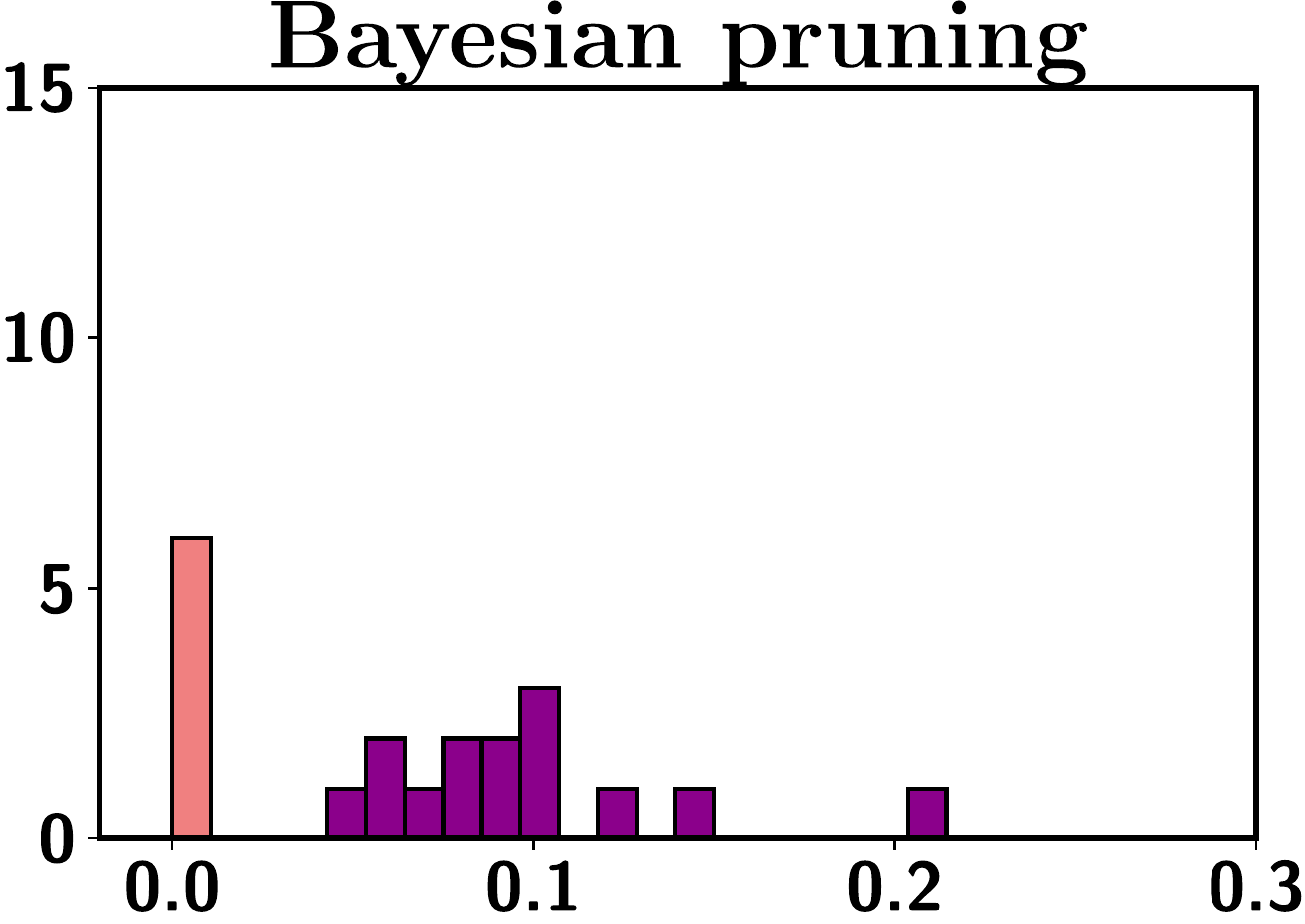}}
	\label{}
		\hspace{0.1in}
	\subfigure[]
	{\includegraphics[width=3.2cm, height=3.2cm]{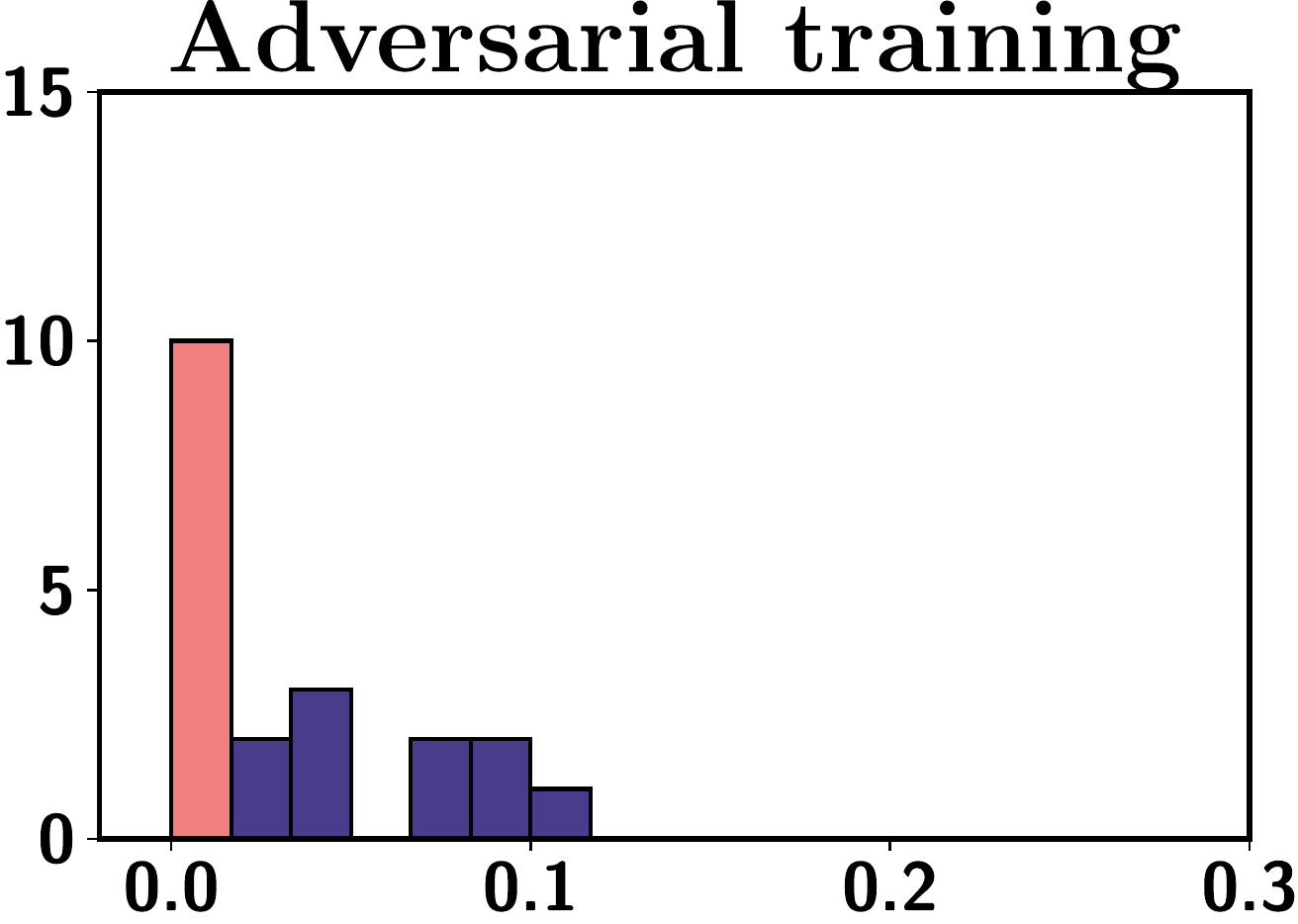}}
	\label{}
		\hspace{0.1in}
	\subfigure[]
	{\includegraphics[width=3.2cm, height=3.2cm]{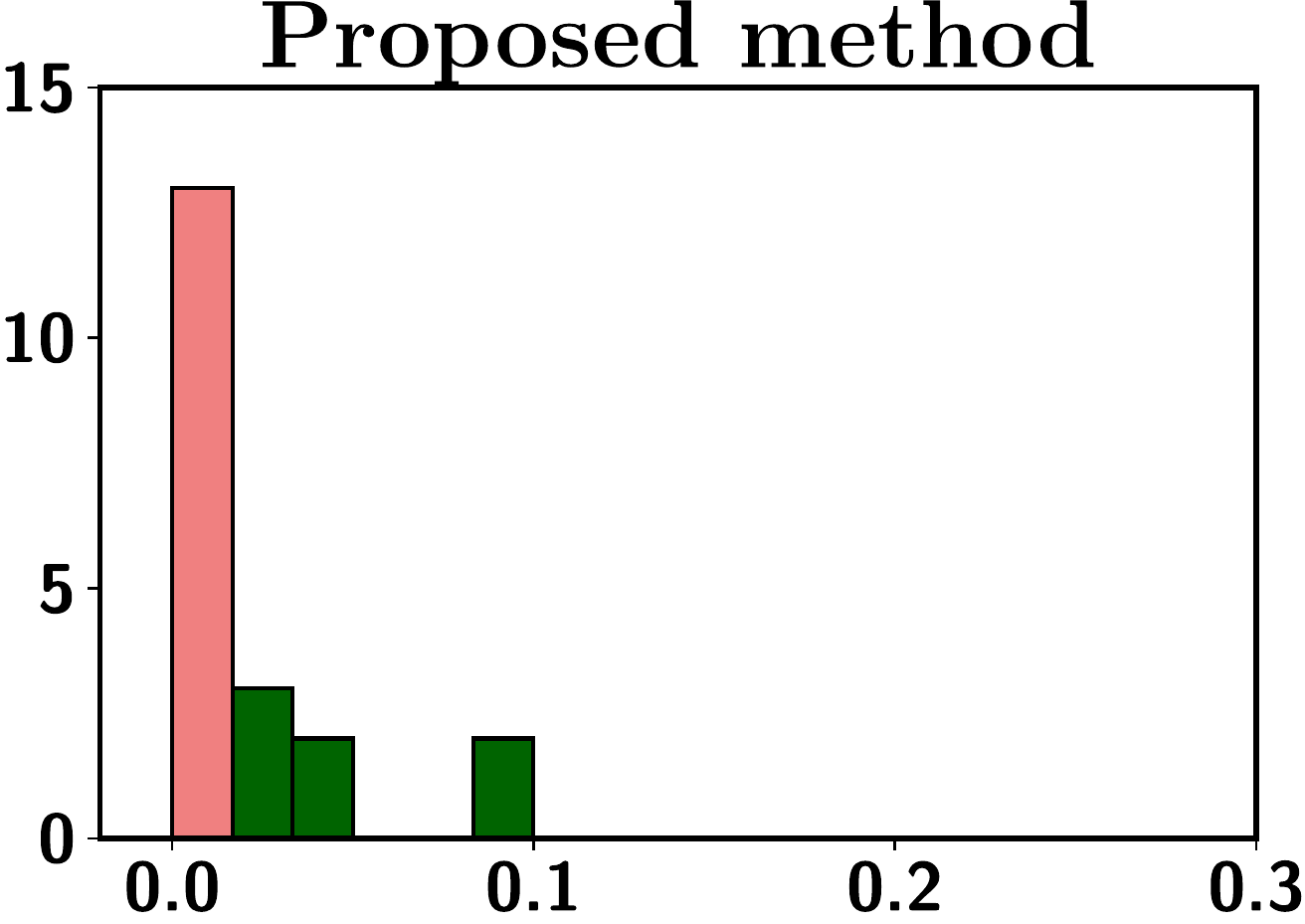}}
	\label{}\\
	\subfigure[]
	{\includegraphics[width=3.2cm, height=3.2cm]{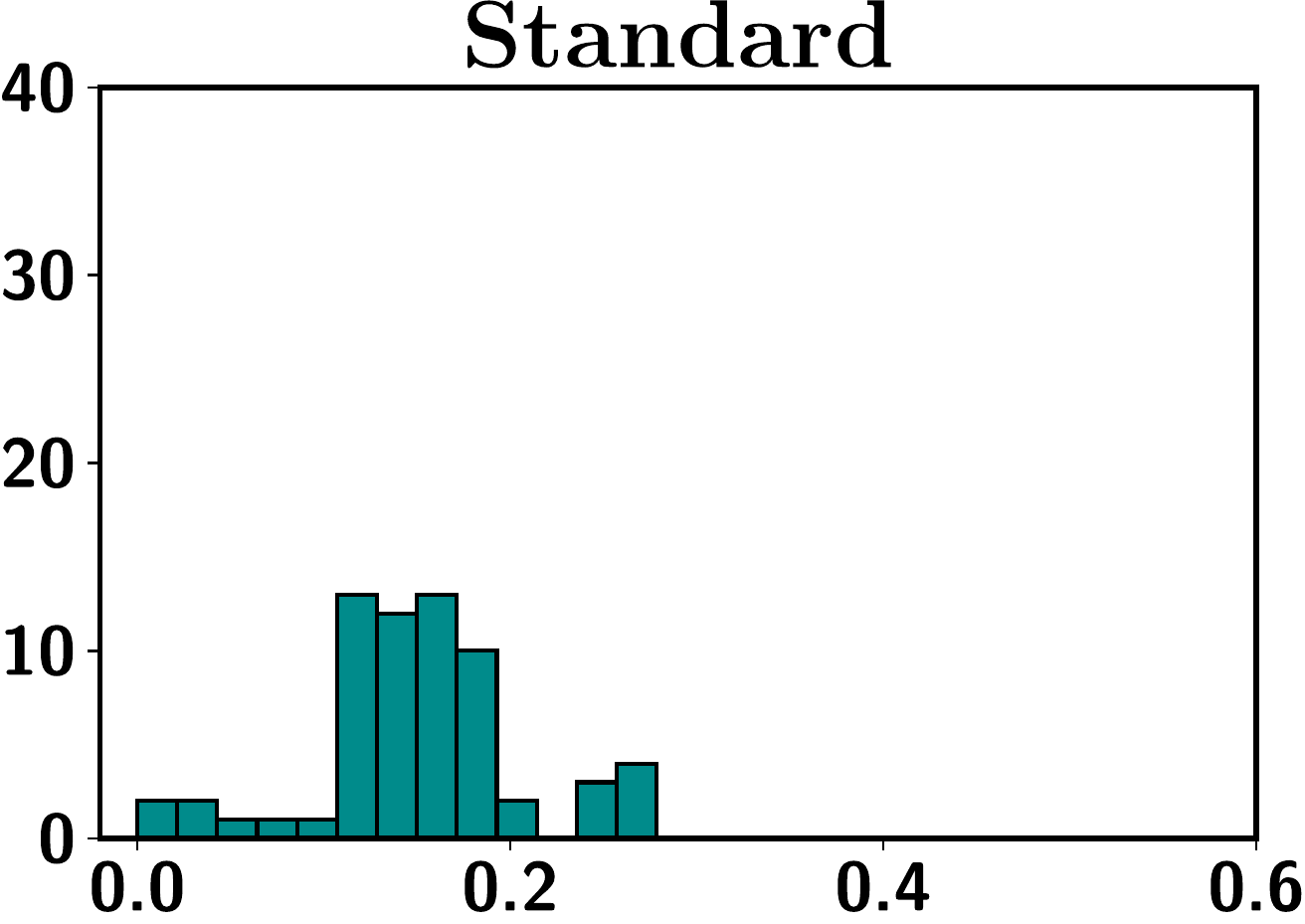}}
	\label{}
		\hspace{0.1in}
	\subfigure[]
	{\includegraphics[width=3.2cm, height=3.2cm]{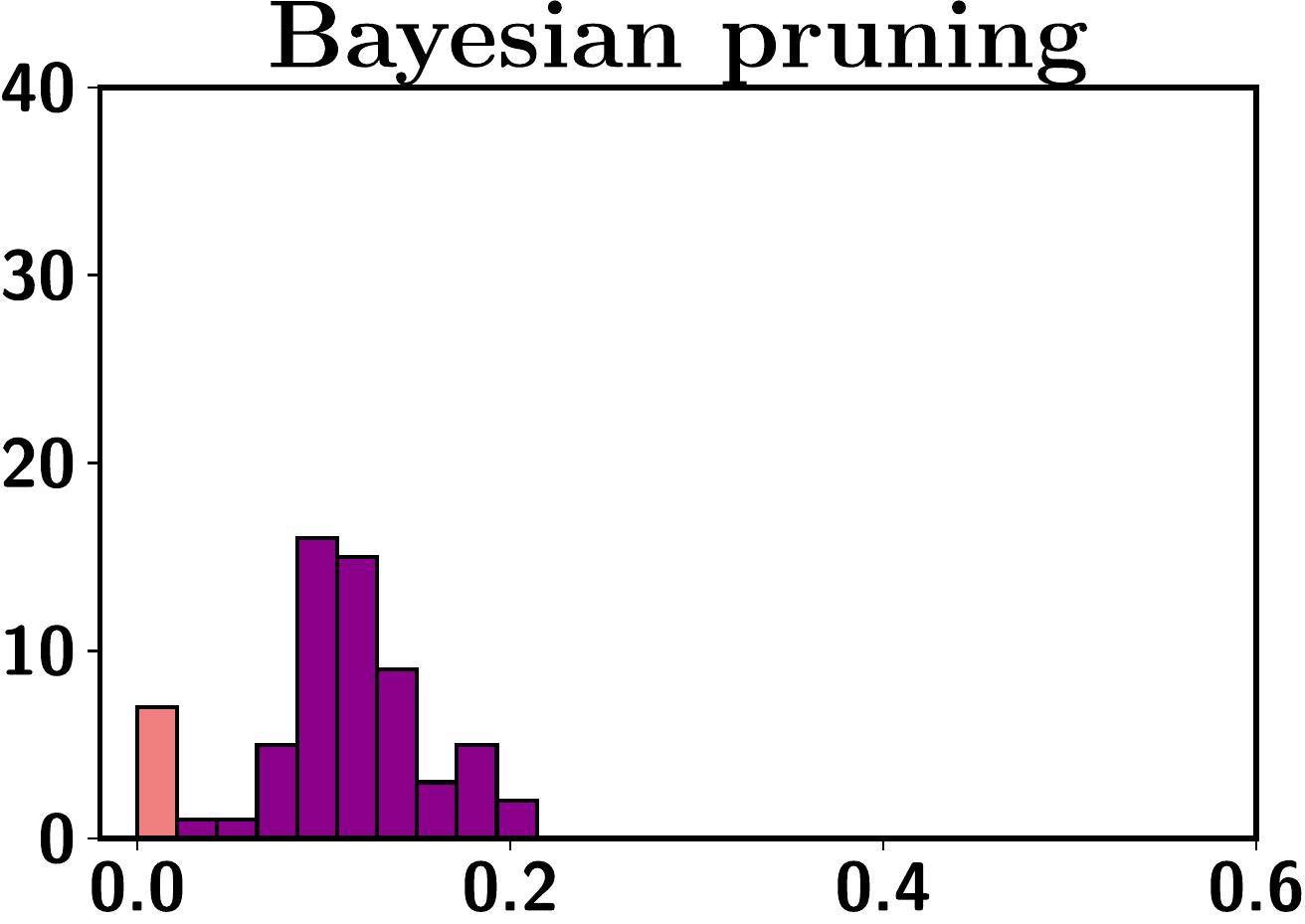}}
	\label{}
		\hspace{0.1in}
	\subfigure[]
	{\includegraphics[width=3.2cm, height=3.2cm]{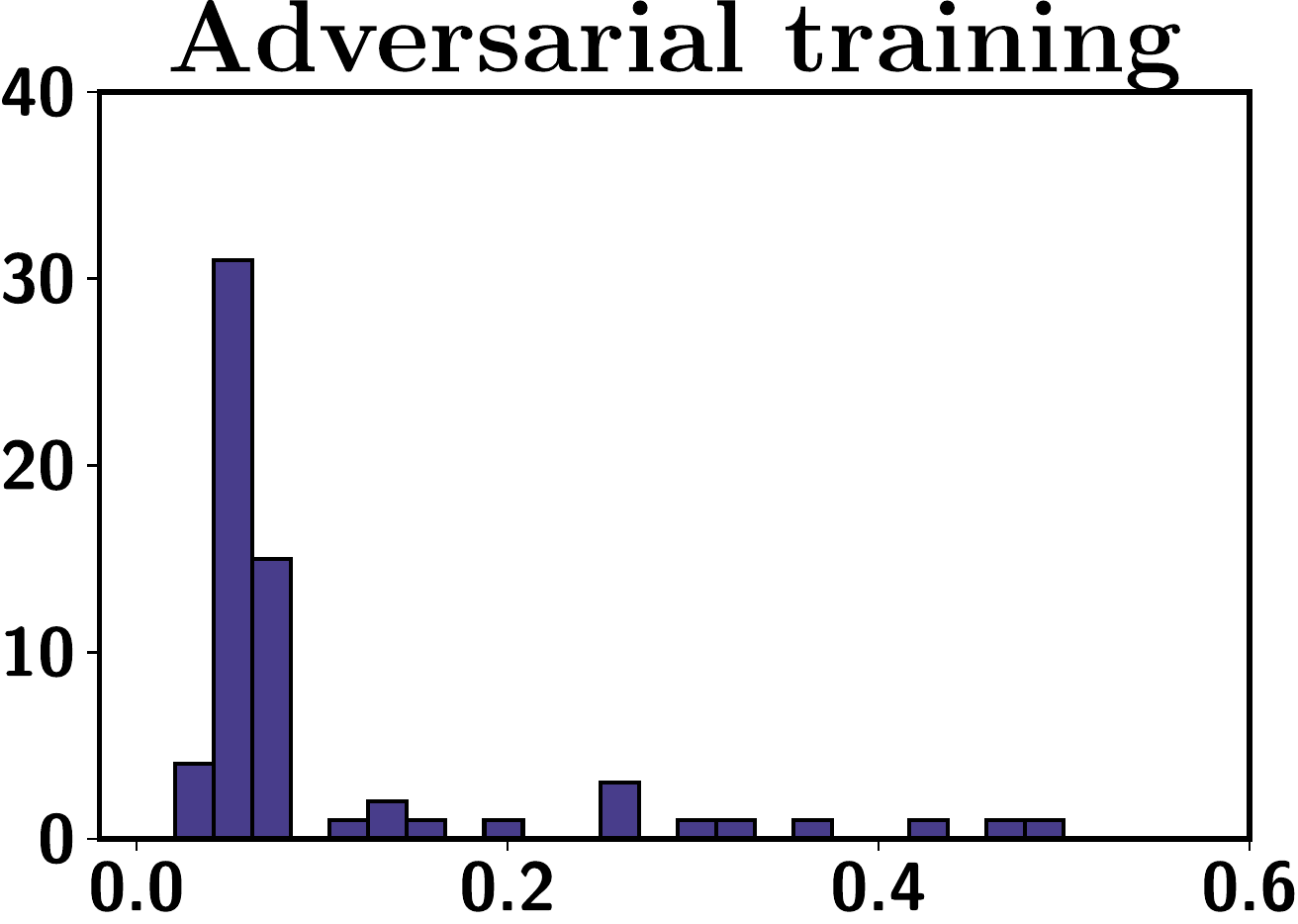}}
	\label{}
		\hspace{0.1in}
	\subfigure[]
	{\includegraphics[width=3.2cm, height=3.2cm]{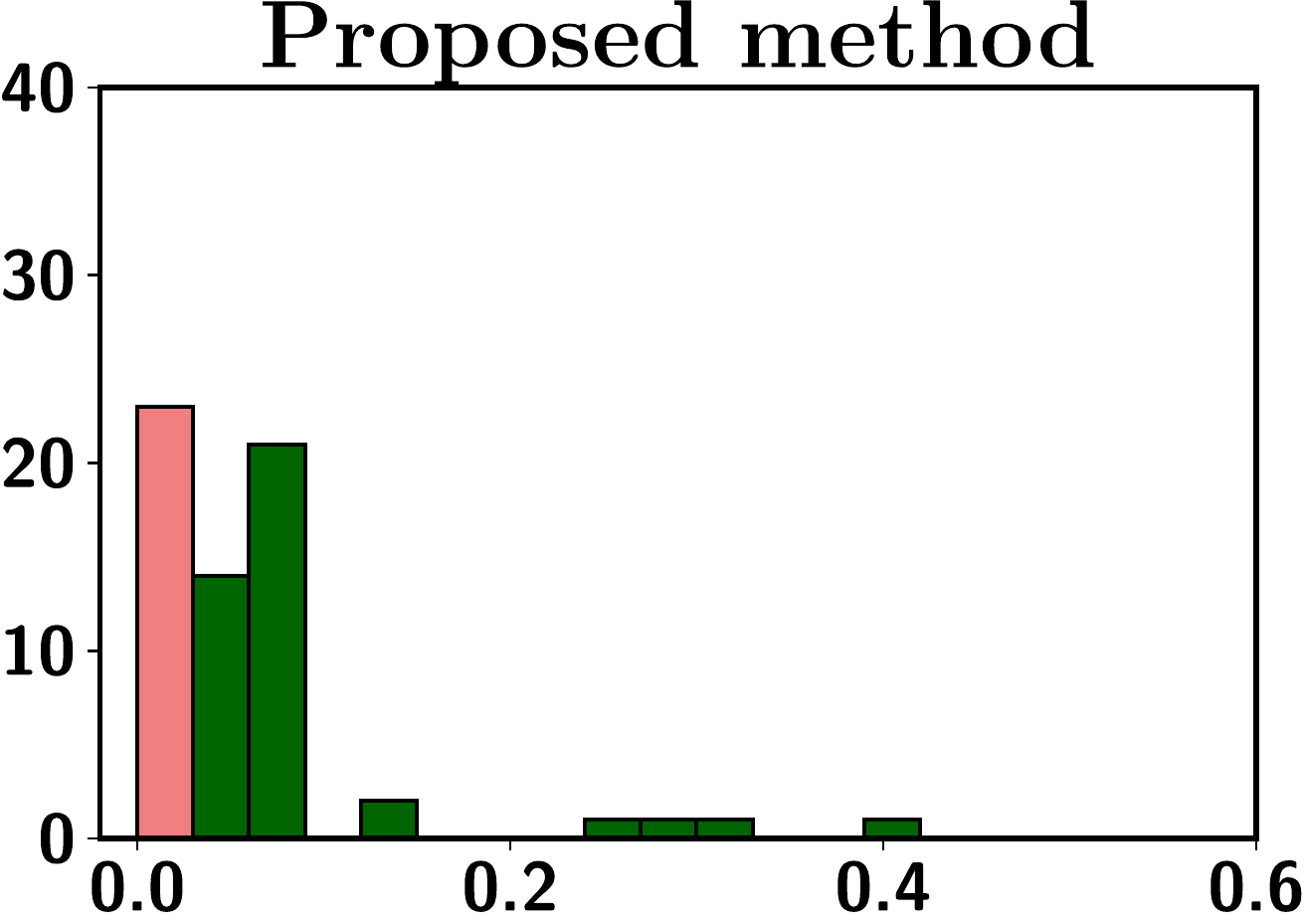}}
	\label{}\\
	\subfigure[]
		{\includegraphics[width=3.2cm, height=3.2cm]{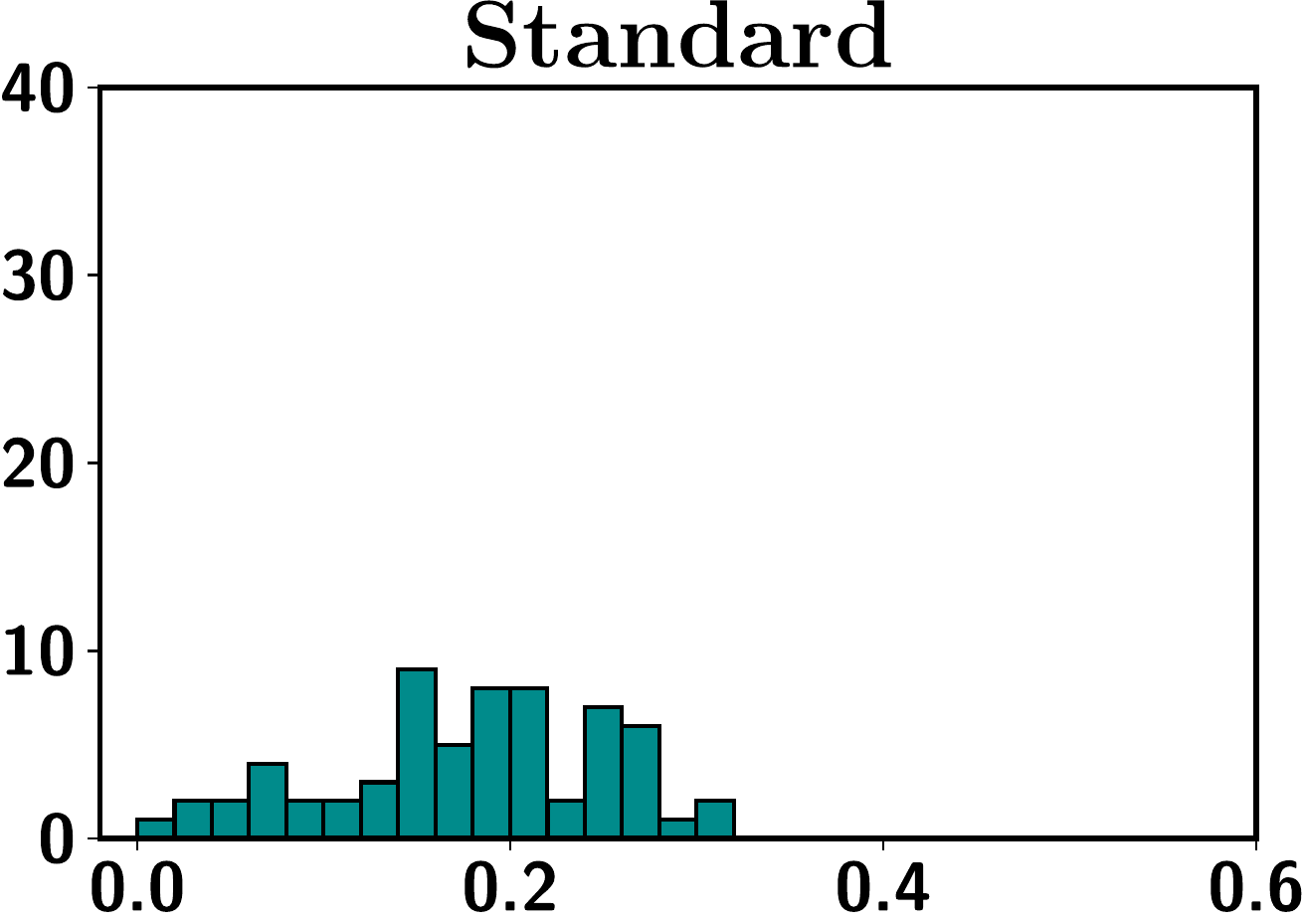}}
	\label{}
		\hspace{0.1in}
	\subfigure[]
	{\includegraphics[width=3.2cm, height=3.2cm]{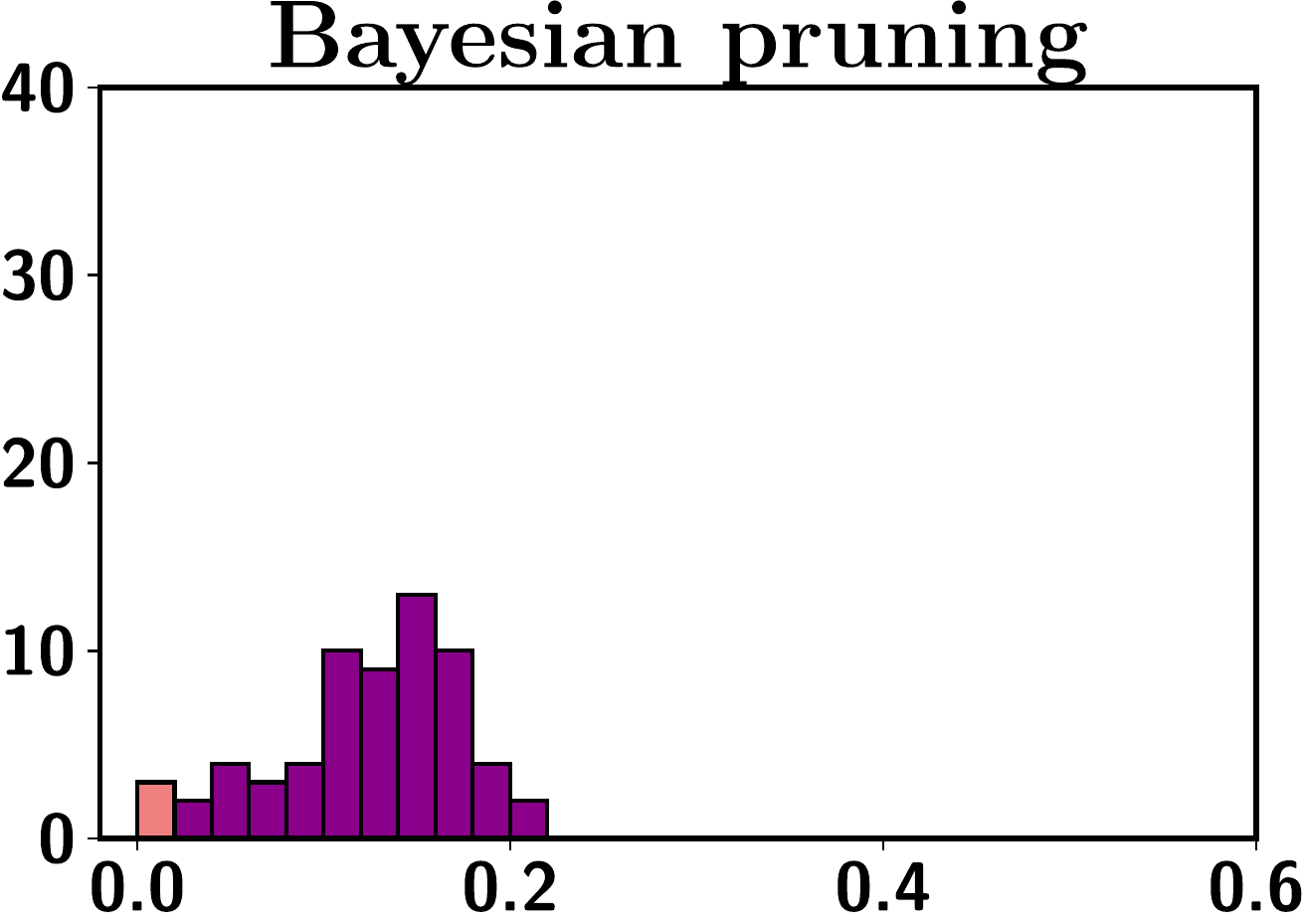}}
	\label{}
		\hspace{0.1in}
	\subfigure[]
	{\includegraphics[width=3.2cm, height=3.2cm]{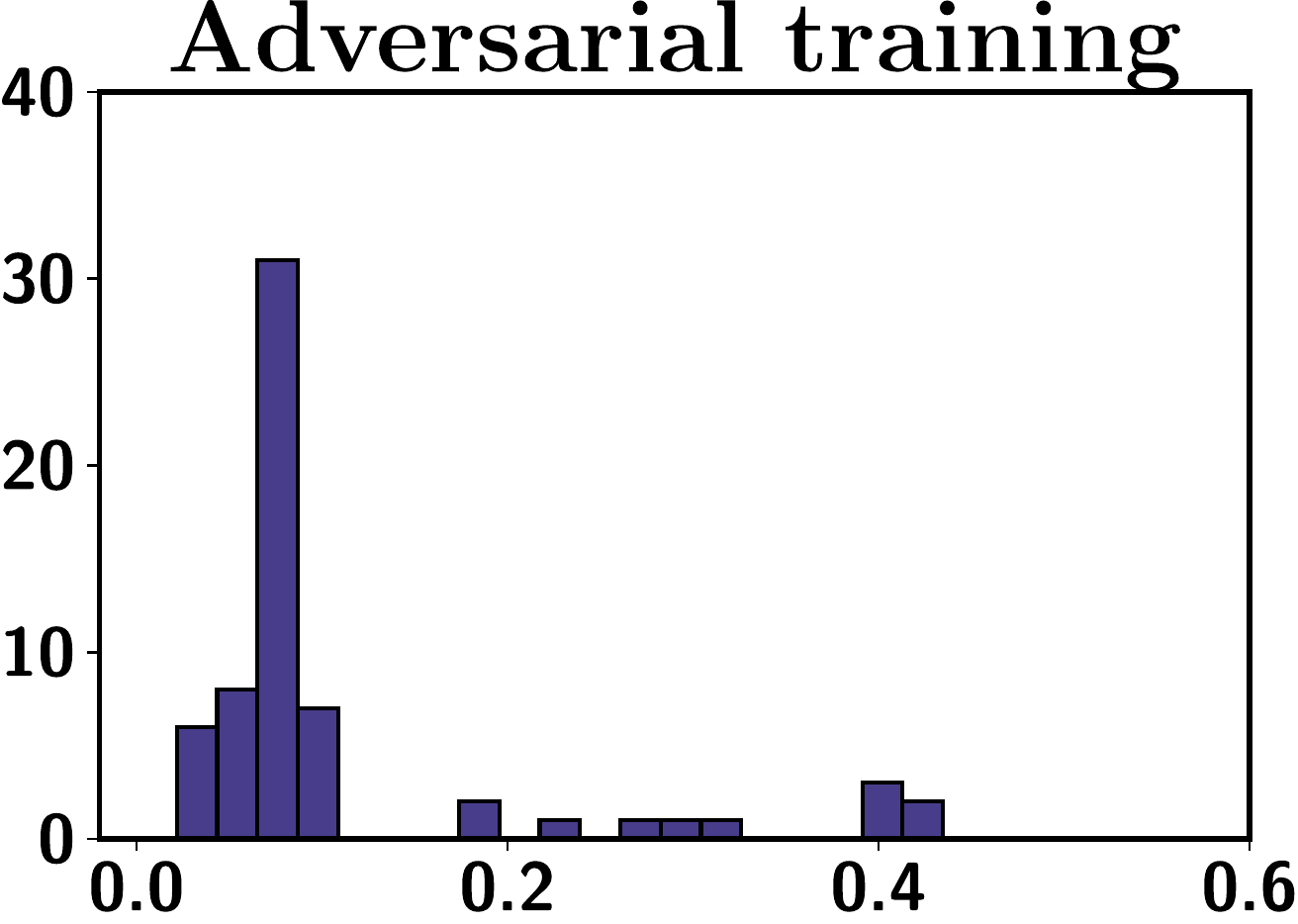}}
	\label{}
		\hspace{0.1in}
	\subfigure[]
	{\includegraphics[width=3.2cm, height=3.2cm]{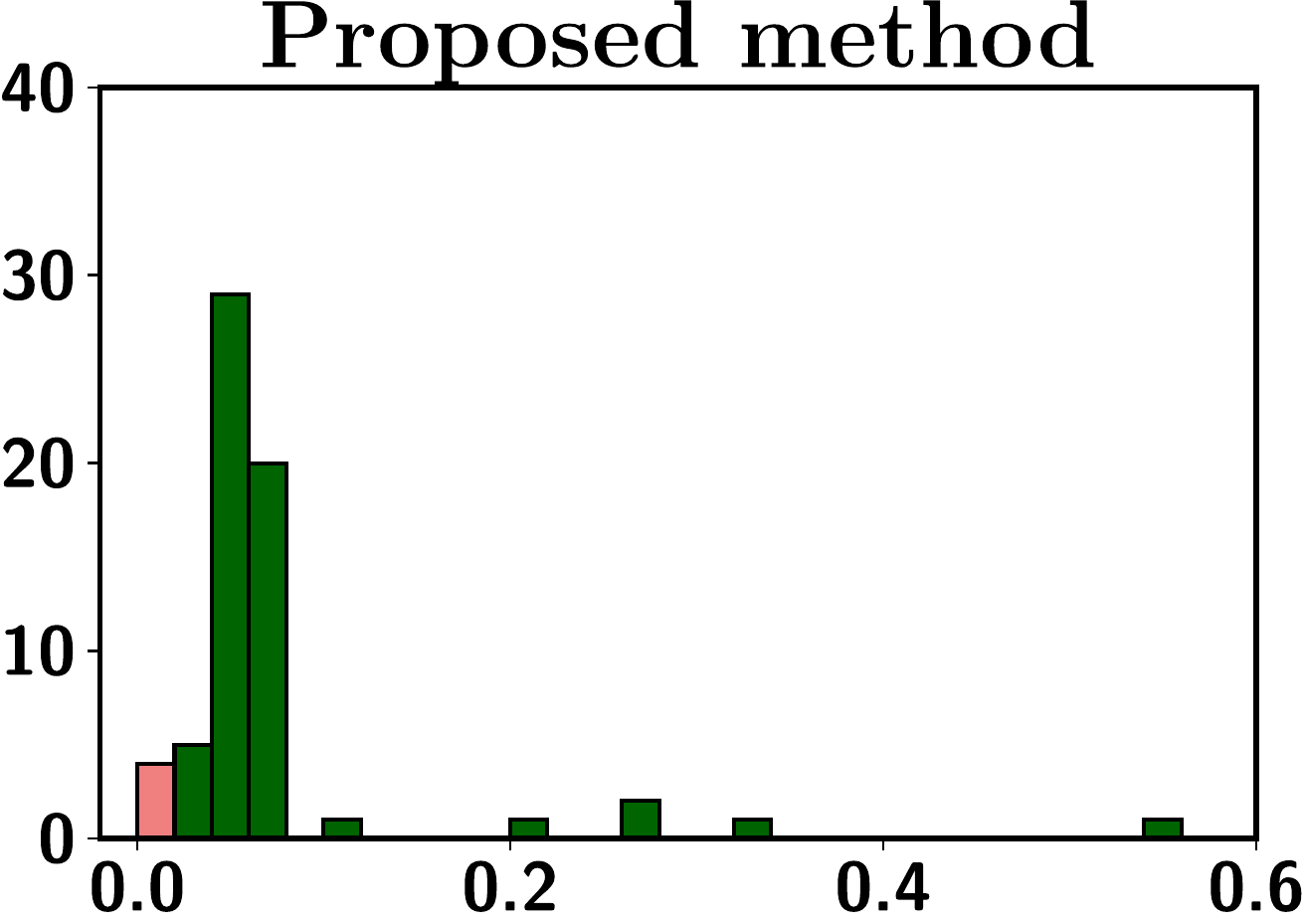}}
	\label{}
		\hspace{0.1in}
	\vspace{-0.1in}
	\caption{\small Histogram of vulnerability of the features for the input layer for MNIST in the top row, CIFAR-10 in the middle and CIFAR-100 in the bottom with the number of zeros shown in orange color.\label{fig:hist_appendix}} 
\end{figure*}

%% file: motivation_vis.tex
\begin{figure*}
	\centering
		\subfigure[]
	{\includegraphics[width=1.9cm, height=2.3cm]{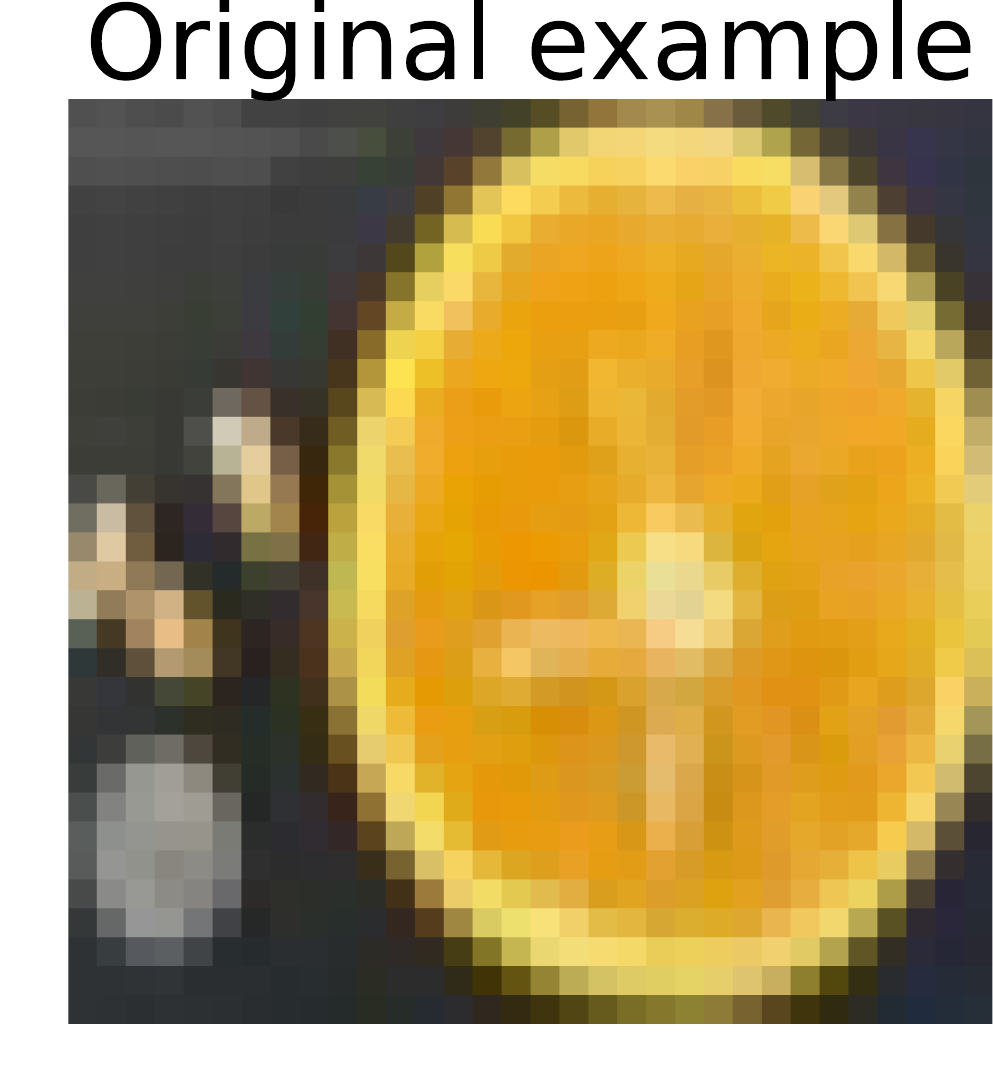}}
	\label{}
	\subfigure[]
		{\includegraphics[width=1.9cm, height=2.3cm]{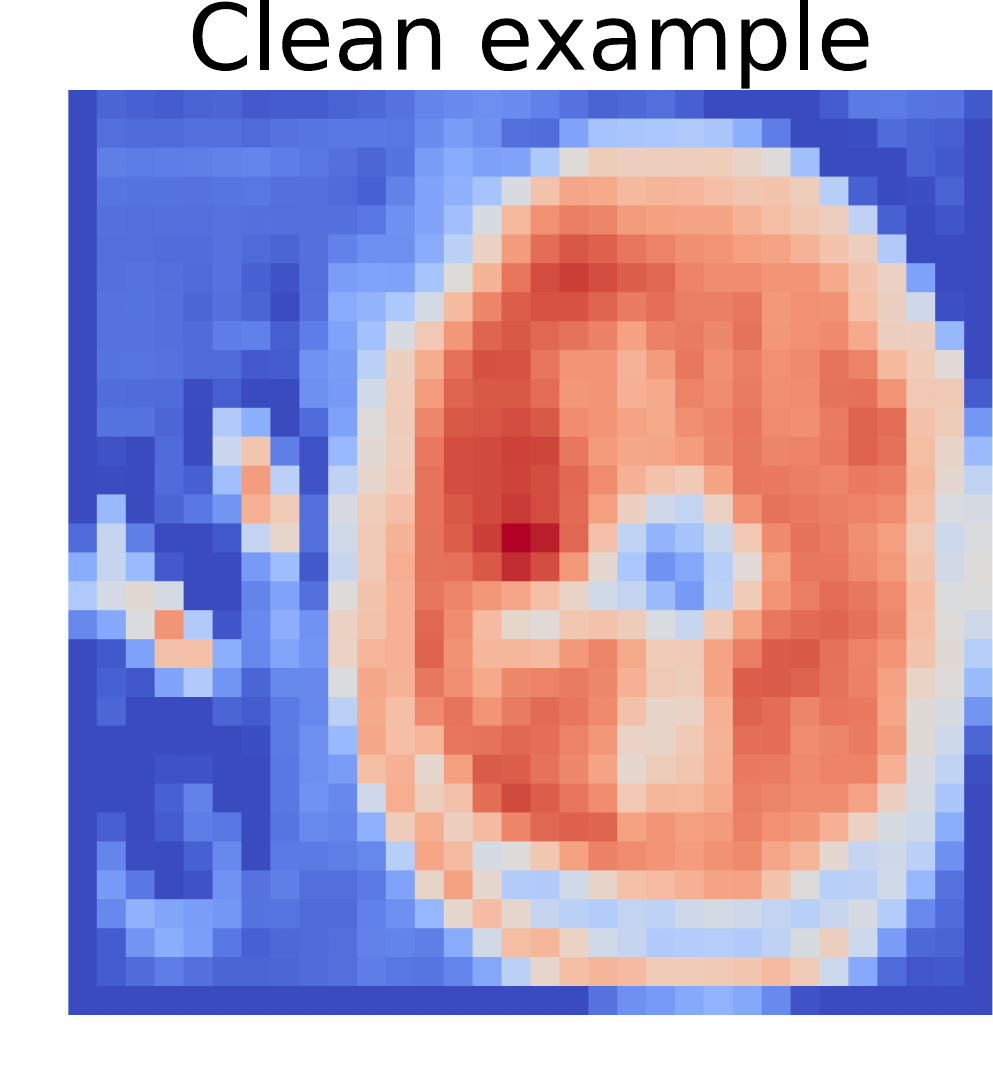}}
		\label{}
	\subfigure[]
		{\includegraphics[width=1.9cm, height=2.3cm]{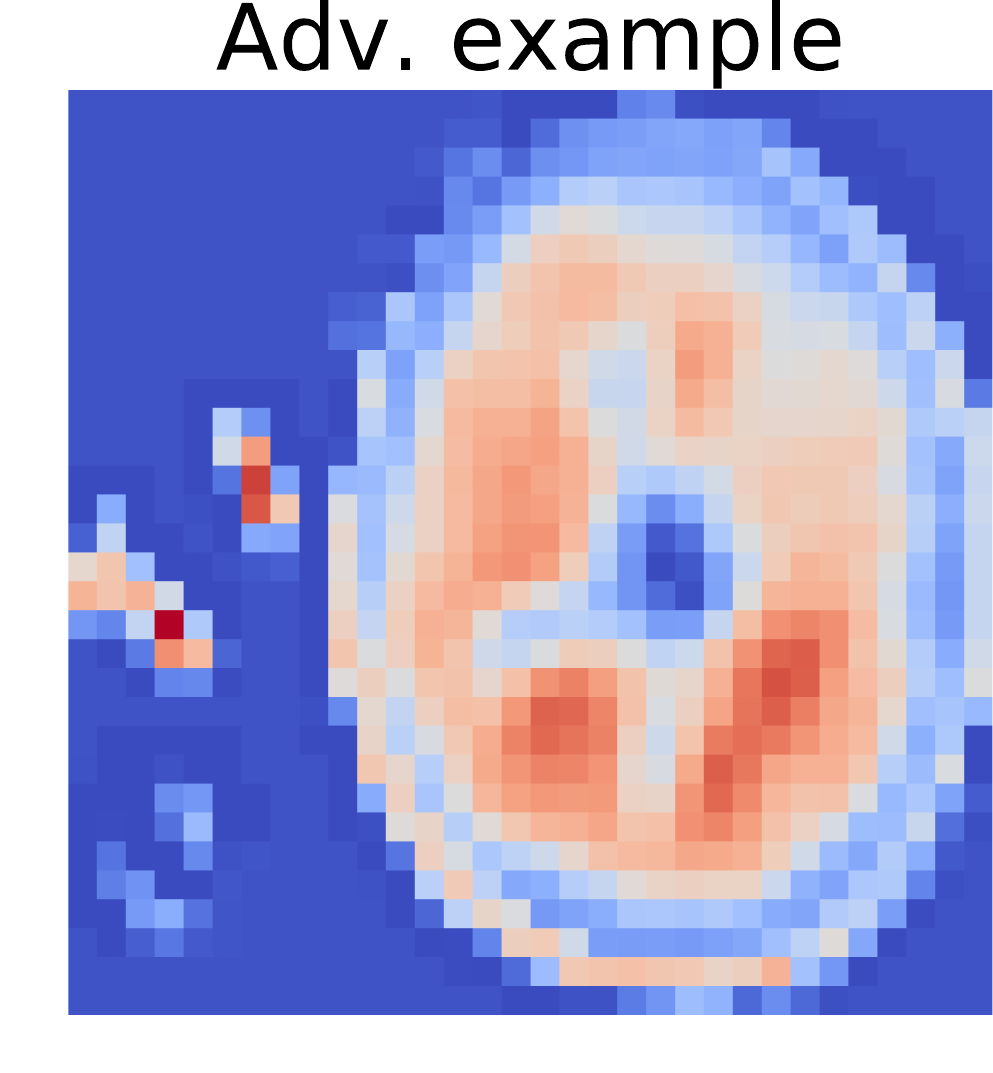}}
		\label{}
	\subfigure[]
		{\includegraphics[width=1.9cm, height=2.3cm]{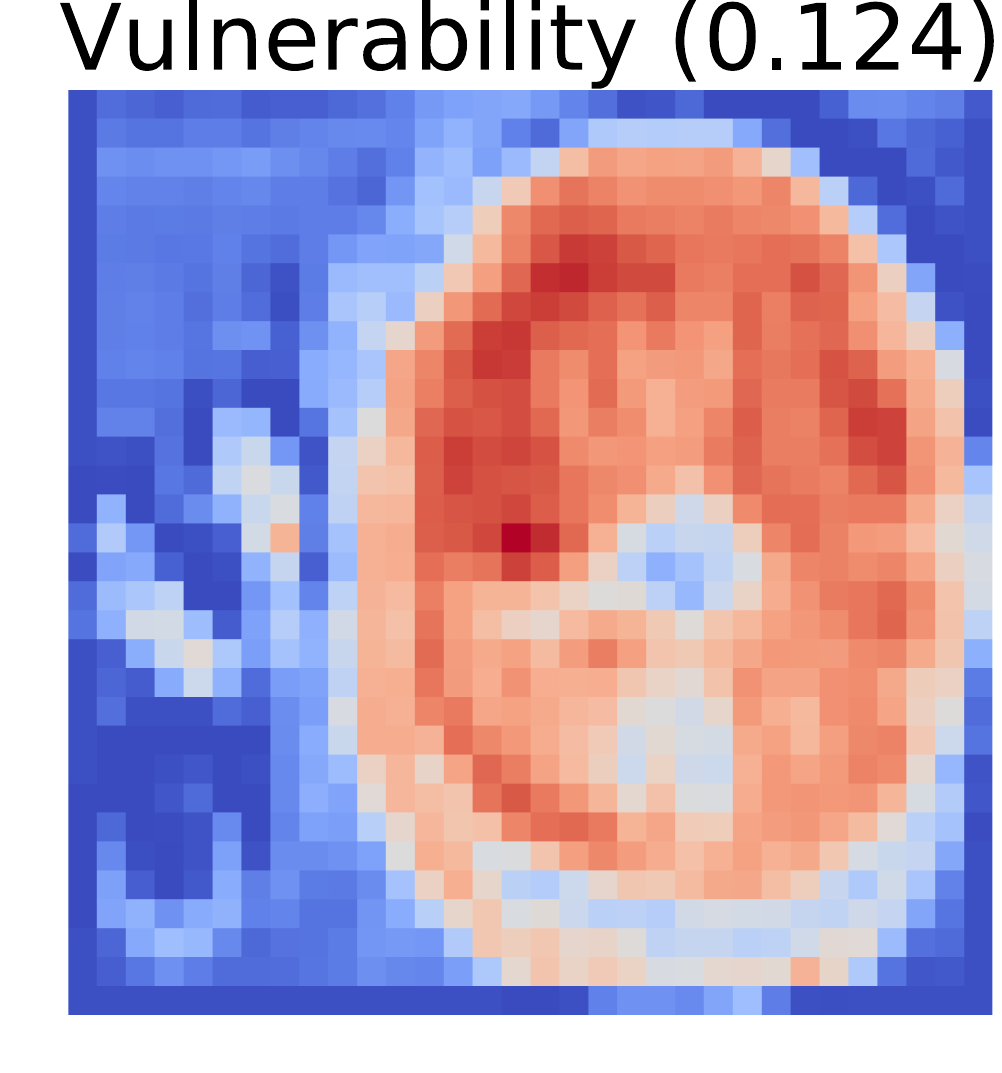}}
		\label{}
		\subfigure[]
		{\includegraphics[width=1.9cm, height=2.3cm]{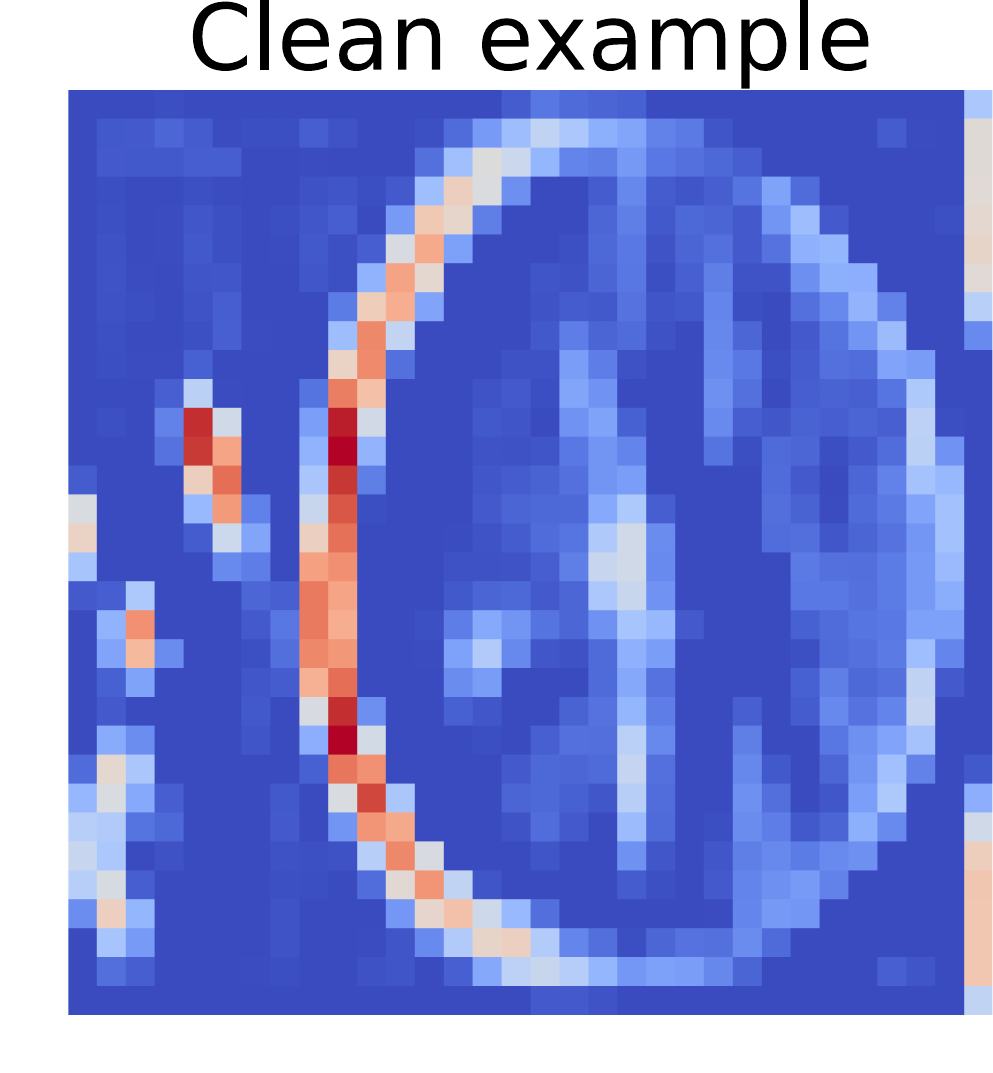}}
		\label{}
		\subfigure[]
		{\includegraphics[width=1.9cm, height=2.3cm]{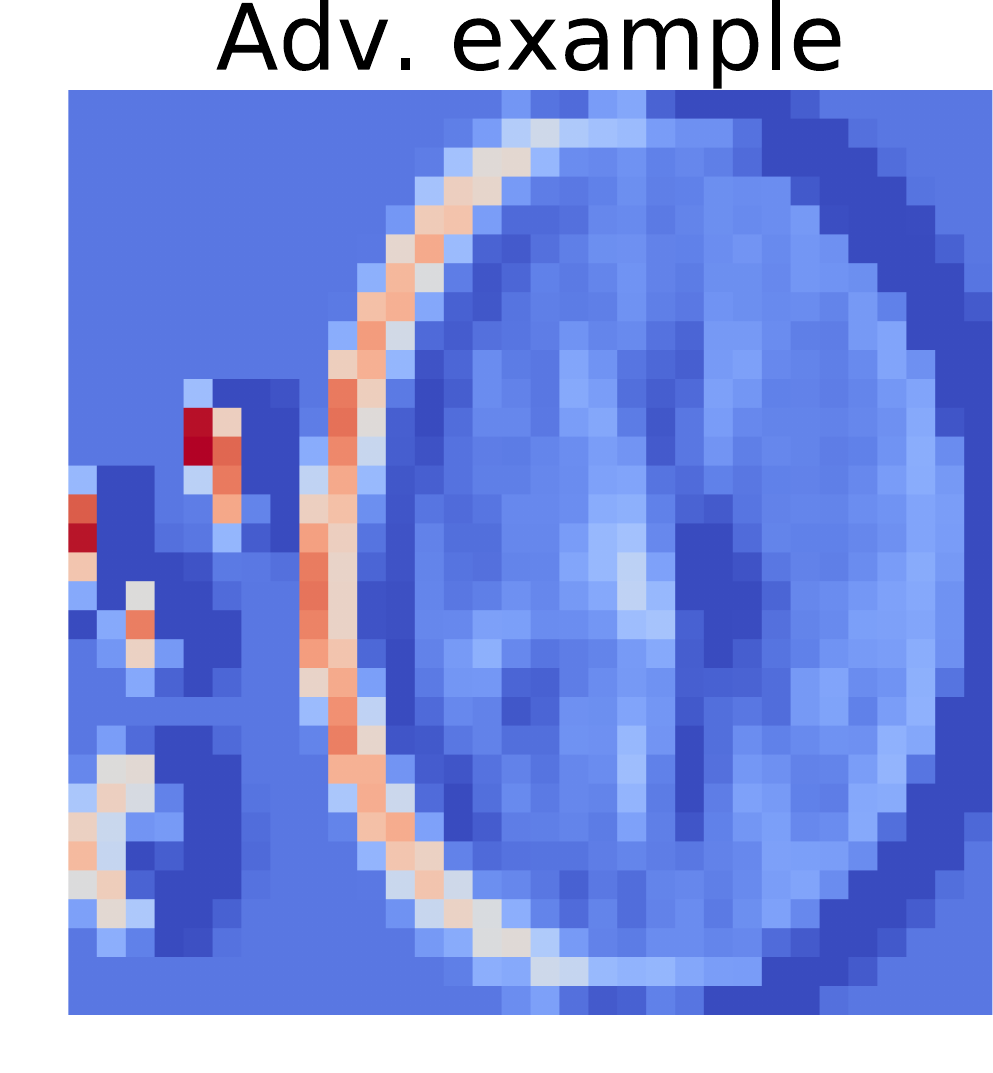}}
		\label{}
		\subfigure[]
		{\includegraphics[width=2.0cm, height=2.3cm]{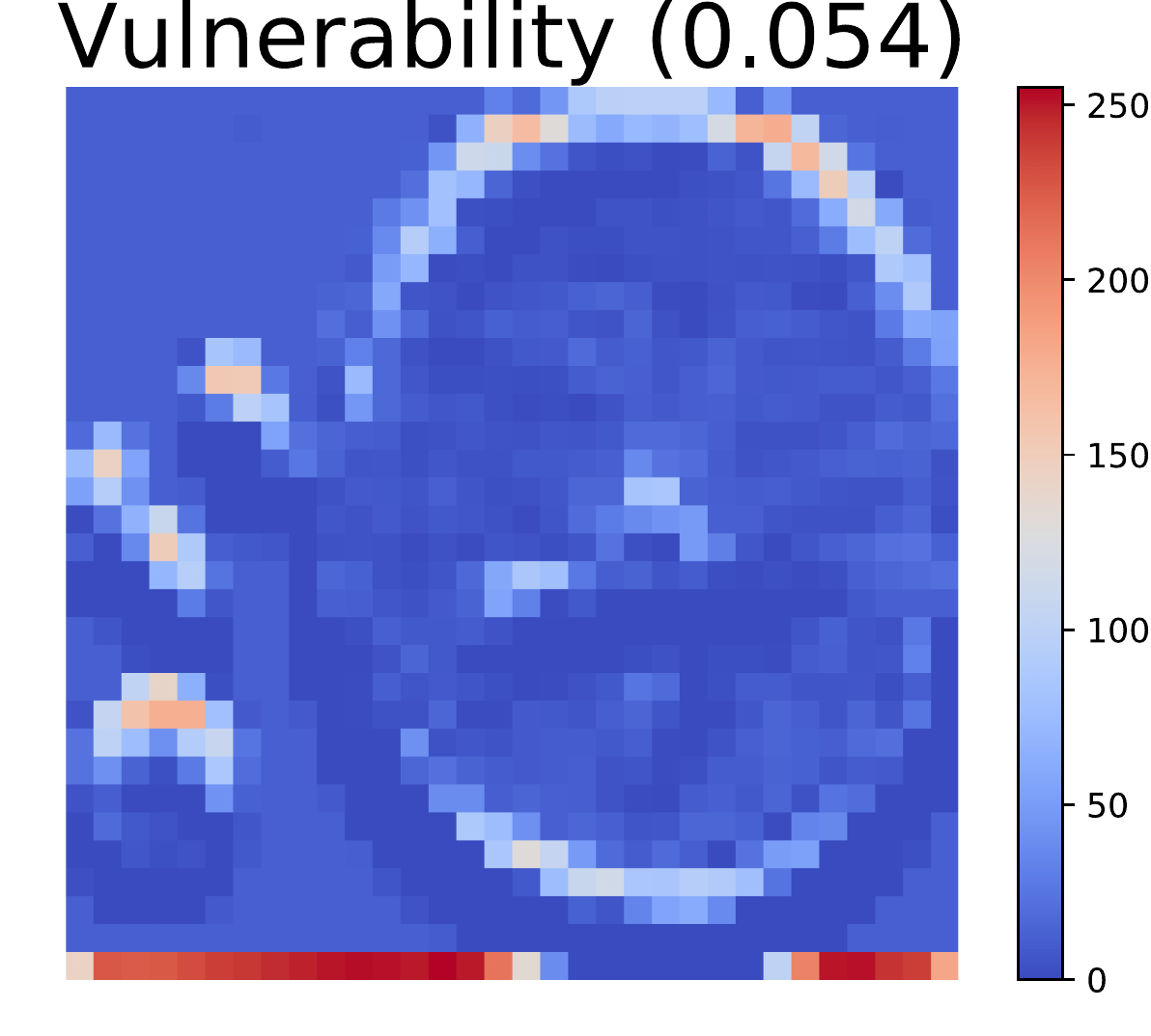}}
		\label{}
		\vspace{-0.1in}
	\caption{\small Visualization of convolutional features of first layer of adversarial trained VGG-16 network with CIFAR-100 dataset. \textbf{b) - d)} represents the vulnerable latent-feature with high vulnerability (vulnerable feature) on b) clean example, c) Adversarial example d) Vulnerability (difference between clean and adversarial example)  \textbf{e) - f)} represents the vulnerable latent-feature with low vulnerability (robust feature) on e) clean example, f) Adversarial example g) Vulnerability (difference between clean and adversarial example) \label{fig:motivation}} 
\end{figure*}